\PassOptionsToPackage{hyphens}{url}
\PassOptionsToPackage{most}{tcolorbox}
\PassOptionsToPackage{table,dvipsnames}{xcolor}
\RequirePackage{fix-cm}
\documentclass[11pt]{article}
\let\standardaddcontentsline\addcontentsline
\AddToHook{package/hyperref/before}{%
  \let\addcontentsline\standardaddcontentsline
}
\usepackage[preprint]{arxiv}
\usepackage{times}
\usepackage[utf8]{inputenc}
\usepackage[T1]{fontenc}
\usepackage{microtype}
\usepackage{amsmath,amssymb,bm}
\usepackage{graphicx,booktabs,multirow,tabularx}
\usepackage{etoolbox,float,xspace,subcaption,placeins,needspace}
\usepackage[most]{tcolorbox}
\usepackage[shortlabels,inline]{enumitem}
\usepackage{comment}
\usepackage{adjustbox}

\usepackage{colortbl}
\usepackage{caption}
\usepackage{xspace}
\usepackage{textcomp}
\usepackage{marvosym}
\definecolor{successTableBest}{RGB}{255,224,224}
\definecolor{successTableSecond}{RGB}{221,235,250}
\newcommand{\ours}{TCP4D\xspace}
\newcommand{\SE}{\ensuremath{\mathrm{SE}(3)}\xspace}

\usepackage{etoc}
\usepackage{tabularx}

\AtBeginDocument{\etocdepthtag.toc{main}}


\newlength{\tcptextwidth}
\newlength{\tcppanelwidth}
\DeclareCaptionFont{panelcaption}{\fontsize{8.5}{10}\selectfont}
\newcommand{\textwithpanel}[3][.47]{%
  \par\addvspace{0.7\baselineskip}%
  \begingroup
  \setlength{\tcptextwidth}{#1\linewidth}%
  \setlength{\tcppanelwidth}{\dimexpr\linewidth-\tcptextwidth-1.2em\relax}%
  \noindent\begin{minipage}[t]{\tcptextwidth}%
    \vspace{0pt}\raggedright\setlength{\emergencystretch}{1em}#2%
  \end{minipage}\hspace{1.2em}%
  \begin{minipage}[t]{\tcppanelwidth}%
    \vspace{0pt}#3%
  \end{minipage}%
  \par\endgroup\addvspace{0.7\baselineskip}%
}
\newenvironment{panelfigure}{%
  \captionsetup{type=figure,font=panelcaption,skip=5pt}\centering
}{\par}
\newenvironment{paneltable}{%
  \captionsetup{type=table,font=panelcaption,skip=5pt}\centering
}{\par}

\AtBeginDocument{\raggedbottom}
\setkeys{Gin}{clip=true}

\usepackage{wrapfig}
\newsavebox{\tcpfigurebox}
\newcommand{\textwithfigure}[3][.47]{%
  \par\addvspace{0.7\baselineskip}%
  \setlength{\tcppanelwidth}{\dimexpr\linewidth-#1\linewidth-1.2em\relax}%
  \savebox{\tcpfigurebox}{%
    \begin{minipage}[t]{\tcppanelwidth}#3\end{minipage}%
  }%
  \Needspace{\dimexpr\ht\tcpfigurebox+\dp\tcpfigurebox+4\baselineskip\relax}%
  \noindent\begin{wrapfigure}{r}{\tcppanelwidth}%
    \vspace{-\intextsep}%
    \usebox{\tcpfigurebox}%
  \end{wrapfigure}%
  #2\par
}

\BeforeBeginEnvironment{abstract}{\begingroup\tcbset{breakable}}
\AfterEndEnvironment{abstract}{\endgroup}
\setcitestyle{authoryear,round,citesep={;},aysep={,},yysep={;}}
\hypersetup{
  hypertexnames=false,
  pdftitle={TCP4D: A Unified Tool-Center-Point Tracking and Generation Model for Robot Manipulation},
  pdfauthor={Haosong Peng, Xiangyang Liu, Zhengshen Zhang, Yunfeng Fan, Tianyu Qi, Xin Xia, Yufeng Zhan, Wenchao Xu}
}

\title{TCP4D: A Unified Tool-Center-Point Tracking and Generation Model for Robot Manipulation}
\author{\parbox{\textwidth}{\centering\normalfont\fontsize{10}{12}\selectfont
  \mbox{Haosong Peng\textsuperscript{1*}},
  \mbox{Xiangyang Liu\textsuperscript{2*}},
  \mbox{Zhengshen Zhang\textsuperscript{3}},
  \mbox{Yunfeng Fan\textsuperscript{4}},\\
  \mbox{Tianyu Qi\textsuperscript{5}},
  \mbox{Xin Xia\textsuperscript{6}},
  \mbox{Yufeng Zhan\textsuperscript{2\Letter}},
  \mbox{Wenchao Xu\textsuperscript{1\Letter}}\endgraf
  \medskip
  {\footnotesize
    \mbox{\textsuperscript{1}\,The Hong Kong University of Science and Technology}\quad
    \mbox{\textsuperscript{2}\,Beijing Institute of Technology}\quad
    \mbox{\textsuperscript{3}\,National University of Singapore}\quad
    \mbox{\textsuperscript{4}\,The Hong Kong Polytechnic University}\quad
    \mbox{\textsuperscript{5}\,Sun Yat-sen University}\quad
    \mbox{\textsuperscript{6}\,Zhejiang University}\endgraf
    \smallskip
    Project page: \url{https://livioni.github.io/TCP4D/}\endgraf
  }
}}

\begin{document}
\setlength{\headheight}{24pt}
\fancyhf{}
\cfoot{\thepage}
\renewcommand{\headrulewidth}{0.4pt}
\fancypagestyle{firstpage}{
  \fancyhf{}
  \lhead{\includegraphics[height=18pt]{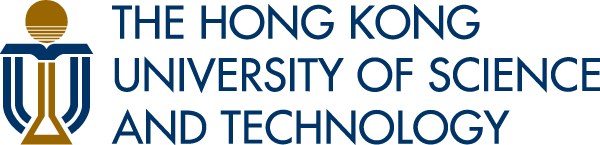}}
  \rhead{\small\textcolor[HTML]{4D4D4D}{\today\quad Preprint}}
  \cfoot{\thepage}
  \renewcommand{\headrulewidth}{0.4pt}
}
\maketitle
\begingroup
  \renewcommand{\thefootnote}{\fnsymbol{footnote}}
  \footnotetext[1]{Equal Contribution. \textsuperscript{\Letter}Corresponding Authors.}
\endgroup
\vspace{-10mm}
\begin{figure}[H]
  \centering
  \includegraphics[width=\linewidth]{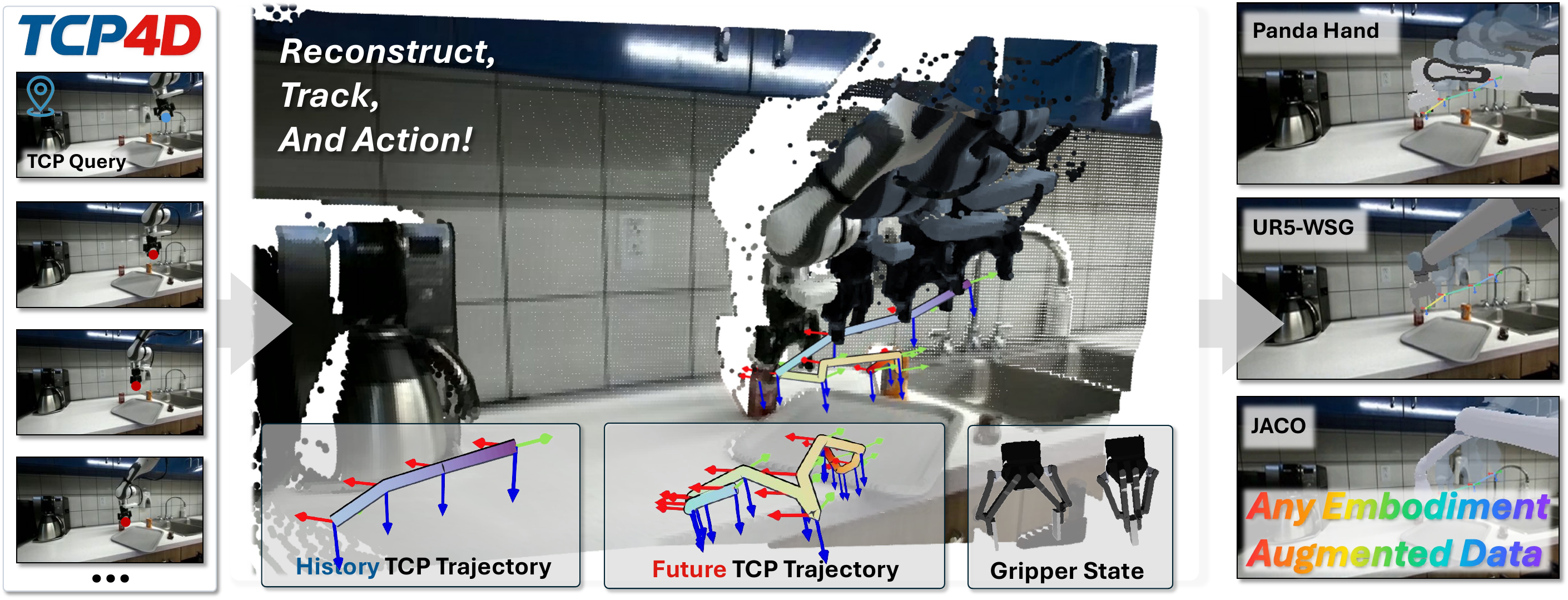}
  \caption{We propose \textbf{\ours}, a unified model that tracks tool-center-point (TCP) poses and gripper states from RGB videos and generates task-conditioned future TCP trajectories and gripper states, enabling robot manipulation and cross-embodiment data augmentation.}
  \label{fig:teaser}
\end{figure}

\begin{abstract}
Robot manipulation videos provide rich visual experience, yet recovering robot action and gripper-state annotations from RGB observations remains challenging. 
Using the embodiment-agnostic tool-center-point (TCP) pose as the action space, we present \textbf{TCP4D}, a unified model in which a shared 4D scene representation both tracks observed TCP motion and generates future TCP trajectories.
Given a monocular video and initial 2D TCP queries, the \textbf{tracker} directly recovers metric TCP poses in SE(3) and gripper states alongside scene geometry and camera parameters.
The \textbf{generator} samples local features from the same 4D representation along the recovered TCP tracks, then combines them with robot observations and language instructions to predict future TCP states. 
Our new \textbf{cross-embodiment augmentation} pipeline converts existing RGB videos into demonstrations for alternative embodiments by retargeting the recovered motion and synthesizing the corresponding observations, without recorded source-robot states.
Experiments show that the tracker achieves absolute trajectory errors of 7.60 and 16.07 mm on DROID and RoboLab and adapts to a new embodiment with only ten demonstrations. 
The generator achieves 69.19\% mean success on RoboTwin and outperforms well-established policy models on two real-world tasks, while joint training on source and augmented demonstrations improves target-robot success by 41.5\% over augmented-only training.
\end{abstract}
   
\section{Introduction}
\label{sec:intro}

\looseness=-1
Large-scale video data of robot manipulation~\cite{open_x_embodiment_rt_x_2023,agibotworld} offer a rich source of experience for learning diverse skills.
These data provide an important foundation for modern vision-language-action (VLA) models~\cite{pan2026world,intelligence2025pi05,Falcon} and world action models (WAMs)~\cite{dreamzero,fastwam,worldguide,sgwam}, whose training relies on action supervision.
However, not all videos include synchronized action annotations, and recovering these labels from RGB video alone remains a nontrivial challenge.

\looseness=-1
Recovering the end effector's motion offers a practical route to obtaining such supervision from videos.
This motion can be represented by the tool-center-point (TCP) pose (i.e., \SE), which specifies the position and orientation of the end effector's tool frame.
With a common reference frame, TCP provides a unified action representation: the same pose encodes the same task-space goal across robot platforms, even when their joint configurations differ, since each robot can reach the pose through its own inverse kinematics (IK)~\cite{OnePolicy,unifyrobotactionscamera} if within its work range.

Recent 4D point tracking methods~\cite{luo2026fourrc,karhade2026any4d,lu2026track4world,sucar2026vdpm4dvideoreconstruction} recover scene geometry and track 3D surface points from RGB video.
The TCP, however, is the origin of a virtual tool frame and can lie in free space between the gripper fingers.
Its image projection can therefore fall on a background pixel.
Moreover, point trackers predict only 3D positions not tool orientation, so even an accurate trajectory of the TCP origin does not determine the full TCP pose.
Recovering full TCP poses from these trackers therefore relies on additional robot-specific priors and multi-stage post-processing, along which errors accumulate and may lead to inaccurate poses or tracking failures.

To address this challenge, we propose \textbf{\ours}, a unified model that tracks observed TCP motion and generates future TCP trajectories from a shared 4D scene representation.
Given a monocular RGB video and an initial 2D TCP query for each arm, the \textit{Tracker} directly estimates TCP poses in \SE and gripper states throughout the observed sequence, together with metric scene geometry and camera parameters, without additional robot priors or post-processing.
The Tracker further adapts to new embodiments in a few-shot manner.
TCP poses provide an action interface for robot control, allowing predicted future TCP trajectories to be executed as robot actions.
The \textit{Generator} extracts local visual features along the tracked motion.
A Diffusion Transformer (DiT) combines these features with the shared scene representation to generate task-conditioned future TCP and gripper states, serving as a manipulation policy.

Since the recovered TCP motion is embodiment-agnostic, \ours further enables a novel cross-embodiment data augmentation pipeline that converts existing RGB videos of a source robot into demonstrations for target embodiments.
Unlike existing methods that require recorded source-robot states~\cite{chen2024roviaug,ji2025oxeauge}, our pipeline recovers TCP trajectories and gripper states directly from video, removes the source robot via segmentation and inpainting, and renders the target embodiment driven by the recovered trajectory through IK.
A single video thus yields demonstrations for any target embodiment that can reach the recovered motion, without additional data collection.

Extensive experiments validate \ours on TCP tracking, manipulation, and cross-embodiment data augmentation.
For tracking, \ours achieves absolute trajectory errors of 7.60 and 16.07\,mm on DROID and RoboLab, respectively, and 8.85\,mm on the UR5 using just ten demonstrations.
As a manipulation policy, it achieves 69.19\% average success on RoboTwin and 78.33\% across two real-world tasks.
Joint training on source and augmented demonstrations improves success rates on both source and target robots.
Our main contributions are as follows:
\begin{itemize}
    \item We introduce \ours, a unified model that directly recovers TCP poses and gripper states from RGB videos and generates future TCP trajectories conditioned on observations and instructions.
  \ours adapts effectively to a new robot embodiment with only a few demonstrations.
    \item We develop a novel cross-embodiment augmentation pipeline that transforms RGB demonstrations into target-robot training data through recovered TCP motion, without requiring recorded source-robot state trajectories.
    \item Extensive experiments across diverse tasks in simulation benchmarks and real-world settings demonstrate the strong performance of \ours in TCP tracking and manipulation, as well as the effectiveness of its cross-embodiment data augmentation.
\end{itemize}

\section{Related Work}

\noindent\textbf{4D Reconstruction and Point Tracking.}
Recent feed-forward approaches to 4D reconstruction include dense geometric prediction~\cite{wang2026vggtomega,omnivggt,sucar2026vdpm4dvideoreconstruction,lu2026track4world,jiang2026omnix} and query-based frameworks~\cite{zhang2026efficiently,luo2026fourrc} that provide flexible access to scene geometry and motion across space and time.
Beyond tracking observed motion, recent methods~\cite{zhang2026molmomotionforecastingpointtrajectories,duisterhof2026pointzero} predict future point-cloud dynamics through 3D trajectory forecasting and completion.
These approaches primarily characterize scene geometry and point positions, leaving local orientation implicit.
Concurrent work MoSE3~\cite{cheng2026mose3} extends motion estimation to dense per-pixel \SE transforms by fitting rigid motions within soft clusters of tracked scene points.
In contrast, \ours directly tracks the \SE pose of a virtual TCP, without requiring a corresponding physical surface point or rigid-body grouping.

\noindent\textbf{Point-based Manipulation Policy.}
Prior work uses predicted 2D point tracks to guide manipulation policies~\cite{bharadhwaj2024track2actpredictingpointtracks,cao2026tract,anypointtrajectory}.
In 3D, world models predict geometric motion to support planning and action learning~\cite{zhi20253dflowaction,lee2026tracegen,huang2026pointworld,wang2026unitas,hu2026tracewamlearningact3d}.
Other approaches infer waypoints or poses from visual predictions~\cite{huang20263dway,zhang2025generativevisualforesightmeets}, or incorporate motion cues into VLA policies~\cite{wang2026track4action,yuan2026motionvlainjectinggeometricmotion}.
Point-based policies also use depth-derived scene geometry to ground action prediction.
DP3~\cite{ze20243dp3} conditions action diffusion on compact features from sparse point clouds, while Act3D~\cite{gervet2023act3d3dfeaturefield} searches a 3D feature field for the next end-effector position.
3D Diffuser Actor~\cite{ke20243ddiffuseractorpolicy} denoises pose trajectories conditioned on 3D scene features.
\ours uses TCP pose and gripper state as a shared interface between motion tracking and action generation, tracking them across observed frames in a 4D scene representation reconstructed from RGB video.

\noindent\textbf{Cross-Embodiment Augmentation Pipeline.}
Existing approaches to robot data augmentation include simulation-based demonstration synthesis~\cite{mandlekar2023mimicgen,dai2024digitalcousins} and image-space semantic augmentation~\cite{yu2023rosie,bharadhwaj2024roboagent}, which expand motion and scene diversity without additional real-world collection.
For embodiment transfer, Mirage~\cite{chen2024mirage} uses test-time cross-painting to present a target robot as the source robot expected by a pretrained policy.
RoVi-Aug~\cite{chen2024roviaug} uses diffusion-based image generation to replace source-robot appearances with target embodiments while retaining the associated task-space motion.
OXE-AugE~\cite{ji2025oxeauge} further scales embodiment augmentation through simulation-based trajectory retargeting, background inpainting, and target-robot rendering.
However, RoVi-Aug and OXE-AugE assume access to recorded source end-effector poses and actions.
In contrast, \ours recovers TCP trajectories in \SE and gripper states from RGB videos and obtains target joint configurations through IK, enabling augmentation without recorded source-robot states.

\section{Methodology}
\label{sec:method}
\begin{figure}[!t]
\centering
\includegraphics[width=\linewidth]{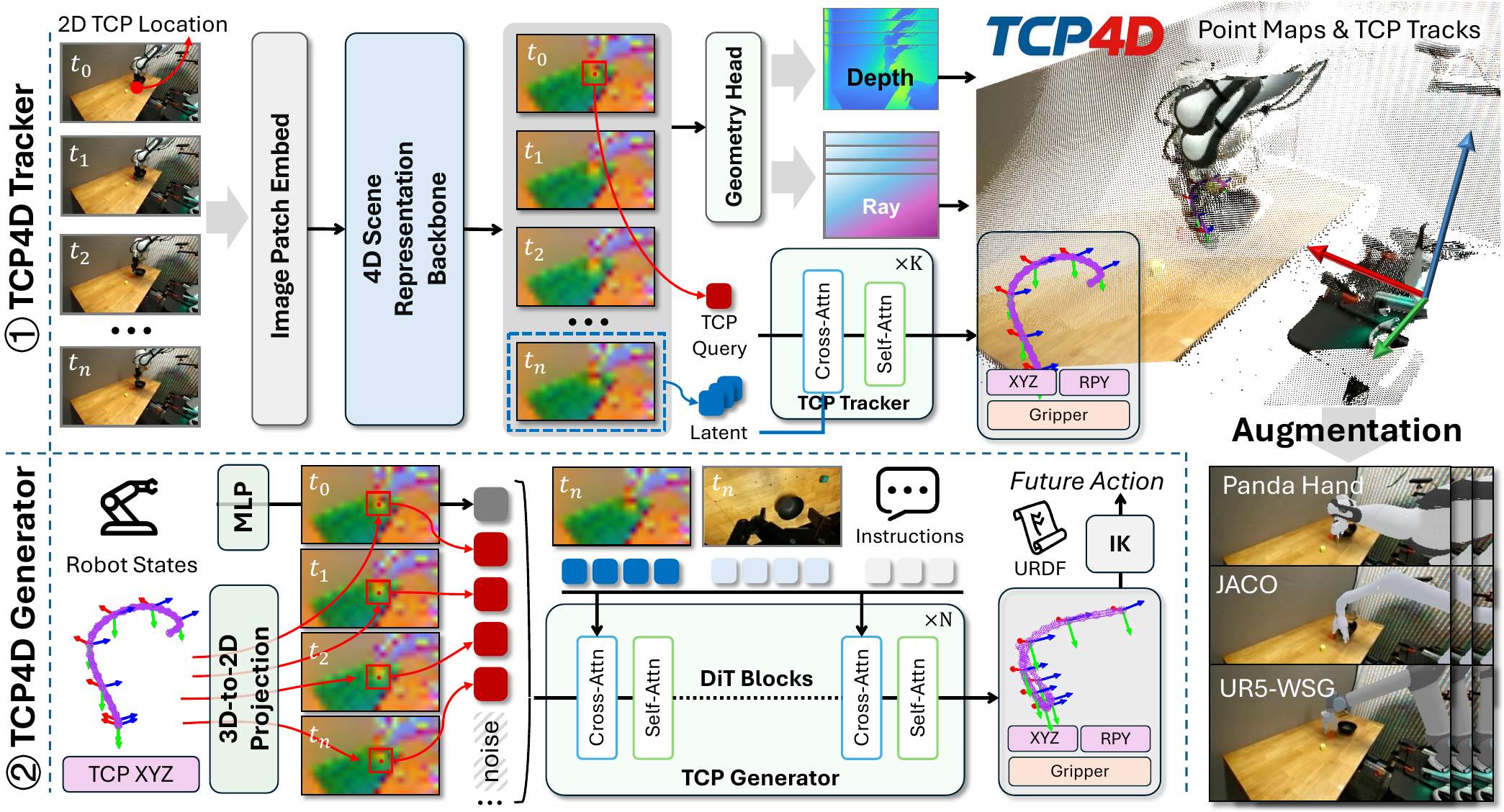}
\caption{\textbf{Overview of \ours.} \textcircled{\raisebox{0.2pt}{\scriptsize 1}}
 \ours Tracker takes a monocular video as input and constructs a 4D scene representation. 
Given the TCP’s 2D location in the initial frame as a query, it decodes the scene geometry, the TCP’s \(\mathrm{SE}(3)\) motion, and the gripper state.
\textcircled{\raisebox{0.2pt}{\scriptsize 2}}
 \ours Generator extracts local features along the TCP tracks to form queries and incorporates the robot states. 
Conditioned on the current 4D scene representation, the wrist RGB view, and language instruction, it predicts the TCP motion and gripper state over the next \(h\) steps.}
\label{fig:overview}
\vspace{-3mm}
\end{figure}

We present \ours, a feed-forward framework that tracks and predicts TCP motion in \SE from monocular videos of robot manipulation, as illustrated in Fig.~\ref{fig:overview}.
We first formulate the tasks in Sec.~\ref{sec:problem}, introduce the \ours Tracker and Generator in Sec.~\ref{sec:tracker} and~\ref{sec:generator}, and describe two-stage training in Sec.~\ref{sec:objectives}.
Finally, Sec.~\ref{sec:data} introduces a novel downstream data augmentation pipeline using \ours.

\subsection{Problem Formulation}
\label{sec:problem}

\noindent\textbf{TCP Tracking.}
Given an RGB sequence $\mathbf I=\{I_i\}_{i=1}^{N}$ and a 2D TCP query $\mathbf u_1^a=\langle x_1^a,y_1^a\rangle$ specifying the TCP's pixel coordinates in the reference frame, our goal is to track the TCP state $\mathbf x_i^a=(\mathbf T_i^a,g_i^a)\in\SE\times[0,1]$ at each observed frame. 
Here, $\mathbf T_i^a=(\mathbf R_i^a,\mathbf t_i^a)\in\SE$ denotes the TCP pose in frame $i$'s camera coordinate system, comprising orientation $\mathbf R_i^a\in\mathrm{SO}(3)$ and position $\mathbf t_i^a\in\mathbb R^3$;
$g_i^a$ denotes gripper state, and $a\in\{l,r\}$ indexes left or right robot arms. 
\ours jointly predicts scene geometry as metric depth maps $\hat{\mathbf D}_i\in\mathbb R^{H\times W}$ and camera-ray maps $\hat{\mathbf M}_i\in\mathbb R^{\frac{H}{2}\times\frac{W}{2}\times6}$ encoding camera intrinsics and extrinsics relative to the first camera frame through per-pixel ray directions and origins.
We write the \ours tracker $\mathcal T_\theta$ as:
\begin{equation}
    \bigl\{\{\hat{\mathbf x}_i^a\},\hat{\mathbf D}_i,\hat{\mathbf M}_i\bigr\}_{i=1}^{N}
    =\mathcal T_\theta(\mathbf I,\{\mathbf u_1^a\}).
    \label{eq:tracking}
\end{equation}

\noindent\textbf{TCP Generation.}
Given a language instruction $\ell$, our goal is to generate the future TCP states over an $h$-step horizon conditioned on the available observations at time $t$.
These observations comprise $L$ past and current RGB frames $\{I_i\}_{i=t-L+1}^{t}$,  the current joint and gripper state $\mathbf s_t$, and current wrist-camera RGB observations $I_t^{w,a}$.
We collect these observations as
$\mathcal O_t=(\{I_i\}_{i=t-L+1}^{t},\mathbf s_t,\{I_t^{w,a}\})$
and write the language-conditioned generator $\mathcal G_\theta$ as
\begin{equation}
    \hat{\mathbf A}_t
    =\mathcal G_\theta(\boldsymbol\epsilon;\mathcal O_t,\ell)
    =\bigl\{\hat{\mathbf x}_{t+k}^{a}\bigr\}_{k=1,\ldots,h\,},
    \label{eq:generation}
\end{equation}
where $\boldsymbol\epsilon\sim\mathcal N(\mathbf 0,\mathbf I)$ is the initial
noise for trajectory generation. 
All future TCP poses are expressed in the camera coordinate system.

\subsection{\ours Tracker}
\label{sec:tracker}

\noindent\textbf{4D Scene Representation Encoding.}
Following 4RC~\cite{luo2026fourrc}, we encode the input video into a unified 4D scene representation using a DINOv2-based Transformer~\cite{oquab2023dinov2}.
Each frame is divided into patch tokens and augmented with a learnable camera
token and a time token. 
The backbone alternates frame-wise self-attention with global self-attention across frames to capture spatial structure and temporal correspondences. 
We write
\begin{equation}
    \mathcal F=\mathcal E_\theta(\mathbf I)
    =\bigl\{(\mathbf F_i,\mathbf c_i,\mathbf e_i)\bigr\}_{i=1}^{N},
    \label{eq:scene_encoding}
\end{equation}
where $\mathbf F_i$ contains the encoded 4D representation tokens, $\mathbf c_i$ is the
camera token summarizing camera geometry, and $\mathbf e_i$ is the time token
aggregating temporal context for frame $i$. 

\noindent\textbf{Geometry Head.}
A geometry head with a Dual-DPT~\cite{lin2025depth} decoder maps the encoded 4D representation tokens to metric depth maps $\hat{\mathbf D}_i\in\mathbb R^{H\times W}$ and camera-ray maps $\hat{\mathbf M}_i\in\mathbb R^{\frac H2\times\frac W2\times6}$,
along with their confidence maps. 
A lightweight camera head additionally decodes $\mathbf c_i$ into camera pose relative to
the robot base and field of view.

\noindent\textbf{Sparse TCP Queries.}
To identify each arm's TCP, a visual query encoder $\mathcal Q_\theta$ bilinearly samples a $3\times3$ neighborhood from $\mathbf F_1$ around $\mathbf u_1^a$. 
A residual adapter and learned arm and offset embeddings encode local appearance and query identity. 
We extract these sparse query tokens at four backbone levels and repeat them across target frames.
We first encode TCP queries in the reference frame and then decode their
representations at each target frame:
\begin{equation}
        \mathbf Q_1=\mathcal Q_\theta
        \bigl(\mathbf F_1,\{\mathbf u_1^a\}_a\bigr),\quad
        \mathbf Z_i=\mathcal D_\theta
        \bigl(\mathbf Q_1,\mathbf F_i;\mathbf e_i\bigr),
    \label{eq:sparse_decoder}
\end{equation}
where $\mathbf Q_1$ contains the sparse TCP query tokens for all arms and $\mathbf Z_i$ contains their decoded representations at frame $i$.
Each layer first applies cross-attention, with the query sequence providing queries and $\mathbf F_i$ providing keys and values, and then applies self-attention within the query sequence. 
Reference-frame sampling positions and target-frame patch positions enter the attention
operations through 2D RoPE~\cite{rope}. 
The time token $\mathbf e_i$ additionally modulates the self-attention and feed-forward layers through AdaLN~\cite{perez2018film}.
Finally, a lightweight prediction head regresses TCP translation $\hat{\mathbf t}_i^a$, a continuous 6D rotation representation~\cite{zhou2019continuity}, gripper state $\hat g_i^a$, and pose confidence. 
The rotation prediction is projected to $\mathrm{SO}(3)$ to form $\hat{\mathbf T}_i^a$. 
Together with $\hat g_i^a$, this yields the tracked state $\hat{\mathbf x}_i^a$.

\subsection{\ours Generator}
\label{sec:generator}

\noindent\textbf{Tracking-conditioned Representation.}
The Generator builds on the 4D scene feature $\mathbf F_i$ and TCP tracks $\{\hat{\mathbf x}_i^a\}_{i=t-L+1}^t$.
We project each tracked TCP translation $\hat{\mathbf t}_i^a$ into its observed frame using the predicted camera intrinsics to obtain image queries $\hat{\mathbf u}_i^a$, and extract local TCP tokens with the shared encoder $\mathcal Q_\theta(\mathbf F_i,\{\hat{\mathbf u}_i^a\}_a)$.
We add spatial and temporal embeddings to the local TCP tokens to encode their image locations and observation order.

\noindent\textbf{Conditional Motion Generation.}
A DiT takes a single sequence formed by concatenating an embedding of the robot state $\mathbf s_t$, the local TCP tokens, and noisy future-action tokens.
Self-attention jointly updates this sequence, while cross-attention uses current scene features $\mathbf F_t$, wrist-image features, and language features as keys and values.
A shared frozen DINOv2 encodes the wrist images $I_t^{w,a}$, and a frozen T5-base encoder~\cite{raffel2020exploring} encodes the instruction $\ell$.
We train the DiT with flow matching~\cite{lipman2023flow}.
For a normalized target action sequence $\mathbf A$, the noisy input and target velocity are
\begin{equation}
    \mathbf A_\tau=(1-\tau)\boldsymbol\epsilon+\tau\mathbf A,
    \qquad \mathbf v^\star=\mathbf A-\boldsymbol\epsilon,
    \label{eq:flow_path}
\end{equation}
where $\tau\in[0,1)$ is flow time.
The output head maps the future-action tokens to predicted velocities $\mathbf v_\phi(\mathbf A_\tau,\tau;\mathcal O_t,\ell)$.
At inference, four Euler updates transform $\boldsymbol\epsilon$ into the
future action sequence, which is denormalized and decoded into
$\hat{\mathbf A}_t$.

\subsection{Training Objectives}
\label{sec:objectives}

We train \ours in two stages with different objectives and training
strategies. Loss definitions, normalization, and component weights are
provided in Appx.~\ref{app:losses}.

\noindent\textbf{Stage 1: Geometry and TCP Reconstruction.}
We jointly train the visual backbone, geometry head, and TCP Tracker to reconstruct metric scene geometry and TCP states for the available arms:
\begin{equation}
    \mathcal L_{\mathrm{stage1}}
    =\mathcal L_{\mathrm{geo}}
    +\lambda_{\mathrm{tcp}}\mathcal L_{\mathrm{tcp}}^{(1)}.
    \label{eq:stage1}
\end{equation}
The geometry loss $\mathcal L_{\mathrm{geo}}$ combines confidence-weighted depth and ray regression with a multiscale depth-gradient term.
TCP supervision $\mathcal L_{\mathrm{tcp}}^{(1)}$ combines a confidence-weighted rigid-point pose error, rotational alignment, temporal velocity consistency, and gripper prediction.

\noindent\textbf{Stage 2: Conditional TCP Generation.}
We initialize the backbone and Tracker modules from Stage 1 and jointly fine-tune them with the conditioning modules and DiT using:
\begin{equation}
    \mathcal L_{\mathrm{stage2}}
    =\mathcal L_{\mathrm{geo}}
    +\lambda_{\mathrm{tcp}}\mathcal L_{\mathrm{tcp}}^{(2)}
    +\lambda_{\mathrm{fm}}\mathcal L_{\mathrm{fm}}.
    \label{eq:stage2}
\end{equation}
The flow-matching loss $\mathcal L_{\mathrm{fm}}$ supervises the predicted velocities of position, rotation, and gripper components for both arms.
$\mathcal L_{\mathrm{tcp}}^{(2)}$ retains pose and gripper supervision and omits the temporal velocity term.

\subsection{Cross-Embodiment Augmentation Pipeline}
\label{sec:data}

\begin{figure}[!t]
\centering
\includegraphics[width=\linewidth]{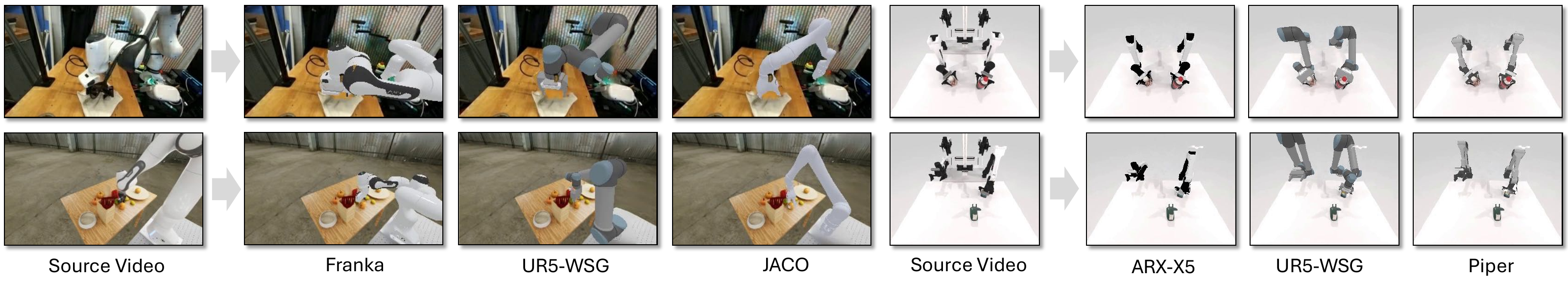}
\caption{\textbf{Example of our cross-embodiment augmented data.} 
Given an RGB video, \ours reconstructs the TCP’s 3D trajectory and gripper state, enabling augmentation on any embodiment that satisfies \textit{reachability}.  
More visualizations in Appx.~\ref{app:pipeline}.  }
\label{fig:vis3}
\vspace{-3mm}
\end{figure}

\textwithfigure[.47]{%
As shown in Fig.~\ref{fig:pipeline}, we use \ours Tracker to recover TCP motion and the gripper state from source demonstration videos and retarget them to alternative embodiments.
We transform the predicted camera-frame poses into a common world frame, align TCP conventions, and solve target-robot inverse kinematics with fixed bases while preserving the world-frame trajectories.
Source-robot masks are obtained either with SAM3~\cite{carion2026sam} or by projecting the source URDF using recorded robot states.
We erase the masked robot, complete the background with a video inpainting model~\cite{zhou2023propainter}, and composite the rendered target robot into the completed scene.
Depth is optional and can refine robot masks when available.
For example, when the projected robot model overlaps with other scene elements in the image, depth information helps determine which is in front and refine the robot mask accordingly.
Fig.~\ref{fig:vis3} shows an example of our cross-embodiment augmented data.
Further implementation and export details are provided in Appx.~\ref{app:pipeline}.
}{%
\begin{panelfigure}
\centering
\includegraphics[width=\linewidth]{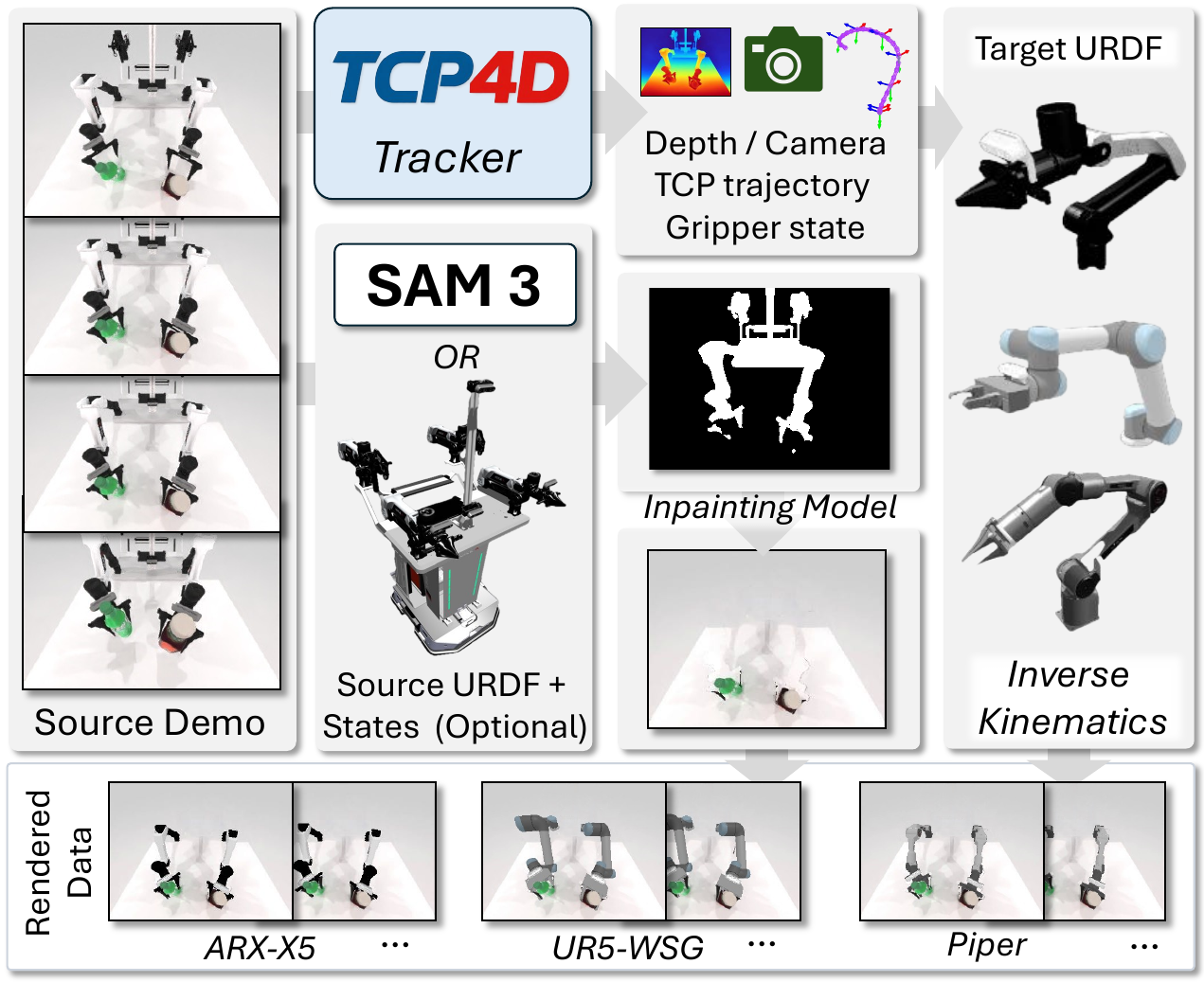}
\caption{\textbf{Cross-embodiment augmentation pipeline.}
Given a video, the TCP4D tracker reconstructs the TCP trajectory and gripper state. 
We remove and inpaint the original robot region, then retarget the motion to alternative embodiments using IK.
}
\label{fig:pipeline}
\end{panelfigure}
}

\begin{table}[t]
\centering
\caption{\textbf{\ours Tracker evaluation on DROID and Robolab.}
All results are camera-wise macro averages.
Depth is evaluated within 3\,m with per-frame median alignment.
\SE errors use Sim(3) alignment.
CoTracker3 uses VGGT-$\Omega$ depth.
The top-2 results are highlighted as \protect\colorbox{successTableBest}{best} and \protect\colorbox{successTableSecond}{second}, respectively.
The metric definitions are given in Appx.~\ref{app:tracking_metrics}.}
\label{tab:main_table}
\begingroup
\setlength{\tabcolsep}{3.5pt}
\renewcommand{\arraystretch}{1.15}
\small
\resizebox{0.9\linewidth}{!}{%
\begin{tabular}{@{}l*{10}{c}@{}}
\toprule
\multicolumn{11}{c}{\textbf{DROID (Eval set)}} \\
\midrule
 & \multicolumn{3}{c}{Depth} & \multicolumn{2}{c}{TCP Tracking} & \multicolumn{3}{c}{\SE} & Gripper & \\
\cmidrule(lr){2-4}\cmidrule(lr){5-6}\cmidrule(lr){7-9}\cmidrule(lr){10-10}
Method & AbsRel $\downarrow$ & MAE $\downarrow$ & $\delta_1\uparrow$ & APD $\uparrow$ & MPE $\downarrow$ & ATE $\downarrow$ & RRE $\downarrow$ & RTE $\downarrow$ & Acc. $\uparrow$ & SR $\uparrow$ \\
 & -- & (mm) & (\%) & (\%) & (mm) & (mm) & ($^\circ$) & (mm) & (\%) & (\%) \\
\midrule
VGGT-$\Omega$~\cite{wang2026vggtomega} & 0.2279 & 224.10 & 66.19 & - & - & - & - & - & - & - \\
DA3-GIANT~\cite{lin2025depth} & 0.2271 & 224.35 & 64.98 & - & - & - & - & - & - & - \\
MapAnything~\cite{keetha2026mapanything} & 0.1994 & 187.42 & 71.90 & - & - & - & - & - & - & - \\
\midrule
4RC$^*$~\cite{luo2026fourrc} & 0.2272 & 215.86 & 68.03 & 8.74 & 107.56 & 84.63 & 8.37 & 13.29 & - & 79.56 \\
Track4World$^*$~\cite{lu2026track4world} & 0.1870 & 182.56 & 72.74 & 3.68 & 186.95 & 92.55 & 10.35 & 12.42 & - & 66.05 \\
Any4D$^*$~\cite{karhade2026any4d} & 0.2132 & 206.72 & 69.29 & 2.32 & 242.01 & 96.71 & 10.76 & 13.10 & - & 56.09 \\
VDPM~\cite{sucar2026vdpm4dvideoreconstruction}$^*$ & 0.2328 & 220.05 & 70.26 & 4.67 & 191.06 & 85.40 & 8.05 & 11.58 & - & 76.82 \\
CoTracker3$^*$~\cite{cotracker} & - & - & - & \cellcolor{successTableSecond}14.04 & \cellcolor{successTableSecond}91.76 & 70.35 & 16.14 & 22.60 & - & 63.23 \\
MoSE3$^\dagger$~\cite{cheng2026mose3} & \cellcolor{successTableSecond}0.1619 & \cellcolor{successTableSecond}156.23 & \cellcolor{successTableSecond}79.16 & 12.38 & 103.43 & \cellcolor{successTableSecond}67.65 & \cellcolor{successTableSecond}1.27 & \cellcolor{successTableSecond}5.96 & - & \cellcolor{successTableSecond}89.64 \\
\midrule
\ours & \cellcolor{successTableBest}0.0330 & \cellcolor{successTableBest}31.27 & \cellcolor{successTableBest}97.74 & \cellcolor{successTableBest}74.78 & \cellcolor{successTableBest}11.85 & \cellcolor{successTableBest}7.60 & \cellcolor{successTableBest}1.03 & \cellcolor{successTableBest}3.40 & \cellcolor{successTableBest}95.84 & \cellcolor{successTableBest}100.00 \\
\bottomrule
\end{tabular}
}
\par\medskip
\resizebox{0.9\linewidth}{!}{%
\begin{tabular}{@{}l*{10}{c}@{}}
\toprule
\multicolumn{11}{c}{\textbf{RoboLab (Unseen)}} \\
\midrule
 & \multicolumn{3}{c}{Depth} & \multicolumn{2}{c}{TCP Tracking} & \multicolumn{3}{c}{\SE} & Gripper & \\
\cmidrule(lr){2-4}\cmidrule(lr){5-6}\cmidrule(lr){7-9}\cmidrule(lr){10-10}
Method & AbsRel $\downarrow$ & MAE $\downarrow$ & $\delta_1\uparrow$ & APD $\uparrow$ & MPE $\downarrow$ & ATE $\downarrow$ & RRE $\downarrow$ & RTE $\downarrow$ & Acc. $\uparrow$ & SR $\uparrow$ \\
 & -- & (mm) & (\%) & (\%) & (mm) & (mm) & ($^\circ$) & (mm) & (\%) & (\%) \\
\midrule
VGGT-$\Omega$~\cite{wang2026vggtomega} & 0.1742 & 114.29 & 74.77 & - & - & - & - & - & - & - \\
DA3-GIANT~\cite{lin2025depth} & 0.1625 & 104.68 & 78.93 & - & - & - & - & - & - & - \\
MapAnything~\cite{keetha2026mapanything} & \cellcolor{successTableSecond}0.1426 & 93.13 & 83.34 & - & - & - & - & - & - & - \\
\midrule
4RC$^*$~\cite{luo2026fourrc} & 0.1795 & 108.26 & 79.12 & 4.92 & 164.53 & 91.37 & 7.16 & 8.59 & - & 90.62 \\
Track4World$^*$~\cite{lu2026track4world} & 0.1849 & 112.01 & 73.57 & 2.44 & 208.05 & 95.43 & 9.50 & 8.88 & - & 82.85 \\
Any4D$^*$~\cite{karhade2026any4d} & 0.1507 & 95.56 & 83.89 & 1.15 & 238.02 & 106.49 & 8.41 & 8.42 & - & 74.63 \\
VDPM~\cite{sucar2026vdpm4dvideoreconstruction}$^*$ & 0.1690 & 108.93 & 81.31 & \cellcolor{successTableSecond}9.90 & \cellcolor{successTableSecond}111.90 & \cellcolor{successTableSecond}70.90 & \cellcolor{successTableSecond}5.78 & 8.59 & - & \cellcolor{successTableSecond}96.33 \\
CoTracker3$^*$~\cite{cotracker} & - & - & - & 8.78 & 131.39 & 85.12 & 15.80 & 17.63 & - & 75.38 \\
MoSE3$^\dagger$~\cite{cheng2026mose3} & \cellcolor{successTableBest}0.1142 & \cellcolor{successTableBest}67.23 & \cellcolor{successTableSecond}89.37 & 8.58 & 130.18 & 85.45 & \cellcolor{successTableBest}0.99 & \cellcolor{successTableBest}4.65 & - & 94.18 \\
\midrule
\ours & 0.1718 & \cellcolor{successTableSecond}92.61 & \cellcolor{successTableBest}92.17 & \cellcolor{successTableBest}43.71 & \cellcolor{successTableBest}28.42 & \cellcolor{successTableBest}16.07 & 6.80 & \cellcolor{successTableSecond}4.83 & \cellcolor{successTableBest}81.92 & \cellcolor{successTableBest}100.00 \\
\bottomrule
\end{tabular}
}
\par\smallskip
{\footnotesize\raggedright
$^*$ TCP poses are recovered by rigid fitting of point tracks.
$^\dagger$ MoSE3 uses its native dense-point \SE outputs followed by post-processing.\par}
\endgroup
\end{table}

\section{Experiments}
We evaluate \ours across four dimensions: (1) TCP tracking performance, (2) the effectiveness of cross-embodiment data augmentation, (3) robot manipulation performance of the TCP generator, and (4) real-world manipulation performance on a physical robot platform.

\subsection{Training Setup}
\label{sec:training_setup}

\noindent\textbf{Datasets and Benchmark.}
We use the subset of DROID~\cite{khazatsky2024droid} annotated by PointWorld~\cite{huang2026pointworld}, comprising 33,966 episodes (67,932 camera sequences) of Franka Panda 7DOF Arm. We use 90\% of the episodes for training and randomly sample 100 episodes from the remaining 10\% as the evaluation set.
For in-the-wild evaluation, we additionally collect 100 episodes spanning 100 distinct tasks in RoboLab~\cite{yang2026robolab}.
To evaluate robot manipulation, we use RoboTwin~\cite{chen2025robotwin} and collect 50 clean demonstrations per task for 48 tasks, using the third-person view.
We train on these demonstrations and evaluate under both \emph{clean2clean} and \emph{clean2random} settings.

\noindent\textbf{Implementation Details.}
In Stage~1, we train two separate TCP trackers on DROID and RoboTwin, respectively, initializing their visual backbones and geometry modules from pretrained 4RC~\cite{luo2026fourrc} weights.
RGB image resolutions are $320\times180$ for DROID and $320\times240$ for RoboTwin.
We sample Stage~1 clips with random starting frames, a per-clip temporal stride randomly chosen from 1--5, and temporal reversal.
In Stage~2, we use only RoboTwin data to train the TCP generator.
Each sample contains $L=4$ consecutive observation frames paired with the next $h=32$ TCP states, preserving chronological order.
The Stage~1 tracker and the full \ours model contain approximately 1.522B and 2.272B parameters, respectively, including frozen modules and retained heads.
Further implementation details are provided in Appx.~\ref{app:implementation}.

\subsection{\ours Tracking Performance}
\label{sec:tracking_performance}

\noindent\textbf{Baselines.}
We compare \ours with VGGT-Omega~\cite{wang2026vggtomega}, DA3~\cite{lin2025depth}, MapAnything~\cite{keetha2026mapanything}, 4RC~\cite{luo2026fourrc}, VDPM~\cite{sucar2026vdpm4dvideoreconstruction}, Any4D~\cite{karhade2026any4d}, Track4World~\cite{lu2026track4world}, CoTracker3~\cite{cotracker}, and MoSE3~\cite{cheng2026mose3} on DROID and RoboLab videos.
Since no existing method directly tracks the TCP, we post-process the point tracks predicted by point-tracking baselines to recover the TCP trajectory and pose.

\noindent\textbf{Evaluation Metrics.}
We report AbsRel, MAE, and $\delta_1$ for depth within 3\,m; \textit{Average Percentage of Points} (APD) and \textit{Mean Position Error} (MPE) for 3D TCP tracking; \textit{Absolute Trajectory Error} (ATE), \textit{Relative Rotation Error} (RRE), and \textit{Relative Translation Error} (RTE) for \SE tracking; and tracking \textit{Success Rate} (SR). 
Gripper Accuracy (Acc.) uses a 0.5 threshold.
Metric definitions and evaluation protocols are provided in Appx.~\ref{app:tracking_metrics}.

\begin{figure}[!t]
\centering
\includegraphics[width=\linewidth]{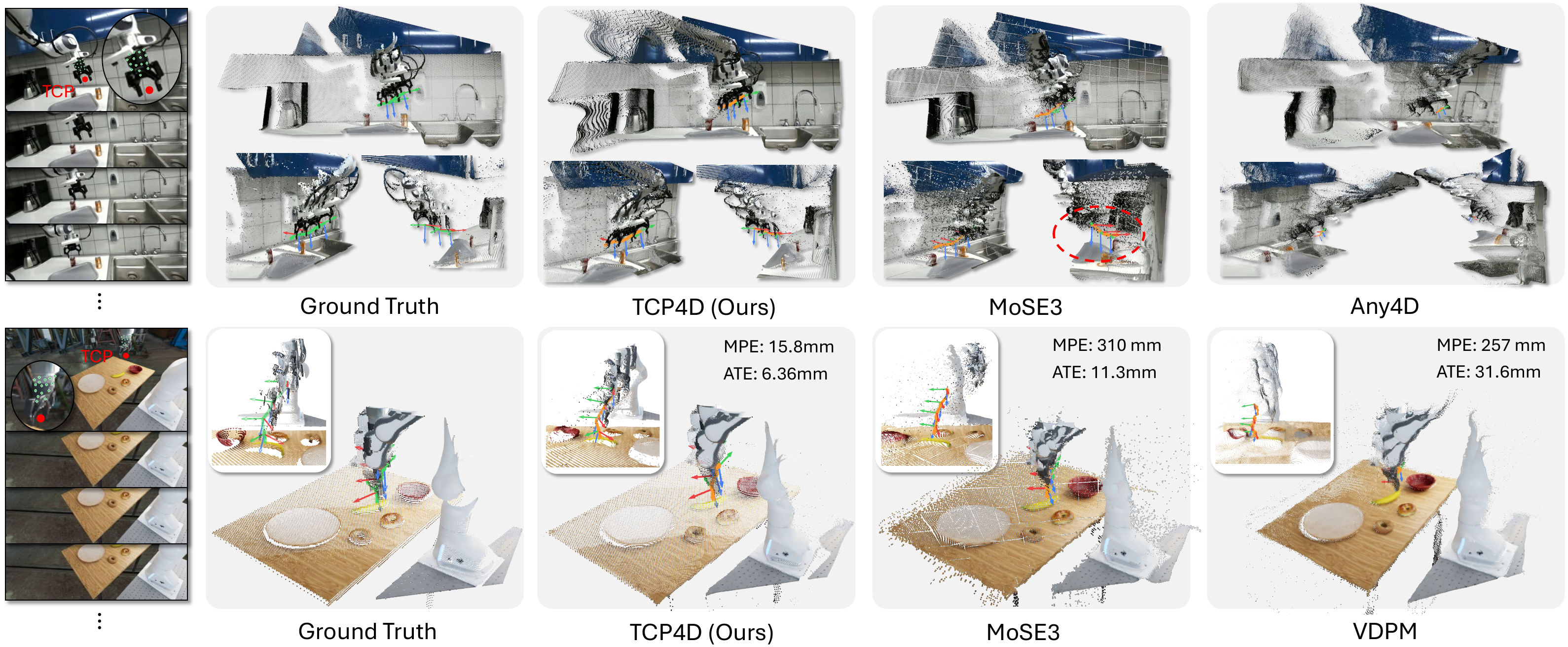}
\caption{\textbf{Qualitative comparison of TCP tracking.} We compare TCP tracking by \ours against dense tracking baselines followed by post-processing. 
The left side shows the input video. 
The red point in the first frame marks the query used by \ours, while the green points indicate the rigid point group tracked by the baselines.
More visualizations are provided in Appx.~\ref{app:tracker}. }
\label{fig:vis1}
\end{figure}

\noindent\textbf{Results.}
As shown in Table~\ref{tab:main_table}, \ours achieves state-of-the-art performance on most evaluated metrics across both benchmarks.
On DROID and RoboLab, it achieves APD scores of 74.78\% and 43.71\%, respectively, more than four times those of the strongest baselines, and reduces ATE to 7.60 and 16.07 mm.
\ours also achieves 100\% SR on both datasets, successfully processing all evaluated tracking windows, compared with the best baseline SRs of 89.64\% and 96.33\%, respectively.
Competing point-tracking methods rely on tracking rigid point groups and fitting rigid transformations in post-processing to recover TCP poses; inaccurate or inconsistent point tracks can yield unreliable pose estimates or prevent successful recovery.
By directly predicting TCP poses and gripper states within a unified feed-forward framework, \ours avoids this recovery step, delivering accurate and reliable tracking even in unseen manipulation scenarios.
Fig.~\ref{fig:vis1} visually compares \ours with other baselines.
More results can be found in Appx.~\ref{app:tracker}.

\noindent\textbf{Few-Shot Cross-Embodiment Adaptation.}
\ours demonstrates strong few-shot cross-embodiment adaptation capabilities after pretraining on the DROID dataset. 
We evaluate it on the \emph{Berkeley AUTOLAB UR5} dataset~\cite{BerkeleyUR5Website}, which contains 1,000 demonstration episodes using UR5-WSG across different tasks, comprising 97,939 video frames in total.
We fine-tune \ours under two settings: \emph{(1) three-shot} transfer using \emph{three} randomly selected episodes and \emph{(2) few-shot} transfer using \emph{ten} randomly selected episodes.
\textwithpanel[.47]{%
As shown in Table~\ref{tab:autolab_tracking}, \emph{three-shot} adaptation improves APD from 14.39\% for MoSE3 to 48.27\% and reduces ATE from 44.56 to 12.22\,mm.
With ten episodes, APD further increases to 64.28\%, ATE decreases to 8.85\,mm, and gripper accuracy reaches 98.48\%.
These results demonstrate that \ours Tracker can be easily transferred to new environments and robot embodiments using only a small amount of data.
The visualizations are provided in Appx.~\ref{app:adaptation}.
}{%
\begin{paneltable}
\centering
\footnotesize
\setlength{\tabcolsep}{2.5pt}
\caption{\textbf{\ours few-shot cross-embodiment adaptation on Berkeley AUTOLAB UR5.}
Results are averages over the 100-episode evaluation set.}
\label{tab:autolab_tracking}
\begin{adjustbox}{max width=\linewidth}
\begin{tabular}{lcccccc}
\toprule
 & \multicolumn{2}{c}{TCP Tracking} & \multicolumn{3}{c}{\SE} & Gripper \\
\cmidrule(lr){2-3}\cmidrule(lr){4-6}\cmidrule(l){7-7}
Method & APD $\uparrow$ & MPE $\downarrow$ & ATE $\downarrow$ & RRE $\downarrow$ & RTE $\downarrow$ & Acc. $\uparrow$ \\
 & (\%) & (mm) & (mm) & ($^\circ$) & (mm) & (\%) \\
\midrule
MoSE3$^\dagger$ & 14.39 & 88.07 & 44.56 & 1.17 & 6.45 & -- \\
Three-Shot & 48.27 & 18.48 & 12.22 & 1.03 & 4.28 & 97.05 \\
Few-Shot & \textbf{64.28} & \textbf{11.08} & \textbf{8.85} & \textbf{0.96} & \textbf{3.87} & \textbf{98.48} \\
\bottomrule
\end{tabular}
\end{adjustbox}
\end{paneltable}

}

\subsection{Effectiveness of Cross-embodiment Data}

\label{sec:augmentation_effectiveness}

\begin{figure}[t]
\centering
\includegraphics[width=\linewidth]{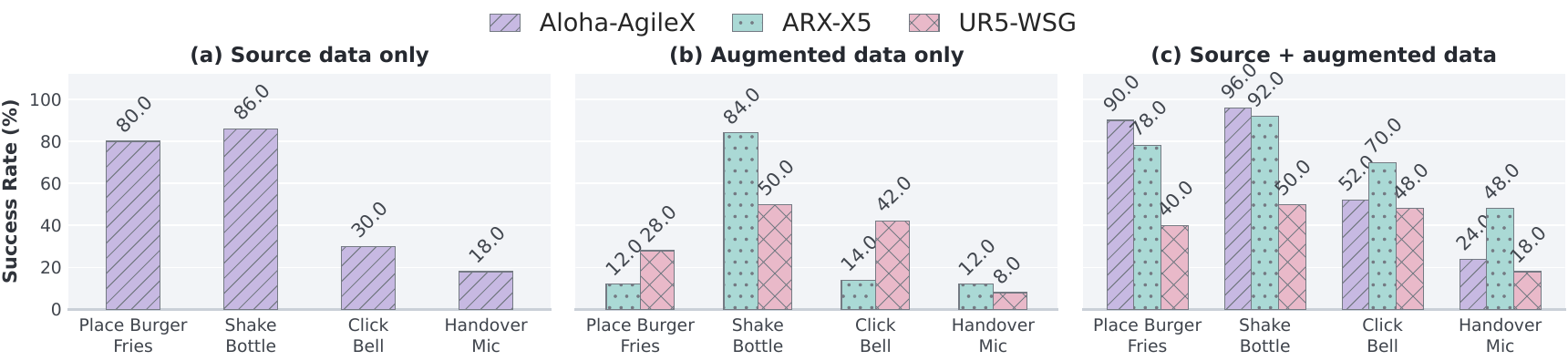}
\caption{\textbf{Cross-embodiment policy evaluation.} Task success rates under three training-data settings on three embodiments. For each task, we collect 50 source demonstrations on Aloha-AgileX and augment them to ARX-X5 and UR5-WSG using our pipeline.}
\label{fig:augmentation_training}
\vspace{-3mm}
\end{figure}

\noindent\textbf{Setup.}
We evaluate whether cross-embodiment demonstrations generated using our pipeline benefit downstream policy learning on four tasks. 
For each task, we collect 50 source demonstrations on Aloha-AgileX and augment them to ARX-X5 and UR5-WSG, yielding 199 and 198 target demonstrations across the four tasks, respectively.
We use DP~\cite{chi2025diffusion} under three training-data settings: \emph{(a) source-only}, trained on Aloha-AgileX; \emph{(b) augmented-only}, trained separately on each target robot; and \emph{(c) source + augmented}, jointly trained on all three robots and evaluated on each.
Policies are trained separately for each task under the same budget. 
Each evaluated policy--embodiment pair is tested over 50 episodes per task. 
The implementation details and data statistics are in Appx.~\ref{app:cross_exp}.

\noindent\textbf{Results.}
As shown in Fig.~\ref{fig:augmentation_training}, combining source and augmented demonstrations improves Aloha performance on all four tasks, raising its mean success rate from 53.5\% to 65.5\%. \textit{Click Bell} shows the largest improvement, from 30\% to 52\%, demonstrating the effectiveness of augmented demonstrations for policy learning on the source embodiment.
On ARX-X5, joint training increases the mean success rate from 30.5\% to 72.0\% relative to augmented-only training. The gains span all four tasks, including \textit{Place Burger Fries} (12\% $\rightarrow$ 78\%) and \textit{Click Bell} (14\% $\rightarrow$ 70\%). On UR5-WSG, joint training improves the mean success rate from 32.0\% to 39.0\%, with gains on \textit{Place Burger Fries} (28\% $\rightarrow$ 40\%), \textit{Click Bell} (42\% $\rightarrow$ 48\%), and \textit{Handover Mic} (8\% $\rightarrow$ 18\%). 
These results show that our pipeline enables policy learning on target robots without original demonstrations from those embodiments, while also improving performance when augmented demonstrations are combined with existing data, highlighting the utility of \ours for cross-embodiment data augmentation.

\begin{figure}[!t]
\centering
\includegraphics[width=\linewidth]{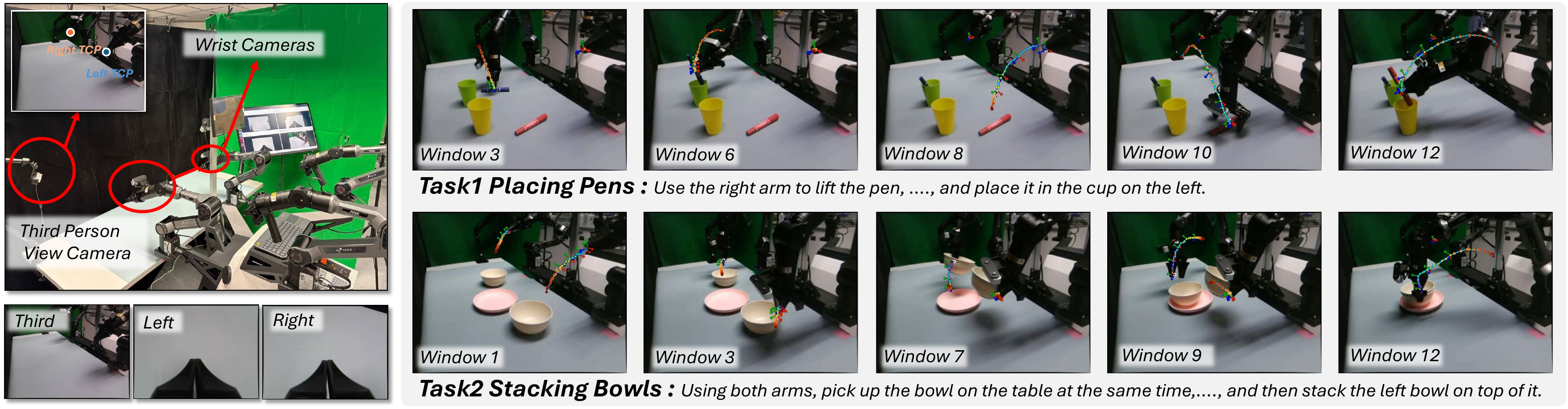}
\caption{\textbf{Real-world experiments setup and task rollouts.}
The right panel shows the rollout of tasks \textit{1) placing pens} and \textit{2) stacking bowls}, with the predicted TCP trajectory over the next 32 steps displayed in each inference window.
The color gradient from blue to orange indicates the temporal order of future actions.
Complete rollout visualizations are in Appx.~\ref{app:real_world_per}.  }
\label{fig:real_rollout}
\vspace{-3mm}
\end{figure}

\subsection{\ours Manipulation Performance}
\label{sec:manipulation_performance}

\noindent\textbf{Results.}
We evaluate \ours on 48 RoboTwin tasks with 100 trials per task and report the success rates in Table~\ref{tab:robotwin2}.
\ours achieves the highest reported average success rate of 69.19\% under \emph{clean2clean}.
It surpasses established manipulation policies ACT, RDT, and DP3 by 39.94\%, 35.27\%, and 14.92\%, respectively, and also outperforms several VLA models, including EventVLA (65.71\%), GalaxeaVLA (62.73\%), and $\pi_0$ (45.21\%).
These results support the effectiveness of predicting future TCP motion from 4D scene representations and highlight the potential of this paradigm for robotic manipulation.
Complete results for all 48 tasks are provided in Appx.~\ref{app:robotwin}.

The average success rate drops to 10.94\% under \emph{clean2random}, where environmental
randomization introduces a distribution shift.
Our TCP generator is trained solely on \textit{clean} demonstrations.
Incorporating large embodiment data pretraining or 3D geometric data from a broader mix of sources could improve generalization to randomized environments, a direction we leave for future work.

\begin{table}[!ht]
\centering
\caption{\textbf{Average success rates (\%) on RoboTwin~2.0 over 48 tasks.}
“View” indicates the primary input camera view used by each method.
The top-2 results in each column are highlighted as
\protect\colorbox{successTableBest}{best} and
\protect\colorbox{successTableSecond}{second}, respectively.}
\label{tab:robotwin2}
\begingroup
\setlength{\tabcolsep}{4pt}
\resizebox{0.85\linewidth}{!}{%
\begin{tabular*}{\linewidth}{@{\extracolsep{\fill}}c|lccc}
\toprule
\textbf{Type} & \textbf{Method} & \textbf{View} & \textbf{Easy (C2C)} & \textbf{Hard (C2R)} \\
\midrule
\multirow{7}{*}{\rotatebox[origin=c]{90}{\textbf{VLA\&WAM}}}
  & EventVLA~\cite{yang2026eventvla}              & head  & \cellcolor{successTableSecond}65.71 & 14.50 \\
    & AHA-WAM~\cite{cai2026ahawam}                & head  & 64.38 & 2.77 \\
  & Xiaomi Robotics-0~\cite{cai2026xiaomirobotics0} & head  & 62.79 & \cellcolor{successTableBest}17.85 \\
  & GalaxeaVLA~\cite{liu2026GalaxeaVLA}           & head  & 62.73 & 12.31 \\
  & Discrete Forcing~\cite{wang2026discreteforcing} & head  & 59.25 & 11.04 \\
  & StarVLA~\cite{community2026starvla}           & head  & 46.73 & 3.21 \\
  & $\pi_0$~\cite{black2026pi0}                  & head  & 45.21 & \cellcolor{successTableSecond}16.15 \\
\midrule
\multirow{5}{*}{\rotatebox[origin=c]{90}{\textbf{Policy}}}
  & DP3~\cite{ze20243dp3}                       & head  & 54.27 & 4.04 \\
  & RDT~\cite{liu2025rdt}                       & head  & 33.92 & 13.25 \\
  & ACT~\cite{zhao2023act}                       & head  & 29.25 & 1.79 \\
  & DP~\cite{chi2025diffusion}                       & head  & 27.31 & 0.67\\
\cmidrule(l){2-5}
  & \ours  (Ours)                                     & third & \cellcolor{successTableBest}69.19 & 10.94 \\
\bottomrule
\end{tabular*}
}
\endgroup
\end{table}

\subsection{Real-World Evaluation}

\noindent\textbf{Real-world setup.}
We conduct real-world experiments on the CoBoT Magic platform.
We evaluate two tasks: \textit{1) placing pens} and \textit{2) stacking bowls} with the left and right arms.
Both tasks require large TCP displacements.
We collect 60 demonstration trajectories per task for training.
We use the cuRobo~\cite{curobo_v2} library to solve IK.

\noindent\textbf{Evaluation protocol.}
We compare \ours with two established policies, DP3~\cite{ze20243dp3} and ACT~\cite{zhao2023act}, using the same demonstration data for fine-tuning and the same evaluation protocol for all methods.
For each task and method, we conduct 30 trials and report binary task success and a stage-wise progress score.
The task-specific progress score counts the number of completed stages in a predefined task sequence, up to a maximum of four, and its rubric is summarized in Appx.~\ref{app:real_world_per}.

\textwithfigure[.51]{%
\noindent\textbf{Results.}
As shown in Fig.~\ref{fig:real_results}, \ours outperforms DP3 and ACT on both real-world tasks in success rate and progress score.
It achieves success rates of 83.33\% on \textit{placing pens} and 73.33\% on \textit{stacking bowls}.
Averaged across the two tasks, \ours attains a success rate of 78.33\%, compared with 63.33\% for DP3 and 31.67\% for ACT, yielding gains of 15.00 and 46.67 percentage points, respectively.
}{%
\begin{panelfigure}
\centering
\includegraphics[width=\linewidth]{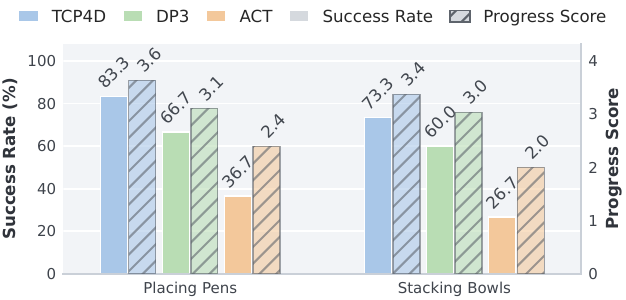}
\caption{\textbf{Real-world task performance.}
 Solid bars report success rates using the left axis, while hatched bars report average progress scores using the right axis. }
\label{fig:real_results}
\end{panelfigure}
}
Its mean progress score is 3.50 out of 4, indicating that it completes more task stages across the trials.
DP3 conditions its policy on sparse 3D point clouds reconstructed from depth observations.
In contrast, \ours infers scene geometry and TCP motion directly from RGB videos, achieving higher task success without requiring measured depth at inference.
These results demonstrate the effectiveness of our 4D scene geometric representation for manipulation while reducing reliance on depth sensing at deployment.

\FloatBarrier
\section{Conclusion}
\label{sec:conclusion}

We presented \ours, a unified model using a shared 4D scene representation to track TCP motion from RGB video and generate future actions conditioned on the recovered tracks.
Our cross-embodiment pipeline converts videos into target-robot demonstrations.
Experiments demonstrate accurate tracking, few-shot adaptation, effective manipulation, and improved policy learning with augmented data.
\ours has the potential to scale up with more diverse data and robot embodiments.

\clearpage
\bibliography{main}

\begin{thebibliography}{73}
\providecommand{\natexlab}[1]{#1}

\bibitem[{Bharadhwaj et~al.(2024{\natexlab{a}})Bharadhwaj, Mottaghi, Gupta, and Tulsiani}]{bharadhwaj2024track2actpredictingpointtracks}
Homanga Bharadhwaj, Roozbeh Mottaghi, Abhinav Gupta, and Shubham Tulsiani. 2024{\natexlab{a}}.
\newblock \href {https://doi.org/10.1007/978-3-031-73116-7\_18} {Track2act: Predicting point tracks from internet videos enables generalizable robot manipulation}.
\newblock In \emph{Computer Vision - {ECCV} 2024 - 18th European Conference, Milan, Italy, September 29-October 4, 2024, Proceedings, Part {LXXVI}}, volume 15134 of \emph{Lecture Notes in Computer Science}, pages 306--324. Springer.

\bibitem[{Bharadhwaj et~al.(2024{\natexlab{b}})Bharadhwaj, Vakil, Sharma, Gupta, Tulsiani, and Kumar}]{bharadhwaj2024roboagent}
Homanga Bharadhwaj, Jay Vakil, Mohit Sharma, Abhinav Gupta, Shubham Tulsiani, and Vikash Kumar. 2024{\natexlab{b}}.
\newblock \href {https://doi.org/10.1109/icra57147.2024.10611293} {{RoboAgent}: Generalization and efficiency in robot manipulation via semantic augmentations and action chunking}.
\newblock In \emph{2024 IEEE International Conference on Robotics and Automation (ICRA)}, pages 4788--4795. IEEE.

\bibitem[{Black et~al.(2026)Black, Brown, Driess, Esmail, Equi, Finn, Fusai, Groom, Hausman, Ichter, Jakubczak, Jones, Ke, Levine, Li-Bell, Mothukuri, Nair, Pertsch, Shi, Tanner, Vuong, Walling, Wang, and Zhilinsky}]{black2026pi0}
Kevin Black, Noah Brown, Danny Driess, Adnan Esmail, Michael Equi, Chelsea Finn, Niccolo Fusai, Lachy Groom, Karol Hausman, Brian Ichter, Szymon Jakubczak, Tim Jones, Liyiming Ke, Sergey Levine, Adrian Li-Bell, Mohith Mothukuri, Suraj Nair, Karl Pertsch, Lucy~Xiaoyang Shi, and 5 others. 2026.
\newblock \href {https://arxiv.org/abs/2410.24164} {$\pi_0$: A vision-language-action flow model for general robot control}.
\newblock \emph{Preprint}, arXiv:2410.24164.

\bibitem[{Cai et~al.(2026{\natexlab{a}})Cai, Ling, Chu, Liu, Kang, Liang, Xu, Mao, Zhang, Yang, Ying, Zheng, and Mu}]{cai2026ahawam}
Jisong Cai, Long Ling, Shiwei Chu, Zhongshan Liu, Jiayue Kang, Zhixuan Liang, Wenjie Xu, Yinan Mao, Weinan Zhang, Xiaokang Yang, Ru~Ying, Ran Zheng, and Yao Mu. 2026{\natexlab{a}}.
\newblock \href {https://arxiv.org/abs/2606.09811} {{AHA-WAM:Asynchronous} horizon-adaptive world-action modeling with observation-guided context routing}.
\newblock \emph{Preprint}, arXiv:2606.09811.

\bibitem[{Cai et~al.(2026{\natexlab{b}})Cai, Guo, He, Jin, Li, Lin, Liu, Liu, Ma, Ma, Qiu, Qu, Su, Sun, Wang, Wang, Wang, Wu, Xiang, Yang, Ye, Zhang, and Zhou}]{cai2026xiaomirobotics0}
Rui Cai, Jun Guo, Xinze He, Piaopiao Jin, Jie Li, Bingxuan Lin, Futeng Liu, Wei Liu, Fei Ma, Kun Ma, Feng Qiu, Heng Qu, Yifei Su, Qiao Sun, Dong Wang, Donghao Wang, Yunhong Wang, Rujie Wu, Diyun Xiang, and 4 others. 2026{\natexlab{b}}.
\newblock \href {https://arxiv.org/abs/2602.12684} {Xiaomi-robotics-0: An open-sourced vision-language-action model with real-time execution}.
\newblock \emph{Preprint}, arXiv:2602.12684.

\bibitem[{Cao et~al.(2026)Cao, Ji, Zhang, Ge, Fei-Fei, Wu, and Huang}]{cao2026tract}
Zhi Cao, Howard Ji, Kevin Zhang, Kuangzhi Ge, Li~Fei-Fei, Jiajun Wu, and Huang Huang. 2026.
\newblock Tract: Bridging robot control and visual prediction with visual tracks.
\newblock \emph{arXiv preprint arXiv:2608.24101}.

\bibitem[{Carion et~al.(2026)Carion, Gustafson, Hu, Debnath, Hu, Coll{-}Vinent, Ryali, Alwala, Khedr, Huang, Lei, Ma, Guo, Kalla, Marks, Greer, Wang, Sun, R{\"{a}}dle, Afouras, Mavroudi, Xu, Wu, Zhou, Momeni, Hazra, Ding, Vaze, Porcher, Li, Li, Kamath, Cheng, Doll{\'{a}}r, Ravi, Saenko, Zhang, and Feichtenhofer}]{carion2026sam}
Nicolas Carion, Laura Gustafson, Yuan{-}Ting Hu, Shoubhik Debnath, Ronghang Hu, Didac~Suris Coll{-}Vinent, Chaitanya Ryali, Kalyan~Vasudev Alwala, Haitham Khedr, Andrew Huang, Jie Lei, Tengyu Ma, Baishan Guo, Arpit Kalla, Markus Marks, Joseph Greer, Meng Wang, Peize Sun, Roman R{\"{a}}dle, and 19 others. 2026.
\newblock \href {https://proceedings.iclr.cc/paper\_files/paper/2026/hash/e0982cbc81401df3430ee1ff780dc7a2-Abstract-Conference.html} {{SAM} 3: Segment anything with concepts}.
\newblock In \emph{The Fourteenth International Conference on Learning Representations, {ICLR} 2026, Rio de Janeiro, Brazil, April 23-27, 2026}. proceedings.iclr.cc.

\bibitem[{Chen et~al.()Chen, Adebola, and Goldberg}]{BerkeleyUR5Website}
Lawrence~Yunliang Chen, Simeon Adebola, and Ken Goldberg.
\newblock Berkeley {UR5} demonstration dataset.
\newblock https://sites.google.com/view/berkeley-ur5/home.

\bibitem[{Chen et~al.(2024{\natexlab{a}})Chen, Dharmarajan, Hari, Xu, Vuong, and Goldberg}]{chen2024mirage}
Lawrence~Yunliang Chen, Karthik Dharmarajan, Kush Hari, Chenfeng Xu, Quan Vuong, and Ken Goldberg. 2024{\natexlab{a}}.
\newblock \href {https://doi.org/10.15607/RSS.2024.XX.069} {{MIRAGE:} cross-embodiment zero-shot policy transfer with cross-painting}.
\newblock In \emph{Robotics: Science and Systems XX, Delft, The Netherlands, July 15-19, 2024}.

\bibitem[{Chen et~al.(2024{\natexlab{b}})Chen, Xu, Dharmarajan, Cheng, Keutzer, Tomizuka, Vuong, and Goldberg}]{chen2024roviaug}
Lawrence~Yunliang Chen, Chenfeng Xu, Karthik Dharmarajan, Richard Cheng, Kurt Keutzer, Masayoshi Tomizuka, Quan Vuong, and Ken Goldberg. 2024{\natexlab{b}}.
\newblock \href {https://proceedings.mlr.press/v270/chen25a.html} {Rovi-aug: Robot and viewpoint augmentation for cross-embodiment robot learning}.
\newblock In \emph{Conference on Robot Learning, 6-9 November 2024, Munich, Germany}, volume 270 of \emph{Proceedings of Machine Learning Research}, pages 209--233. {PMLR}.

\bibitem[{Chen et~al.(2025)Chen, Chen, Chen, Cai, Liu, Li, Liang, Lin, Ge, Gu et~al.}]{chen2025robotwin}
Tianxing Chen, Zanxin Chen, Baijun Chen, Zijian Cai, Yibin Liu, Zixuan Li, Qiwei Liang, Xianliang Lin, Yiheng Ge, Zhenyu Gu, and 1 others. 2025.
\newblock Robotwin 2.0: A scalable data generator and benchmark with strong domain randomization for robust bimanual robotic manipulation.
\newblock \emph{arXiv preprint arXiv:2506.18088}.

\bibitem[{Cheng et~al.(2026)Cheng, Li, Xia, Cai, Du, and Wang}]{cheng2026mose3}
Jiahuan Cheng, Zhiyi Li, Tian Xia, Ruojin Cai, Yilun Du, and Qianqian Wang. 2026.
\newblock {MoSE3}: Learning world-space {SE} (3) at every pixel.
\newblock \emph{arXiv preprint arXiv:2610.03716}.

\bibitem[{Chi et~al.(2023)Chi, Feng, Du, Xu, Cousineau, Burchfiel, and Song}]{chi2025diffusion}
Cheng Chi, Siyuan Feng, Yilun Du, Zhenjia Xu, Eric Cousineau, Benjamin Burchfiel, and Shuran Song. 2023.
\newblock \href {https://doi.org/10.15607/RSS.2023.XIX.026} {Diffusion policy: Visuomotor policy learning via action diffusion}.
\newblock In \emph{Robotics: Science and Systems XIX, Daegu, Republic of Korea, July 10-14, 2023}.

\bibitem[{Community(2026)}]{community2026starvla}
StarVLA Community. 2026.
\newblock \href {https://arxiv.org/abs/2604.05014} {{StarVLA}: A lego-like codebase for vision-language-action model developing}.
\newblock \emph{Preprint}, arXiv:2604.05014.

\bibitem[{Contributors et~al.(2026)Contributors, Bu, Cai, Chen, Cui, Ding, Feng, Gao, He, Hu, Huang, Jiang, Jiang, Jing, Li, Li, Liu, Liu, Lu, Luo, Luo, Mu, Niu, Pan, Pang, Qiao, Ren, Ruan, Shan, Shen, Shi, Shi, Shi, Sima, Song, Wang, Wang, Wei, Xie, Xu, Yan, Yang, Yang, Yang, Yao, Zeng, Zhang, Zhang, Zhao, Zhao, Zhao, and Zhu}]{agibotworld}
AgiBot~World Contributors, Qingwen Bu, Jisong Cai, Li~Chen, Xiuqi Cui, Yan Ding, Siyuan Feng, Shenyuan Gao, Xindong He, Xuan Hu, Xu~Huang, Shu Jiang, Yuxin Jiang, Cheng Jing, Hongyang Li, Jialu Li, Chiming Liu, Yi~Liu, Yuxiang Lu, and 33 others. 2026.
\newblock \href {https://arxiv.org/abs/2503.06669} {{AgiBot} world colosseo: A large-scale manipulation platform for scalable and intelligent embodied systems}.
\newblock \emph{Preprint}, arXiv:2503.06669.

\bibitem[{Dai et~al.(2024)Dai, Wong, Jiang, Wang, Gokmen, Zhang, Wu, and Fei{-}Fei}]{dai2024digitalcousins}
Tianyuan Dai, Josiah Wong, Yunfan Jiang, Chen Wang, Cem Gokmen, Ruohan Zhang, Jiajun Wu, and Li~Fei{-}Fei. 2024.
\newblock \href {https://proceedings.mlr.press/v270/dai25a.html} {Automated creation of digital cousins for robust policy learning}.
\newblock In \emph{Conference on Robot Learning, 6-9 November 2024, Munich, Germany}, volume 270 of \emph{Proceedings of Machine Learning Research}, pages 4912--4943. {PMLR}.

\bibitem[{Duisterhof et~al.(2026)Duisterhof, Zhang, Hung, Wen, Birchfield, Li, Ramanan, and Ichnowski}]{duisterhof2026pointzero}
Bardienus~P Duisterhof, Kaifeng Zhang, Adam Hung, Bowen Wen, Stan Birchfield, Yunzhu Li, Deva Ramanan, and Jeffrey Ichnowski. 2026.
\newblock {PointZero}: {3D} point track completion for learning transferable {3D} dynamics.
\newblock \emph{arXiv preprint arXiv:2609.19142}.

\bibitem[{Gervet et~al.(2023)Gervet, Xian, Gkanatsios, and Fragkiadaki}]{gervet2023act3d3dfeaturefield}
Th{\'{e}}ophile Gervet, Zhou Xian, Nikolaos Gkanatsios, and Katerina Fragkiadaki. 2023.
\newblock \href {https://proceedings.mlr.press/v229/gervet23a.html} {Act3d: 3d feature field transformers for multi-task robotic manipulation}.
\newblock In \emph{Conference on Robot Learning, CoRL 2023, 6-9 November 2023, Atlanta, GA, {USA}}, volume 229 of \emph{Proceedings of Machine Learning Research}, pages 3949--3965. {PMLR}.

\bibitem[{Heo et~al.(2024)Heo, Park, Han, and Yun}]{rope}
Byeongho Heo, Song Park, Dongyoon Han, and Sangdoo Yun. 2024.
\newblock \href {https://doi.org/10.1007/978-3-031-72684-2_17} {Rotary position embedding for vision transformer}.
\newblock In \emph{European Conference on Computer Vision (ECCV)}, pages 289--305, Cham. Springer Nature Switzerland.

\bibitem[{Hu et~al.(2026)Hu, Wu, Shi, Zhang, Yang, Zhang, Zhang, Xu, Peng, Li, and Zhang}]{hu2026tracewamlearningact3d}
Xu~Hu, Changjie Wu, Linzhe Shi, Hang Zhang, Chen Yang, Lingjun Zhang, Guowen Zhang, Mu~Xu, Junran Peng, Qing Li, and Zhaoxiang Zhang. 2026.
\newblock \href {https://arxiv.org/abs/2610.36840} {{Trace-WAM}: Learning to act via {3D} motion prediction for world action models}.
\newblock \emph{Preprint}, arXiv:2610.36840.

\bibitem[{Huang et~al.(2026{\natexlab{a}})Huang, Chao, Mousavian, Liu, Fox, Mo, and Fei{-}Fei}]{huang2026pointworld}
Wenlong Huang, Yu{-}Wei Chao, Arsalan Mousavian, Ming{-}Yu Liu, Dieter Fox, Kaichun Mo, and Li~Fei{-}Fei. 2026{\natexlab{a}}.
\newblock \href {https://openaccess.thecvf.com/content/CVPR2026/html/Huang\_PointWorld\_Scaling\_3D\_World\_Models\_for\_In-The-Wild\_Robotic\_Manipulation\_CVPR\_2026\_paper.html} {Pointworld: Scaling 3d world models for in-the-wild robotic manipulation}.
\newblock In \emph{{IEEE/CVF} Conference on Computer Vision and Pattern Recognition, {CVPR} 2026, Denver, CO, USA, June 3-7, 2026}, pages 20765--20779. Computer Vision Foundation.

\bibitem[{Huang et~al.(2026{\natexlab{b}})Huang, Li, Zhang, Zhang, Chen, Wang, Liu, Tomizuka, and Ji}]{huang20263dway}
Ziqin Huang, Yingyue Li, Chenyangguang Zhang, Ruida Zhang, Yuxin Chen, Gu~Wang, Xingyu Liu, Masayoshi Tomizuka, and Xiangyang Ji. 2026{\natexlab{b}}.
\newblock \href {https://doi.org/10.1007/978-3-032-37152-2_2} {{3DWay}: Generalizing robot manipulation via {3D} consistent waypoints}.
\newblock In \emph{European Conference on Computer Vision (ECCV)}, pages 20--38. Springer, Springer Nature Switzerland.

\bibitem[{Intelligence et~al.(2025)Intelligence, Black, Brown, Darpinian, Dhabalia, Driess, Esmail, Equi, Finn, Fusai, Galliker, Ghosh, Groom, Hausman, Ichter, Jakubczak, Jones, Ke, LeBlanc, Levine, Li-Bell, Mothukuri, Nair, Pertsch, Ren, Shi, Smith, Springenberg, Stachowicz, Tanner, Vuong, Walke, Walling, Wang, Yu, and Zhilinsky}]{intelligence2025pi05}
Physical Intelligence, Kevin Black, Noah Brown, James Darpinian, Karan Dhabalia, Danny Driess, Adnan Esmail, Michael Equi, Chelsea Finn, Niccolo Fusai, Manuel~Y. Galliker, Dibya Ghosh, Lachy Groom, Karol Hausman, Brian Ichter, Szymon Jakubczak, Tim Jones, Liyiming Ke, Devin LeBlanc, and 17 others. 2025.
\newblock \href {https://arxiv.org/abs/2504.16054} {$\pi_{0.5}$: a vision-language-action model with open-world generalization}.
\newblock \emph{Preprint}, arXiv:2504.16054.

\bibitem[{Ji et~al.(2025)Ji, Polavaram, Chen, Bajamahal, Ma, Adebola, Xu, and Goldberg}]{ji2025oxeauge}
Guanhua Ji, Harsha Polavaram, Lawrence~Yunliang Chen, Sandeep Bajamahal, Zehan Ma, Simeon Adebola, Chenfeng Xu, and Ken Goldberg. 2025.
\newblock \href {https://arxiv.org/abs/2512.13100} {{OXE-AugE}: A large-scale robot augmentation of {OXE} for scaling cross-embodiment policy learning}.
\newblock \emph{Preprint}, arXiv:2512.13100.

\bibitem[{Jiang et~al.(2026)Jiang, Wang, Wang, Cao, Wu, Luo, Hu, Gao, and Guo}]{jiang2026omnix}
Yanqin Jiang, Tengfei Wang, Zhengwei Wang, Chenjie Cao, Junta Wu, Wenhan Luo, Weiming Hu, Jin Gao, and Chunchao Guo. 2026.
\newblock \href {https://doi.org/10.1007/978-3-032-37235-2_19} {{OmniX}: Any-view and any-time {4D} reconstruction via feed-forward trajectory fields}.
\newblock In \emph{European Conference on Computer Vision (ECCV)}, pages 313--331. Springer, Springer Nature Switzerland.

\bibitem[{Karaev et~al.(2025)Karaev, Makarov, Wang, Neverova, Vedaldi, and Rupprecht}]{cotracker}
Nikita Karaev, Yuri Makarov, Jianyuan Wang, Natalia Neverova, Andrea Vedaldi, and Christian Rupprecht. 2025.
\newblock \href {https://doi.org/10.1109/ICCV51701.2025.00568} {{CoTracker3}: Simpler and better point tracking by pseudo-labeling real videos}.
\newblock In \emph{2025 IEEE/CVF International Conference on Computer Vision (ICCV)}, pages 1--10. IEEE.

\bibitem[{Karhade et~al.(2026)Karhade, Keetha, Zhang, Gupta, Sharma, Scherer, and Ramanan}]{karhade2026any4d}
Jay Karhade, Nikhil~Varma Keetha, Yuchen Zhang, Tanisha Gupta, Akash Sharma, Sebastian~A. Scherer, and Deva Ramanan. 2026.
\newblock \href {https://openaccess.thecvf.com/content/CVPR2026/html/Karhade\_Any4D\_Unified\_Feed-Forward\_Metric\_4D\_Reconstruction\_CVPR\_2026\_paper.html} {Any4d: Unified feed-forward metric 4d reconstruction}.
\newblock In \emph{{IEEE/CVF} Conference on Computer Vision and Pattern Recognition, {CVPR} 2026, Denver, CO, USA, June 3-7, 2026}, pages 14578--14589. Computer Vision Foundation.

\bibitem[{Ke et~al.(2024)Ke, Gkanatsios, and Fragkiadaki}]{ke20243ddiffuseractorpolicy}
Tsung{-}Wei Ke, Nikolaos Gkanatsios, and Katerina Fragkiadaki. 2024.
\newblock \href {https://proceedings.mlr.press/v270/ke25a.html} {3d diffuser actor: Policy diffusion with 3d scene representations}.
\newblock In \emph{Conference on Robot Learning, 6-9 November 2024, Munich, Germany}, volume 270 of \emph{Proceedings of Machine Learning Research}, pages 1949--1974. {PMLR}.

\bibitem[{Keetha et~al.(2026)Keetha, M{\"{u}}ller, Sch{\"{o}}nberger, Porzi, Zhang, Fischer, Knapitsch, Zauss, Weber, Antunes, Luiten, Lopez{-}Antequera, Bul{\`{o}}, Richardt, Ramanan, Scherer, and Kontschieder}]{keetha2026mapanything}
Nikhil~Varma Keetha, Norman M{\"{u}}ller, Johannes~L. Sch{\"{o}}nberger, Lorenzo Porzi, Yuchen Zhang, Tobias Fischer, Arno Knapitsch, Duncan Zauss, Ethan Weber, Nelson Antunes, Jonathon Luiten, Manuel Lopez{-}Antequera, Samuel~Rota Bul{\`{o}}, Christian Richardt, Deva Ramanan, Sebastian~A. Scherer, and Peter Kontschieder. 2026.
\newblock \href {https://doi.org/10.1109/3DV69130.2026.00054} {Mapanything: Universal feed-forward metric 3d reconstruction; map-anything.github.io}.
\newblock In \emph{International Conference on 3D Visio, 3DV 2026, Vancouver, BC, Canada, March 20-23, 2026}, pages 499--509. {IEEE}.

\bibitem[{Khazatsky et~al.(2024)Khazatsky, Pertsch, Nair, Balakrishna, Dasari, Karamcheti, Nasiriany, Srirama, Chen, Ellis, Fagan, Hejna, Itkina, Lepert, Ma, Miller, Wu, Belkhale, Dass, Ha, Jain, Lee, Lee, Memmel, Park, Radosavovic, Wang, Zhan, Black, Chi, Hatch, Lin, Lu, Mercat, Rehman, Sanketi, Sharma, Simpson, Vuong, Walke, Wulfe, Xiao, Yang, Yavary, Zhao, Agia, Baijal, Castro, Chen, Chen, Chung, Drake, Foster, Gao, Herrera, Heo, Hsu, Hu, Jackson, Le, Li, Lin, Ma, Maddukuri, Mirchandani, Morton, Nguyen, O'Neill, Scalise, Seale, Son, Tian, Tran, Wang, Wu, Xie, Yang, Yin, Zhang, Bastani, Berseth, Bohg, Goldberg, Gupta, Gupta, Jayaraman, Lim, Malik, Mart{\'{\i}}n{-}Mart{\'{\i}}n, Ramamoorthy, Sadigh, Song, Wu, Yip, Zhu, Kollar, Levine, and Finn}]{khazatsky2024droid}
Alexander Khazatsky, Karl Pertsch, Suraj Nair, Ashwin Balakrishna, Sudeep Dasari, Siddharth Karamcheti, Soroush Nasiriany, Mohan~Kumar Srirama, Lawrence~Yunliang Chen, Kirsty Ellis, Peter~David Fagan, Joey Hejna, Masha Itkina, Marion Lepert, Yecheng~Jason Ma, Patrick~Tree Miller, Jimmy Wu, Suneel Belkhale, Shivin Dass, and 79 others. 2024.
\newblock \href {https://doi.org/10.15607/RSS.2024.XX.120} {{DROID:} {A} large-scale in-the-wild robot manipulation dataset}.
\newblock In \emph{Robotics: Science and Systems XX, Delft, The Netherlands, July 15-19, 2024}.

\bibitem[{Lee et~al.(2026)Lee, Jung, Chun, Lee, Cai, Huang, Talreja, Dao, Liang, Huang, and Huang}]{lee2026tracegen}
Seungjae Lee, Yoonkyo Jung, Inkook Chun, Yao{-}Chih Lee, Zikui Cai, Hongjia Huang, Aayush Talreja, Tan~Dat Dao, Yongyuan Liang, Jia{-}Bin Huang, and Furong Huang. 2026.
\newblock \href {https://openaccess.thecvf.com/content/CVPR2026/html/Lee\_TraceGen\_World\_Modeling\_in\_3D\_Trace\_Space\_Enables\_Learning\_from\_CVPR\_2026\_paper.html} {Tracegen: World modeling in 3d trace space enables learning from cross-embodiment videos}.
\newblock In \emph{{IEEE/CVF} Conference on Computer Vision and Pattern Recognition, {CVPR} 2026, Denver, CO, USA, June 3-7, 2026}, pages 20721--20731. Computer Vision Foundation.

\bibitem[{Lin et~al.(2026)Lin, Chen, Liew, Chen, Li, Zhao, Peng, Guo, Zhou, Shi, Feng, and Kang}]{lin2025depth}
Haotong Lin, Sili Chen, Jun~Hao Liew, Donny~Y. Chen, Zhenyu Li, Yang Zhao, Sida Peng, Hengkai Guo, Xiaowei Zhou, Guang Shi, Jiashi Feng, and Bingyi Kang. 2026.
\newblock \href {https://proceedings.iclr.cc/paper\_files/paper/2026/hash/e4cd50120b6d7e8daff1749d6bbaa889-Abstract-Conference.html} {Depth anything 3: Recovering the visual space from any views}.
\newblock In \emph{The Fourteenth International Conference on Learning Representations, {ICLR} 2026, Rio de Janeiro, Brazil, April 23-27, 2026}. proceedings.iclr.cc.

\bibitem[{Lipman et~al.(2023)Lipman, Chen, Ben{-}Hamu, Nickel, and Le}]{lipman2023flow}
Yaron Lipman, Ricky T.~Q. Chen, Heli Ben{-}Hamu, Maximilian Nickel, and Matthew Le. 2023.
\newblock \href {https://openreview.net/forum?id=PqvMRDCJT9t} {Flow matching for generative modeling}.
\newblock In \emph{The Eleventh International Conference on Learning Representations, {ICLR} 2023, Kigali, Rwanda, May 1-5, 2023}. OpenReview.net.

\bibitem[{Liu et~al.(2025)Liu, Wu, Li, Tan, Chen, Wang, Xu, Su, and Zhu}]{liu2025rdt}
Songming Liu, Lingxuan Wu, Bangguo Li, Hengkai Tan, Huayu Chen, Zhengyi Wang, Ke~Xu, Hang Su, and Jun Zhu. 2025.
\newblock \href {https://openreview.net/forum?id=yAzN4tz7oI} {{RDT-1B:} a diffusion foundation model for bimanual manipulation}.
\newblock In \emph{The Thirteenth International Conference on Learning Representations, {ICLR} 2025, Singapore, April 24-28, 2025}. OpenReview.net.

\bibitem[{Liu et~al.(2026)Liu, Dong, Ye, Yuan, Jiang, Yang, Cao, Liu, Sun, Guo, Liu, Ke, Pan, Wu, Cheng, Ren, Zhang, Cui, Zhao, Zhang, Xu, Yang, Zhang, Niu, Zhu, Zhang, and Zhao}]{liu2026GalaxeaVLA}
Yicheng Liu, Zibin Dong, Baijun Ye, Tianyuan Yuan, Tao Jiang, Anqi Yang, Shicheng Cao, Haonan Liu, Yue Sun, Zihan Guo, Xiao Liu, Dong Ke, Changxun Pan, Chenru Wu, Tailai Cheng, Xiaoshu Ren, Xinlei Zhang, Jianning Cui, Zijie Zhao, and 8 others. 2026.
\newblock \href {https://arxiv.org/abs/2608.11739} {G0.5: One autoregressive stream for robot reasoning and action}.
\newblock \emph{Preprint}, arXiv:2608.11739.

\bibitem[{Lu et~al.(2026)Lu, Xu, Hu, Zhu, Zhao, Yeung, Shan, and Liu}]{lu2026track4world}
Jiahao Lu, Jiayi Xu, Wenbo Hu, Ruijie Zhu, Chengfeng Zhao, Sai{-}Kit Yeung, Ying Shan, and Yuan Liu. 2026.
\newblock \href {https://doi.org/10.1007/978-3-032-37023-5\_11} {Track4world: Feedforward world-centric dense 3d tracking of all pixels}.
\newblock In \emph{Computer Vision - {ECCV} 2026 - 19th European Conference, Malm{\"{o}}, Sweden, September 8-12, 2026, Proceedings, Part {LXIX}}, volume 17069 of \emph{Lecture Notes in Computer Science}, pages 184--204. Springer.

\bibitem[{Luo et~al.(2026)Luo, Zhou, Lan, Pan, and Loy}]{luo2026fourrc}
Yihang Luo, Shangchen Zhou, Yushi Lan, Xingang Pan, and Chen~Change Loy. 2026.
\newblock \href {https://arxiv.org/abs/2602.10094} {{4RC}: {4D} reconstruction via conditional querying anytime and anywhere}.
\newblock \emph{arXiv preprint arXiv:2602.10094}.

\bibitem[{Mandlekar et~al.(2023)Mandlekar, Nasiriany, Wen, Akinola, Narang, Fan, Zhu, and Fox}]{mandlekar2023mimicgen}
Ajay Mandlekar, Soroush Nasiriany, Bowen Wen, Iretiayo Akinola, Yashraj Narang, Linxi Fan, Yuke Zhu, and Dieter Fox. 2023.
\newblock \href {https://proceedings.mlr.press/v229/mandlekar23a.html} {Mimicgen: {A} data generation system for scalable robot learning using human demonstrations}.
\newblock In \emph{Conference on Robot Learning, CoRL 2023, 6-9 November 2023, Atlanta, GA, {USA}}, volume 229 of \emph{Proceedings of Machine Learning Research}, pages 1820--1864. {PMLR}.

\bibitem[{Mousavian et~al.(2019)Mousavian, Eppner, and Fox}]{mousavian2019graspnet}
Arsalan Mousavian, Clemens Eppner, and Dieter Fox. 2019.
\newblock \href {https://doi.org/10.1109/ICCV.2019.00299} {6-dof graspnet: Variational grasp generation for object manipulation}.
\newblock In \emph{2019 {IEEE/CVF} International Conference on Computer Vision, {ICCV} 2019, Seoul, Korea (South), October 27 - November 2, 2019}, pages 2901--2910. {IEEE}.

\bibitem[{{Open X-Embodiment Collaboration} et~al.(2023){Open X-Embodiment Collaboration}, O'Neill, Rehman, Gupta, Maddukuri, Gupta, Padalkar, Lee, Pooley, Gupta, Mandlekar, Jain, Tung, Bewley, Herzog, Irpan, Khazatsky, Rai, Gupta, Wang, Kolobov, Singh, Garg, Kembhavi, Xie, Brohan, Raffin, Sharma, Yavary, Jain, Balakrishna, Wahid, Burgess-Limerick, Kim, Schölkopf, Wulfe, Ichter, Lu, Xu, Le, Finn, Wang, Xu, Chi, Huang, Chan, Agia, Pan, Fu, Devin, Xu, Morton, Driess, Chen, Pathak, Shah, Büchler, Jayaraman, Kalashnikov, Sadigh, Johns, Foster, Liu, Ceola, Xia, Zhao, Frujeri, Stulp, Zhou, Sukhatme, Salhotra, Yan, Feng, Schiavi, Berseth, Kahn, Yang, Wang, Su, Fang, Shi, Bao, Amor, Christensen, Furuta, Bharadhwaj, Walke, Fang, Ha, Mordatch, Radosavovic, Leal, Liang, Abou-Chakra, Kim, Drake, Peters, Schneider, Hsu, Vakil, Bohg, Bingham, Wu, Gao, Hu, Wu, Wu, Sun, Luo, Gu, Tan, Oh, Wu, Lu, Yang, Malik, Silvério, Hejna, Booher, Tompson, Yang, Salvador, Lim, Han, Wang, Rao, Pertsch, Hausman, Go, Gopalakrishnan, Goldberg, Byrne, Oslund, Kawaharazuka, Black, Lin, Zhang, Ehsani, Lekkala, Ellis, Rana, Srinivasan, Fang, Singh, Zeng, Hatch, Hsu, Itti, Chen, Pinto, Fei-Fei, Tan, Fan, Ott, Lee, Weihs, Chen, Lepert, Memmel, Tomizuka, Itkina, Castro, Spero, Du, Ahn, Yip, Zhang, Ding, Heo, Srirama, Sharma, Kim, Irshad, Kanazawa, Hansen, Heess, Joshi, Suenderhauf, Liu, Palo, Shafiullah, Mees, Kroemer, Bastani, Sanketi, Miller, Yin, Wohlhart, Xu, Fagan, Mitrano, Sermanet, Abbeel, Sundaresan, Chen, Vuong, Rafailov, Tian, Doshi, Mart{'i}n-Mart{'i}n, Baijal, Scalise, Hendrix, Lin, Qian, Zhang, Mendonca, Shah, Hoque, Julian, Bustamante, Kirmani, Levine, Lin, Moore, Bahl, Dass, Sonawani, Tulsiani, Song, Xu, Haldar, Karamcheti, Adebola, Guist, Nasiriany, Schaal, Welker, Tian, Ramamoorthy, Dasari, Belkhale, Park, Nair, Mirchandani, Osa, Gupta, Harada, Matsushima, Xiao, Kollar, Yu, Ding, Davchev, Zhao, Armstrong, Darrell, Chung, Jain, Kumar, Vanhoucke, Guizilini, Zhan, Zhou, Burgard, Chen, Chen, Wang, Zhu, Geng, Liu, Liangwei, Li, Pang, Lu, Ma, Kim, Chebotar, Zhou, Zhu, Wu, Xu, Wang, Bisk, Dou, Cho, Lee, Cui, Cao, Wu, Tang, Zhu, Zhang, Jiang, Li, Li, Iwasawa, Matsuo, Ma, Xu, Cui, Zhang, Fu, and Lin}]{open_x_embodiment_rt_x_2023}
{Open X-Embodiment Collaboration}, Abby O'Neill, Abdul Rehman, Abhinav Gupta, Abhiram Maddukuri, Abhishek Gupta, Abhishek Padalkar, Abraham Lee, Acorn Pooley, Agrim Gupta, Ajay Mandlekar, Ajinkya Jain, Albert Tung, Alex Bewley, Alex Herzog, Alex Irpan, Alexander Khazatsky, Anant Rai, Anchit Gupta, and 275 others. 2023.
\newblock Open {X-E}mbodiment: Robotic learning datasets and {RT-X} models.
\newblock \url{https://arxiv.org/abs/2310.08864}.

\bibitem[{Oquab et~al.(2024)Oquab, Darcet, Moutakanni, Vo, Szafraniec, Khalidov, Fernandez, Haziza, Massa, El{-}Nouby, Assran, Ballas, Galuba, Howes, Huang, Li, Misra, Rabbat, Sharma, Synnaeve, Xu, J{\'{e}}gou, Mairal, Labatut, Joulin, and Bojanowski}]{oquab2023dinov2}
Maxime Oquab, Timoth{\'{e}}e Darcet, Th{\'{e}}o Moutakanni, Huy~V. Vo, Marc Szafraniec, Vasil Khalidov, Pierre Fernandez, Daniel Haziza, Francisco Massa, Alaaeldin El{-}Nouby, Mido Assran, Nicolas Ballas, Wojciech Galuba, Russell Howes, Po{-}Yao Huang, Shang{-}Wen Li, Ishan Misra, Michael Rabbat, Vasu Sharma, and 7 others. 2024.
\newblock \href {https://arxiv.org/abs/2304.07193} {{DINOv2}: Learning robust visual features without supervision}.
\newblock \emph{Transactions on Machine Learning Research (TMLR)}.

\bibitem[{Pan et~al.(2026)Pan, Peng, Zhang, Yan, Dai, Huo, Wang, Qi, Wang, Cheng et~al.}]{pan2026world}
Yuhao Pan, Haosong Peng, Zhengshen Zhang, Zhengyang Yan, Yalun Dai, Fushuo Huo, Chujie Wang, Tianyu Qi, Xiucheng Wang, Nan Cheng, and 1 others. 2026.
\newblock World-to-wrist: Task-conditioned future wrist modeling for fine-grained robot manipulation.
\newblock \emph{arXiv preprint arXiv:2608.05369}.

\bibitem[{Peng et~al.(2026)Peng, Li, Dai, Lan, Luo, Qi, Zhang, Zhan, Zhang, Xu, and Liu}]{omnivggt}
Haosong Peng, Hao Li, Yalun Dai, Yushi Lan, Yihang Luo, Tianyu Qi, Zhengshen Zhang, Yufeng Zhan, Junfei Zhang, Wenchao Xu, and Ziwei Liu. 2026.
\newblock \href {https://openaccess.thecvf.com/content/CVPR2026/html/Peng\_OmniVGGT\_Omni-Modality\_Driven\_Visual\_Geometry\_Grounded\_Transformer\_CVPR\_2026\_paper.html} {Omnivggt: Omni-modality driven visual geometry grounded transformer}.
\newblock In \emph{{IEEE/CVF} Conference on Computer Vision and Pattern Recognition, {CVPR} 2026, Denver, CO, USA, June 3-7, 2026}, pages 36485--36497. Computer Vision Foundation.

\bibitem[{Perez et~al.(2018)Perez, Strub, de~Vries, Dumoulin, and Courville}]{perez2018film}
Ethan Perez, Florian Strub, Harm de~Vries, Vincent Dumoulin, and Aaron~C. Courville. 2018.
\newblock \href {https://doi.org/10.1609/AAAI.V32I1.11671} {Film: Visual reasoning with a general conditioning layer}.
\newblock In \emph{Proceedings of the Thirty-Second {AAAI} Conference on Artificial Intelligence, (AAAI-18), the 30th innovative Applications of Artificial Intelligence (IAAI-18), and the 8th {AAAI} Symposium on Educational Advances in Artificial Intelligence (EAAI-18), New Orleans, Louisiana, USA, February 2-7, 2018}, pages 3942--3951. {AAAI} Press.

\bibitem[{Raffel et~al.(2020)Raffel, Shazeer, Roberts, Lee, Narang, Matena, Zhou, Li, and Liu}]{raffel2020exploring}
Colin Raffel, Noam Shazeer, Adam Roberts, Katherine Lee, Sharan Narang, Michael Matena, Yanqi Zhou, Wei Li, and Peter~J. Liu. 2020.
\newblock \href {https://jmlr.org/papers/v21/20-074.html} {Exploring the limits of transfer learning with a unified text-to-text transformer}.
\newblock \emph{J. Mach. Learn. Res.}, 21:140:1--140:67.

\bibitem[{Su et~al.(2026)Su, Chen, Shi, Liu, Zhang, Huang, Zhong, Zhu, Liu, and Liu}]{worldguide}
Yue Su, Sijin Chen, Haixin Shi, Mingyu Liu, Zhengshen Zhang, Ningyuan Huang, Weiheng Zhong, Zhengbang Zhu, Yuxiao Liu, and Xihui Liu. 2026.
\newblock \href {https://doi.org/10.48550/ARXIV.2602.22010} {World guidance: World modeling in condition space for action generation}.
\newblock \emph{CoRR}, abs/2602.22010.

\bibitem[{Sucar et~al.(2026)Sucar, Insafutdinov, Lai, and Vedaldi}]{sucar2026vdpm4dvideoreconstruction}
Edgar Sucar, Eldar Insafutdinov, Zihang Lai, and Andrea Vedaldi. 2026.
\newblock \href {https://openaccess.thecvf.com/content/CVPR2026/html/Sucar\_V-DPM\_4D\_Video\_Reconstruction\_with\_Dynamic\_Point\_Maps\_CVPR\_2026\_paper.html} {{V-DPM:} 4d video reconstruction with dynamic point maps}.
\newblock In \emph{{IEEE/CVF} Conference on Computer Vision and Pattern Recognition, {CVPR} 2026, Denver, CO, USA, June 3-7, 2026}, pages 14502--14511. Computer Vision Foundation.

\bibitem[{Sundaralingam et~al.(2026)Sundaralingam, Murali, and Birchfield}]{curobo_v2}
Balakumar Sundaralingam, Adithyavairavan Murali, and Stan Birchfield. 2026.
\newblock \href {https://arxiv.org/abs/2603.05493} {{cuRoboV2}: Dynamics-aware motion generation with depth-fused distance fields for {High-DoF} robots}.
\newblock \emph{Preprint}, arXiv:2603.05493.

\bibitem[{Sundermeyer et~al.(2021)Sundermeyer, Mousavian, Triebel, and Fox}]{sundermeyer2021contactgraspnet}
Martin Sundermeyer, Arsalan Mousavian, Rudolph Triebel, and Dieter Fox. 2021.
\newblock \href {https://doi.org/10.1109/ICRA48506.2021.9561877} {Contact-graspnet: Efficient 6-dof grasp generation in cluttered scenes}.
\newblock In \emph{{IEEE} International Conference on Robotics and Automation, {ICRA} 2021, Xi'an, China, May 30 - June 5, 2021}, pages 13438--13444. {IEEE}.

\bibitem[{Tan et~al.(2026)Tan, Sun, Qin, Adai, Fu, Zhou, Zhang, Xu, Zhu, Shen, and Xue}]{lingbot-depth2026}
Bin Tan, Changjiang Sun, Xiage Qin, Hanat Adai, Zelin Fu, Tianxiang Zhou, Han Zhang, Yinghao Xu, Xing Zhu, Yujun Shen, and Nan Xue. 2026.
\newblock Masked depth modeling for spatial perception.
\newblock \emph{arXiv preprint arXiv:2601.17895}.

\bibitem[{Wang et~al.(2026{\natexlab{a}})Wang, Wang, Lin, Tian, Zhang, Lu, and Yang}]{wang2026track4action}
Chenyi Wang, Xinkai Wang, Bokai Lin, Jialin Tian, Fucheng Zhang, Cewu Lu, and Lixin Yang. 2026{\natexlab{a}}.
\newblock {Track4Action}: Distilling world-centric 3d tracker into vision-language-action policies.
\newblock \emph{arXiv preprint arXiv:2608.03727}.

\bibitem[{Wang et~al.(2026{\natexlab{b}})Wang, Chen, Zhang, Karaev, Sch{\"o}nberger, Labatut, Bojanowski, Novotny, Vedaldi, and Rupprecht}]{wang2026vggtomega}
Jianyuan Wang, Minghao Chen, Shangzhan Zhang, Nikita Karaev, Johannes Sch{\"o}nberger, Patrick Labatut, Piotr Bojanowski, David Novotny, Andrea Vedaldi, and Christian Rupprecht. 2026{\natexlab{b}}.
\newblock Vggt-$\boldsymbol{\Omega}$.
\newblock \emph{arXiv preprint arXiv:2605.15195}.

\bibitem[{Wang et~al.(2026{\natexlab{c}})Wang, Song, Yu, Zhao, Wang, Chen, Wang, Wang, and Li}]{wang2026discreteforcing}
Jingbo Wang, Wenxuan Song, Wenhao Yu, Han Zhao, Xi~Wang, Jiayi Chen, Donglin Wang, Yan Wang, and Haoang Li. 2026{\natexlab{c}}.
\newblock \href {https://arxiv.org/abs/2609.39526} {Discrete forcing: Infusing discrete guidance into continuous denoising for few-step action experts}.
\newblock \emph{Preprint}, arXiv:2609.39526.

\bibitem[{Wang et~al.(2026{\natexlab{d}})Wang, Su, Zhou, Yue, Liu, Lu, Tian, Fang, Wu, Hu et~al.}]{wang2026unitas}
Ruixiang Wang, Yongyi Su, Wenlve Zhou, Bo~Yue, Hengyan Liu, Dekun Lu, Yuxin Tian, Yihan Fang, Zerui Wu, Xing Hu, and 1 others. 2026{\natexlab{d}}.
\newblock {UNITAS}: A {3D-Native} world action model for embodied manipulation.
\newblock \emph{arXiv preprint arXiv:2610.12099}.

\bibitem[{Wen et~al.(2024)Wen, Lin, So, Chen, Dou, Gao, and Abbeel}]{anypointtrajectory}
Chuan Wen, Xingyu Lin, John Ian~Reyes So, Kai Chen, Qi~Dou, Yang Gao, and Pieter Abbeel. 2024.
\newblock \href {https://doi.org/10.15607/RSS.2024.XX.092} {Any-point trajectory modeling for policy learning}.
\newblock In \emph{Robotics: Science and Systems XX, Delft, The Netherlands, July 15-19, 2024}.

\bibitem[{{Xiaomi Embodied Intelligence Team} et~al.(2026){Xiaomi Embodied Intelligence Team}, of~Macau, Xu, Li, Zhan, Duan, Wang, Luo, Jiang, Li, Du, Wang, Jiang, Liang, Gong, Pan, Zhao, Chen, You, Ma, Liu, Ye, and xin Yang}]{OnePolicy}
{Xiaomi Embodied Intelligence Team}, University of~Macau, Shaoqing Xu, Fang Li, Guozhi Zhan, Zhixiang Duan, Yuhan Wang, Yuechen Luo, Shengyin Jiang, Hanbing Li, Zhiying Du, Longlong Wang, Longmei Jiang, Weixiang Liang, Ying Gong, Yong Pan, Ziping Zhao, Zhiyuan Chen, Yangwei You, and 4 others. 2026.
\newblock \href {https://arxiv.org/abs/2608.26058} {One policy, many embodiments: Unified camera-centric action geometry pre-training for heterogeneous embodied manipulation}.
\newblock \emph{Preprint}, arXiv:2608.26058.

\bibitem[{Xie et~al.(2026)Xie, Meng, Diao, Cao, Du, Tu, Leng, Wang, Li, Bai, Wu, and Jiang}]{unifyrobotactionscamera}
Sicheng Xie, Lingchen Meng, Zijie Diao, Haidong Cao, Zhiying Du, Shuyuan Tu, Jiaqi Leng, Qiuyue Wang, Mingsheng Li, Shuai Bai, Zuxuan Wu, and Yu-Gang Jiang. 2026.
\newblock \href {https://arxiv.org/abs/2511.17001} {Unify robot actions in camera frame}.
\newblock \emph{Preprint}, arXiv:2511.17001.

\bibitem[{Yang et~al.(2026{\natexlab{a}})Yang, Tu, Yang, Mao, Dong, Chen, Peng, Xiong, Cao, Dai, Zhou, Mu, and Wang}]{yang2026eventvla}
Ganlin Yang, Zhangzheng Tu, Yuqiang Yang, Sitong Mao, Junyi Dong, Tianxing Chen, Jiaqi Peng, Jing Xiong, Jiafei Cao, Jifeng Dai, Wengang Zhou, Yao Mu, and Tai Wang. 2026{\natexlab{a}}.
\newblock \href {https://arxiv.org/abs/2606.20092} {{EventVLA}: Event-driven visual evidence memory for long-horizon vision-language-action policies}.
\newblock \emph{Preprint}, arXiv:2606.20092.

\bibitem[{Yang et~al.(2026{\natexlab{b}})Yang, Dagli, Zook, Hadfield, Goyal, Birchfield, Ramos, and Tremblay}]{yang2026robolab}
Jenai~Xuning Yang, Rishit Dagli, Alex Zook, Hugo Hadfield, Ankit Goyal, Stan Birchfield, Fabio Ramos, and Jonathan Tremblay. 2026{\natexlab{b}}.
\newblock Robolab: A high-fidelity simulation benchmark for analysis of task generalist policies.
\newblock \emph{arXiv preprint arXiv:2604.09860}.

\bibitem[{Ye et~al.(2026)Ye, Ge, Zheng, Gao, Yu, Kurian, Indupuru, Tan, Zhu, Xiang, Malik, Lee, Liang, Ranawaka, Gu, Xu, Wang, Hu, Narayan, Bjorck, Wang, Kim, Niu, Zheng, Xie, Wu, Wang, Julian, Xu, Du, Chebotar, Reed, Kautz, Zhu, Fan, and Jang}]{dreamzero}
Seonghyeon Ye, Yunhao Ge, Kaiyuan Zheng, Shenyuan Gao, Sihyun Yu, George Kurian, Suneel Indupuru, You~Liang Tan, Chuning Zhu, Jiannan Xiang, Ayaan Malik, Kyungmin Lee, William Liang, Nadun Ranawaka, Jiasheng Gu, Yinzhen Xu, Guanzhi Wang, Fengyuan Hu, Avnish Narayan, and 17 others. 2026.
\newblock \href {https://arxiv.org/abs/2602.15922} {World action models are zero-shot policies}.
\newblock \emph{Preprint}, arXiv:2602.15922.

\bibitem[{Yu et~al.(2023)Yu, Xiao, Tompson, Stone, Wang, Brohan, Singh, Tan, M, Peralta, Hausman, Ichter, and Xia}]{yu2023rosie}
Tianhe Yu, Ted Xiao, Jonathan Tompson, Austin Stone, Su~Wang, Anthony Brohan, Jaspiar Singh, Clayton Tan, Dee M, Jodilyn Peralta, Karol Hausman, Brian Ichter, and Fei Xia. 2023.
\newblock \href {https://doi.org/10.15607/RSS.2023.XIX.027} {Scaling robot learning with semantically imagined experience}.
\newblock In \emph{Robotics: Science and Systems XIX, Daegu, Republic of Korea, July 10-14, 2023}.

\bibitem[{Yuan et~al.(2026{\natexlab{a}})Yuan, Zhao, Guo, Sui, Yu, Liu, and Wang}]{yuan2026motionvlainjectinggeometricmotion}
Shanglin Yuan, Weiheng Zhao, Xianda Guo, Wei Sui, Li~Yu, Wenyu Liu, and Xinggang Wang. 2026{\natexlab{a}}.
\newblock \href {https://arxiv.org/abs/2606.08288} {{MotionVLA}: Injecting geometric motion into vision-language-action model}.
\newblock \emph{Preprint}, arXiv:2606.08288.

\bibitem[{Yuan et~al.(2026{\natexlab{b}})Yuan, Dong, Liu, and Zhao}]{fastwam}
Tianyuan Yuan, Zibin Dong, Yicheng Liu, and Hang Zhao. 2026{\natexlab{b}}.
\newblock \href {https://arxiv.org/abs/2603.16666} {{Fast-WAM}: Do world action models need test-time future imagination?}
\newblock \emph{Preprint}, arXiv:2603.16666.

\bibitem[{Ze et~al.(2024)Ze, Zhang, Zhang, Hu, Wang, and Xu}]{ze20243dp3}
Yanjie Ze, Gu~Zhang, Kangning Zhang, Chenyuan Hu, Muhan Wang, and Huazhe Xu. 2024.
\newblock \href {https://doi.org/10.15607/rss.2024.xx.067} {{3D} diffusion policy: Generalizable visuomotor policy learning via simple {3D} representations}.
\newblock In \emph{Robotics: Science and Systems XX}. Robotics: Science and Systems Foundation.

\bibitem[{Zhang et~al.(2026{\natexlab{a}})Zhang, Moing, Koppula, Rocco, Momeni, Xie, Sun, Sukthankar, Barral, Hadsell, Ghahramani, Zisserman, Zhang, and Sajjadi}]{zhang2026efficiently}
Chuhan Zhang, Guillaume~Le Moing, Skanda Koppula, Ignacio Rocco, Liliane Momeni, Junyu Xie, Shuyang Sun, Rahul Sukthankar, Jo{\"{e}}lle~K. Barral, Raia Hadsell, Zoubin Ghahramani, Andrew Zisserman, Junlin Zhang, and Mehdi S.~M. Sajjadi. 2026{\natexlab{a}}.
\newblock \href {https://openaccess.thecvf.com/content/CVPR2026/html/Zhang\_Efficiently\_Reconstructing\_Dynamic\_Scenes\_One\_D4RT\_at\_a\_Time\_CVPR\_2026\_paper.html} {Efficiently reconstructing dynamic scenes one {D4RT} at a time}.
\newblock In \emph{{IEEE/CVF} Conference on Computer Vision and Pattern Recognition, {CVPR} 2026, Denver, CO, USA, June 3-7, 2026}, pages 7382--7392. Computer Vision Foundation.

\bibitem[{Zhang et~al.(2025)Zhang, Zhang, Pan, Zheng, and Zhang}]{zhang2025generativevisualforesightmeets}
Chuye Zhang, Xiaoxiong Zhang, Wei Pan, Linfang Zheng, and Wei Zhang. 2025.
\newblock \href {https://arxiv.org/abs/2509.00361} {Generative visual foresight meets task-agnostic pose estimation in robotic table-top manipulation}.
\newblock \emph{Preprint}, arXiv:2509.00361.

\bibitem[{Zhang et~al.(2026{\natexlab{b}})Zhang, Zheng, Yang, Argus, Soraki, Han, Anderson, Li, Liu, Duan, Ren, Zhang, and Krishna}]{zhang2026molmomotionforecastingpointtrajectories}
Jianing Zhang, Chenhao Zheng, Yajun Yang, Max Argus, Rustin Soraki, Winson Han, Taira Anderson, Chun-Liang Li, Shuo Liu, Jiafei Duan, Zhongzheng Ren, Jieyu Zhang, and Ranjay Krishna. 2026{\natexlab{b}}.
\newblock \href {https://arxiv.org/abs/2606.18558} {{MolmoMotion}: Forecasting point trajectories in {3D} with language instruction}.
\newblock \emph{Preprint}, arXiv:2606.18558.

\bibitem[{Zhang et~al.(2026{\natexlab{c}})Zhang, Li, Dai, Zhu, Zhou, Liu, Wang, Tay, Chen, Liu, Liu, Li, and Zhou}]{Falcon}
Zhengshen Zhang, Hao Li, Yalun Dai, Zhengbang Zhu, Lei Zhou, Chenchen Liu, Dong Wang, Francis E.~H. Tay, Sijin Chen, Ziwei Liu, Yuxiao Liu, Xinghang Li, and Pan Zhou. 2026{\natexlab{c}}.
\newblock \href {https://proceedings.iclr.cc/paper\_files/paper/2026/hash/3cbe9fcdccb2399bcd6e6d01cbcae1fd-Abstract-Conference.html} {From spatial to actions: Grounding vision-language-action model in spatial foundation priors}.
\newblock In \emph{The Fourteenth International Conference on Learning Representations, {ICLR} 2026, Rio de Janeiro, Brazil, April 23-27, 2026}. proceedings.iclr.cc.

\bibitem[{Zhao et~al.(2026)Zhao, Zhang, Su, Wang, Li, Yang, Tay, Jr., and Zhu}]{sgwam}
Ruiteng Zhao, Zhengshen Zhang, Yue Su, Wenshuo Wang, Jiahui Li, Zhiyuan Yang, Francis E.~H. Tay, Marcelo H.~Ang Jr., and Haiyue Zhu. 2026.
\newblock \href {https://doi.org/10.48550/ARXIV.2608.01397} {{SG-WAM:} self-guided world modeling in geometry-aware policy space}.
\newblock \emph{CoRR}, abs/2608.01397.

\bibitem[{Zhao et~al.(2023)Zhao, Kumar, Levine, and Finn}]{zhao2023act}
Tony~Z. Zhao, Vikash Kumar, Sergey Levine, and Chelsea Finn. 2023.
\newblock \href {https://doi.org/10.15607/RSS.2023.XIX.016} {Learning fine-grained bimanual manipulation with low-cost hardware}.
\newblock In \emph{Robotics: Science and Systems XIX, Daegu, Republic of Korea, July 10-14, 2023}.

\bibitem[{Zhi et~al.(2025)Zhi, Chen, Zhou, Dong, Wu, Han, and Tan}]{zhi20253dflowaction}
Hongyan Zhi, Peihao Chen, Siyuan Zhou, Yubo Dong, Quanxi Wu, Lei Han, and Mingkui Tan. 2025.
\newblock 3dflowaction: Learning cross-embodiment manipulation from 3d flow world model.
\newblock \emph{arXiv preprint arXiv:2506.06199}.

\bibitem[{Zhou et~al.(2023)Zhou, Li, Chan, and Loy}]{zhou2023propainter}
Shangchen Zhou, Chongyi Li, Kelvin C.~K. Chan, and Chen~Change Loy. 2023.
\newblock \href {https://doi.org/10.1109/ICCV51070.2023.00961} {Propainter: Improving propagation and transformer for video inpainting}.
\newblock In \emph{{IEEE/CVF} International Conference on Computer Vision, {ICCV} 2023, Paris, France, October 1-6, 2023}, pages 10443--10452. {IEEE}.

\bibitem[{Zhou et~al.(2019)Zhou, Barnes, Lu, Yang, and Li}]{zhou2019continuity}
Yi~Zhou, Connelly Barnes, Jingwan Lu, Jimei Yang, and Hao Li. 2019.
\newblock \href {https://doi.org/10.1109/CVPR.2019.00589} {On the continuity of rotation representations in neural networks}.
\newblock In \emph{{IEEE} Conference on Computer Vision and Pattern Recognition, {CVPR} 2019, Long Beach, CA, USA, June 16-20, 2019}, pages 5745--5753. Computer Vision Foundation / {IEEE}.

\end{thebibliography}

\clearpage
\appendix
\section*{Appendix}

\etocdepthtag.toc{appendix}
\begingroup
  \etocsetnexttocdepth{subsection}
  \etocsettagdepth{main}{none}
  \etocsettagdepth{appendix}{subsection}
  \etocsettocstyle{\section*{Contents}}{\medskip}
  \hypersetup{linktoc=all}
  \tableofcontents
\endgroup

\clearpage
\section{Detailed Training Objectives}
\label{app:losses}

This section details the losses in Eqs.~\ref{eq:stage1} and~\ref{eq:stage2}, with an emphasis on TCP tracking supervision.
We use the same TCP state $\mathbf x_i^a=(\mathbf T_i^a,g_i^a)$ and pose $\mathbf T_i^a=(\mathbf R_i^a,\mathbf t_i^a)$ as in the main paper.
Here, $\mathbf R_i^a\in\mathrm{SO}(3)$ and $\mathbf t_i^a\in\mathbb R^3$ are the TCP orientation and position in frame $i$'s camera coordinate system, and $g_i^a\in[0,1]$ is gripper state. 
The index $a$ identifies an available arm, with $a\in\{l,r\}$ for bimanual manipulation. For supervised outputs, hats indicate predictions and unhatted quantities denote ground truth.

\subsection{Geometry Supervision}
\label{app:geometry_loss}

Given an RGB video $\mathbf I=\{I_i\}_{i=1}^{N}$, the geometry head predicts depth maps $\hat{\mathbf D}=\{\hat{\mathbf D}_i\}_{i=1}^{N}$ and camera-ray maps $\hat{\mathbf M}=\{\hat{\mathbf M}_i\}_{i=1}^{N}$,
where $\hat{\mathbf D}_i\in\mathbb R^{H\times W}$ and $\hat{\mathbf M}_i\in\mathbb R^{\frac H2\times\frac W2\times6}$, and $H, W$ are image height and width. 
Each ray contains a viewing direction and a camera origin expressed in the first camera frame. We use $\mathbf M_i$ for ray maps and reserve $\mathbf R_i^a$ for TCP rotations. We denote depth and ray confidence maps by $\hat{\mathbf Y}_d$ and $\hat{\mathbf Y}_r$, respectively.

Following 4RC~\cite{luo2026fourrc} and Depth Anything 3~\cite{lin2025depth}, the geometry objective combines depth regression, spatial depth gradients, and ray regression:
\begin{equation}
    \mathcal L_{\mathrm{geo}}
    =\mathcal L_{\mathrm{depth}}+\mathcal L_{\mathrm{grad}}
     +\mathcal L_{\mathrm{ray}}.
    \label{eq:app_geo_total}
\end{equation}
All three terms have unit weight. Depth and camera translations retain metric scale.

\noindent\textbf{Confidence-weighted Regression.}
For a prediction $\hat x\in\mathbb R^d$, target $x$, and positive confidence $\hat y$, define
\begin{equation}
    \begin{aligned}
        \ell_{\mathrm{reg}}(\hat x,x)
        &=\frac1d\|\hat x-x\|_1,\\
        \ell_{\mathrm{conf}}(\hat x,x,\hat y)
        &=\hat y\,\ell_{\mathrm{reg}}(\hat x,x)-\alpha\log\hat y,
    \end{aligned}
    \label{eq:app_geometry_regression}
\end{equation}
where $\alpha=0.2$, matching the confidence regularization used for TCP supervision below. 
We use $d=1$ for depth and $d=6$ for rays.
Let $p$ index a pixel together with its video frame, and let $\Omega_d$ and $\Omega_r$ denote the valid depth and ray-grid locations, respectively.

\noindent\textbf{Depth Loss.}
For predicted depths $\hat{\mathbf D}$, ground truth $\mathbf D$, and confidence $\hat{\mathbf Y}_d$, we average over valid depth pixels:
\begin{equation}
    \mathcal L_{\mathrm{depth}}
    =\frac1{|\Omega_d|}\sum_{p\in\Omega_d}
      \ell_{\mathrm{conf}}(\hat D_p,D_p,\hat Y_{d,p}).
    \label{eq:app_depth_reg}
\end{equation}

\noindent\textbf{Depth-gradient Loss.}
To preserve sharp edges and smooth planar regions, we penalize the discrepancy between predicted and ground-truth depth gradients:
\begin{equation}
    \mathcal L_{\mathrm{grad}}
    =\|\nabla_x\hat{\mathbf D}-\nabla_x\mathbf D\|_1
     +\|\nabla_y\hat{\mathbf D}-\nabla_y\mathbf D\|_1.
    \label{eq:app_depth_grad}
\end{equation}
Here, $\nabla_x$ and $\nabla_y$ are horizontal and vertical finite-difference operators. 
In practice, we evaluate this term over $J=4$ dyadic scales (stride $2^j$, $j=0,\ldots,J-1$) and average the results, using only differences whose endpoints have valid depth.

\noindent\textbf{Ray Loss.}
For the six-channel ray predictions $\hat{\mathbf M}$, ground truth $\mathbf M$, and confidence $\hat{\mathbf Y}_r$, we use
\begin{equation}
    \mathcal L_{\mathrm{ray}}
    =\frac1{|\Omega_r|}\sum_{p\in\Omega_r}
      \ell_{\mathrm{conf}}(\hat{\mathbf M}_p,\mathbf M_p,\hat Y_{r,p}).
    \label{eq:app_ray_loss}
\end{equation}
The factor $1/d$ in Eq.~\ref{eq:app_geometry_regression} averages the residual over the six ray channels.

\Needspace{24\baselineskip}
\subsection{TCP Tracking Supervision}
\label{app:tcp_loss}

\textwithfigure[.51]{%
TCP poses are supervised in each observed frame's own camera coordinate system. 
The arm index $a$ ranges over the available arms (e.g., left arm and right arm), so the same formulation applies to single-arm and bimanual tracking.

\noindent\textbf{Virtual-point Pose Supervision.}
Following point-based pose supervision~\cite{mousavian2019graspnet,sundermeyer2021contactgraspnet},
we supervise TCP translation and orientation by aligning rigidly transformed
reference points.
As shown in Fig.~\ref{fig:virtual}, a TCP center alone provides no orientation constraint. 
We therefore expand each TCP into a rigid set of seven virtual points: its center and six points displaced along the positive and negative local coordinate axes.
Their offsets are
\begin{equation}
    \{\boldsymbol\delta_j\}_{j=1}^{7}
    =\{\mathbf0,\ \pm r\mathbf e_k:\ k=1,2,3\},
    \label{eq:app_virtual_offsets}
\end{equation}
where $\mathbf e_k$ are Cartesian basis vectors and $r=0.03\,\mathrm m$.
}{%
\begin{panelfigure}
\centering
\includegraphics[width=\linewidth]{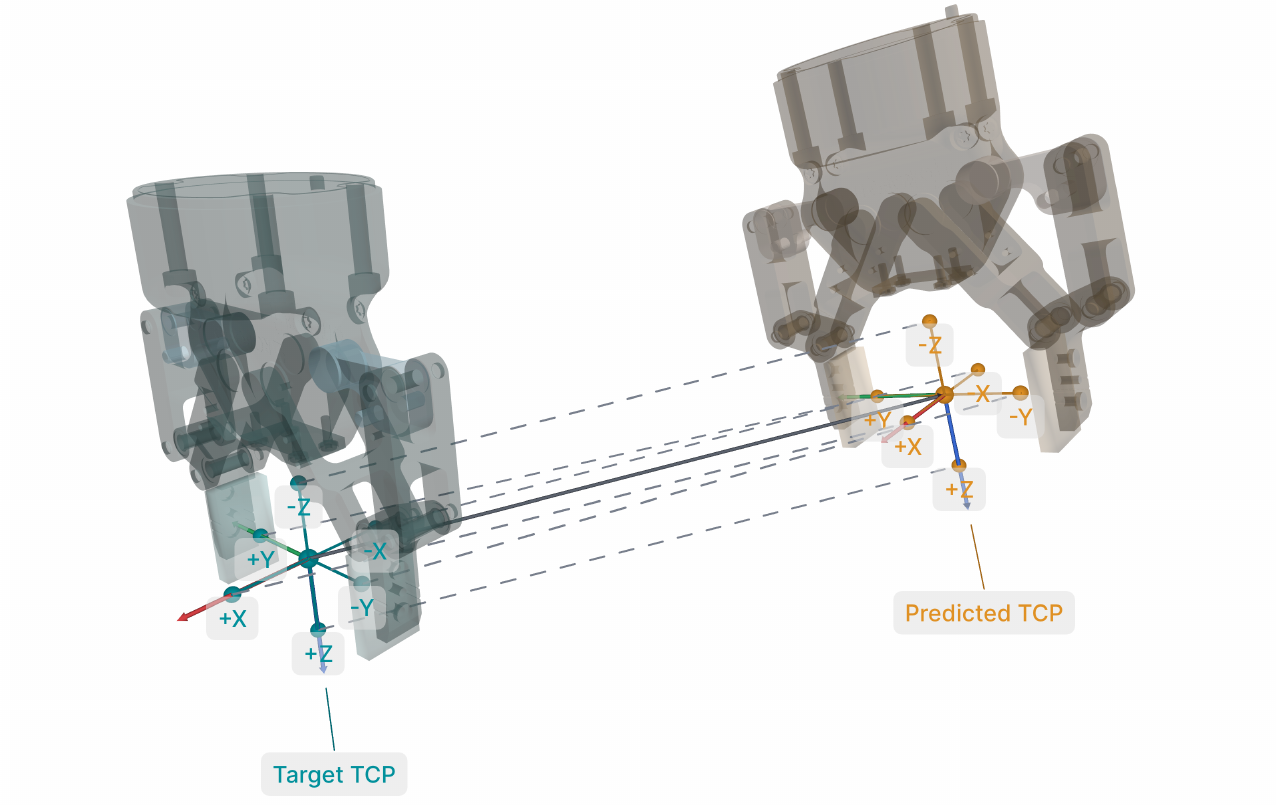}
\caption{\textbf{TCP tracking supervision illustration.} The green Robotiq gripper on the left indicates the ground-truth TCP, while the orange Robotiq gripper on the right indicates the predicted TCP.
We expand each TCP into a rigid set of seven virtual points, thereby augmenting TCP pose \SE supervision without additional annotations. }
\label{fig:virtual}
\end{panelfigure}
}
The target and predicted points are obtained by applying the corresponding
TCP poses:
\begin{equation}
    \begin{aligned}
        \mathbf y_{i,j}^a
        &=\mathbf t_i^a+\mathbf R_i^a\boldsymbol\delta_j,\\
        \hat{\mathbf y}_{i,j}^a
        &=\hat{\mathbf t}_i^a+\hat{\mathbf R}_i^a\boldsymbol\delta_j.
    \end{aligned}
    \label{eq:app_virtual_points}
\end{equation}
These targets require no additional annotations. 
Their spatial extent makes point alignment sensitive to both TCP translation and orientation.
We measure the normalized point error by
\begin{equation}
    e_{i,a}^{\mathrm{point}}
    =\frac17\sum_{j=1}^{7}
      \bar\rho\!\left(
        \frac{\hat{\mathbf y}_{i,j}^a-\mathbf y_{i,j}^a}{s_p}
      \right),
    \label{eq:app_rigid_error}
\end{equation}
where $s_p=0.1\,\mathrm m$ and $\bar\rho$ averages the unit-transition
Smooth L1 penalty over the three coordinates. For a scalar residual $z$,
this penalty is $z^2/2$ when $|z|<1$ and $|z|-1/2$ otherwise.
We additionally supervise rotational alignment using the geodesic angle
\begin{equation}
    d_R(\hat{\mathbf R},\mathbf R)
    =\arccos\!\left(
      \frac{\operatorname{tr}(\hat{\mathbf R}^{\top}\mathbf R)-1}{2}
      \right).
    \label{eq:app_rotation_distance}
\end{equation}
With predicted pose confidence $c_i^a>0$, the pose objective is
\begin{equation}
    \begin{aligned}
        e_{i,a}^{\mathrm{pose}}
        &=e_{i,a}^{\mathrm{point}}
          +\lambda_R\frac{d_R(\hat{\mathbf R}_i^a,\mathbf R_i^a)}{\pi},\\
        \mathcal L_{\mathrm{pose}}
        &=\left\langle
            c_i^a e_{i,a}^{\mathrm{pose}}-\alpha\log c_i^a
          \right\rangle_{\mathrm{valid}},
    \end{aligned}
    \label{eq:app_tcp_pose}
\end{equation}
where $\lambda_R=0.5$ and $\alpha=0.2$.
The average includes valid frame--arm pairs across the minibatch; an invalid reference query excludes that arm's trajectory.
The confidence regularizer discourages reducing confidence solely to suppress reconstruction errors.

\noindent\textbf{Temporal Velocity Consistency.}
We also supervise motion through finite differences of the virtual points.
Let $s_i$ be the timestamp of sampled frame $i$ and
$\Delta_i=s_i-s_{i-1}$. The target and predicted velocities are
\begin{equation}
    \begin{aligned}
        \mathbf v_{i,j}^a
        &=\frac{\mathbf y_{i,j}^a-\mathbf y_{i-1,j}^a}{\Delta_i},\\
        \hat{\mathbf v}_{i,j}^a
        &=\frac{\hat{\mathbf y}_{i,j}^a
                 -\hat{\mathbf y}_{i-1,j}^a}{\Delta_i}.
    \end{aligned}
    \label{eq:app_tcp_velocity}
\end{equation}
Using the actual, signed interval accommodates variable frame sampling
and reversed clips. We define
\begin{equation}
    \begin{aligned}
        e_{i,a}^{v}
        &=\frac17\sum_{j=1}^{7}
          \bar\rho\!\left(
            \frac{\hat{\mathbf v}_{i,j}^a-\mathbf v_{i,j}^a}{s_v}
          \right),\\
        \mathcal L_{\mathrm{temp}}
        &=\left\langle
            \tilde c_i^a e_{i,a}^{v}-\alpha\log\tilde c_i^a
          \right\rangle_{\mathrm{pair}},
    \end{aligned}
    \label{eq:app_tcp_temporal}
\end{equation}
where $s_v=1\,\mathrm{m/s}$ and
$\tilde c_i^a=\min(c_i^a,c_{i-1}^a)$.
The average includes valid adjacent frame pairs for each available arm.
Applying this loss to the extended points constrains temporal changes in both translation and orientation. 
These finite differences are computed in the supplied per-frame camera-coordinate labels.

\noindent\textbf{Gripper and Stage-wise TCP Objectives.}
We supervise predicted gripper state $\hat g_i^a$ against $g_i^a\in[0,1]$ with binary cross-entropy, averaged over the same valid frame--arm pairs. 
Continuous targets use the unweighted loss; binary targets use positive-class reweighting by the ratio of negative to positive labels when both classes are present. 
Denoting this loss by
$\mathcal L_{\mathrm{grip}}$, the two TCP objectives are
\begin{equation}
    \begin{aligned}
        \mathcal L_{\mathrm{tcp}}^{(1)}
        &=\mathcal L_{\mathrm{pose}}
          +0.2\mathcal L_{\mathrm{temp}}
          +0.2\mathcal L_{\mathrm{grip}},\\
        \mathcal L_{\mathrm{tcp}}^{(2)}
        &=\mathcal L_{\mathrm{pose}}+0.2\mathcal L_{\mathrm{grip}}.
    \end{aligned}
    \label{eq:app_tcp_total}
\end{equation}
Thus Stage~1 supervises pose, temporal motion, and gripper reconstruction, while Stage~2 retains pose and gripper reconstruction.

\subsection{Conditional Flow-matching Supervision}
\label{app:flow_loss}

\noindent\textbf{Action Representation and Normalization.}
Ground-truth future TCP poses are transformed into the current observation's
camera coordinate system before normalization. Each arm contributes a
10D action $(\mathbf t,\mathbf r,g)$: metric position $\mathbf t\in\mathbb R^3$,
the first two columns of the rotation matrix $\mathbf r\in\mathbb R^6$,
and gripper state $g\in[0,1]$. Thus the bimanual target trajectory
$\mathbf A$ has shape $h\times20$, where $h=32$ is the prediction horizon.

Position is normalized separately for each arm and coordinate using the
1st and 99th percentiles of valid training targets, pooled across future
steps and excluding padding. An affine transform maps this range
approximately to $[-1,1]$, without clipping outliers.
The 6D rotation remains unscaled, while gripper states are mapped linearly
from $[0,1]$ to $[-1,1]$.
At inference, the same statistics recover metric positions; rotation
columns are orthonormalized and gripper states are thresholded at zero.

\noindent\textbf{Flow-matching Objective.}
The condition $\mathcal O_t$ contains $L$ past and current RGB
frames, the current joint and gripper state $\mathbf s_t$, and wrist-camera
images $I_t^{w,a}$; $\ell$ is the language instruction. Gaussian noise
$\boldsymbol\epsilon\sim\mathcal N(\mathbf0,\mathbf I)$ and flow time
$\tau\in[0,1)$ define the interpolated input $\mathbf A_\tau$ and target
$\mathbf v^\star$ in Eq.~\ref{eq:flow_path}. We write
$\hat{\mathbf v}=\mathbf v_\phi(\mathbf A_\tau,\tau;\mathcal O_t,\ell)$.
The generation objective is
\begin{equation}
    \mathcal L_{\mathrm{fm}}
    =\mathbb E_{\mathbf A,\boldsymbol\epsilon,\tau}
      \left[
        \sum_{q\in\{\mathrm{pos},\mathrm{rot},\mathrm{grip}\}}
        \frac{\left\langle
          \|\hat{\mathbf v}_q-\mathbf v_q^\star\|_2^2
        \right\rangle_f}{d_q}
      \right],
    \label{eq:app_flow_total}
\end{equation}
where $d_{\mathrm{pos}}=3$, $d_{\mathrm{rot}}=6$, and
$d_{\mathrm{grip}}=1$ are the per-arm component dimensions.
For each trajectory, $\langle\cdot\rangle_f$ averages over valid future
steps and available arms; padded steps are excluded. 
The expectation also averages over trajectories and sampled noise and flow time.
All three component means have unit weight; dividing by $d_q$ prevents the
6D rotation from receiving a larger weight solely because of its dimensionality.

\section{Cross-Embodiment Augmentation}
\label{app:pipeline}

\subsection{Augmentation Pipeline}
\ours enables cross-embodiment augmentation by recovering task-space motion directly from RGB demonstrations. As illustrated in Fig.~\ref{fig:pipeline}, the pipeline comprises motion recovery, embodiment retargeting, and visual composition. Given a source video and initial TCP queries, \ours predicts TCP trajectories, gripper states, and camera geometry. The recovered poses are aligned in a common coordinate frame and transferred to the target robot through inverse kinematics using its kinematic model, producing joint configurations that follow the demonstrated task-space motion.

We segment the source robot using SAM3~\cite{carion2026sam} or, when available, its geometric model and recorded states. ProPainter~\cite{zhou2023propainter} completes the masked regions to reconstruct the background. We then render the target robot at the recovered configurations and composite it into the completed frames using the aligned camera geometry. The resulting images are paired with target configurations to form augmented demonstrations. 
Depth from \ours can optionally improve segmentation quality or distinguish foreground and background, but the pipeline does not depend on depth.

During Retarget, if IK fails at the target robot's initial base placement, the pipeline automatically searches nearby robot-base translations and retries IK over the full trajectory (auto\_base is enabled by default). The base orientation remains fixed during this search, and the target arm is accepted only when every frame passes IK. For further details on robot-base adjustment in cross-embodiment augmentation, refer to the OXE-AugE~\cite{ji2025oxeauge} paper and project page.

\subsection{Motion Recovery from RGB}
The key advantage of \ours is that motion information for retargeting can be inferred from video rather than supplied as recorded source-robot states. Table~\ref{tab:augmentation_requirements} compares the required resources and the sources of motion and camera information. Predicted TCP trajectories and gripper states provide the motion representation, while inverse kinematics recovers target joint configurations. Camera geometry can also be predicted when calibration is unavailable. With SAM3 segmentation, source URDFs and recorded joint configurations are unnecessary for robot removal. This extends augmentation to RGB demonstrations without accompanying robot-state trajectories, while retaining the target model for retargeting and rendering.

\begin{table}[!ht]
\centering
\small
\setlength{\tabcolsep}{4pt}
\renewcommand{\arraystretch}{1.12}
\caption{\textbf{Resources and motion signals for cross-embodiment augmentation.} $\checkmark$: supplied as input or recorded metadata (required); $\times$: not required; \textbf{Pred.}: inferred from RGB by \ours. Source URDFs are unnecessary with image-based segmentation.}
\label{tab:augmentation_requirements}
\begin{tabularx}{\linewidth}{@{}>{\raggedright\arraybackslash}p{0.38\linewidth}*{3}{>{\centering\arraybackslash}X}@{}}
\toprule
\textbf{Resource / signal} & \textbf{RoVi-Aug}\par{\small\cite{chen2024roviaug}} & \textbf{OXE-AugE}\par{\small\cite{ji2025oxeauge}} & \ours \\
\midrule
Source robot URDF & $\times$ & $\times$ & $\times$ \\
Target robot URDF & $\times$ & $\checkmark$ & $\checkmark$ \\
Recorded camera intrinsics & $\times$ & $\times$ & \textbf{Pred.} \\
Recorded camera extrinsics & $\times$ & $\times$ & \textbf{Pred.} \\
Source TCP trajectory & $\checkmark$ & $\checkmark$ & \textbf{Pred.} \\
Source joint configurations & $\checkmark$ & $\times$ & $\times$ \\
Source gripper state & $\checkmark$ & $\checkmark$ & \textbf{Pred.} \\
\bottomrule
\end{tabularx}
\end{table}

\section{Implementation Details}
\label{app:implementation}
We provide the architecture configuration, frame sampling strategy, preprocessing, and optimization details in this section, followed by the dataset splits and evaluation settings.

\subsection{Model Architecture and Hyperparameters}
\label{app:training}

\noindent\textbf{Dataset-specific Training.}
We train two independent Stage~1 models: a single-arm TCP tracker on DROID and a bimanual TCP tracker on RoboTwin.
Both models initialize their visual backbones and geometry modules from pretrained 4RC weights.
Stage~2 uses only RoboTwin data and jointly fine-tunes the TCP tracker and scene geometry modules while training the conditioning modules and TCP generator.
The camera head is excluded from optimization in both stages and is enabled for camera evaluation.
The pretrained wrist-image and language encoders remain frozen during Stage~2.
Table~\ref{tab:model_architecture} summarizes the model architecture used for the 48-task RoboTwin experiments.

\begin{table}[!ht]
  \centering
  \caption{\textbf{Model architecture for the \ours Stage 1\&2 experiments.}}
  \label{tab:model_architecture}
  \small
  \setlength{\tabcolsep}{6pt}
  \renewcommand{\arraystretch}{1.08}
  \begin{tabularx}{\linewidth}{@{}>{\raggedright\arraybackslash}p{0.24\linewidth}>{\raggedright\arraybackslash}p{0.30\linewidth}>{\raggedright\arraybackslash}X@{}}
    \toprule
    \textbf{Module} & \textbf{Architecture / setting} & \textbf{Specification} \\
    \midrule
     Shared visual and TCP modules & Visual backbone and geometry head & 4RC backbone and Dual-DPT; trainable \\
     & Sparse TCP modules & Query encoder, decoder, and TCP head; trainable \\
     & TCP query neighborhood & $3\times3$ patches per arm and frame \\
     & Geometry decoding & Enabled \\
     & Camera head & Excluded from optimization in both stages \\
    \midrule
    Conditioning encoders & Robot-state encoder & MLP: $14\rightarrow1024\rightarrow1280$ \\
     & Global visual token grid & $14\times19$ (266 tokens) \\
     & Wrist-image encoder & Shared DINOv2-Large; frozen \\
     & Language encoder & T5-base; frozen \\
     & Maximum language length & 128 tokens \\
     & Image and language condition width & 2048 \\
    \midrule
    Trajectory generator (DiT) & Transformer blocks & 12; alternating cross- and self-attention \\
     & Hidden width & 1280 \\
     & Attention heads & 32 (40 dimensions per head) \\
     & Feed-forward hidden width & 5120 (expansion ratio $4$) \\
     & Dropout & $0.2$ \\
     & Output feature width & 1024 \\
     & Action decoder & MLP: $1024\rightarrow1024\rightarrow20$ \\
    \midrule
    Input and output sequences & Observation history / prediction horizon & $L=4$ frames / $h=32$ steps \\
     & Local TCP history tokens & $4\times2\times9=72$ \\
     & DiT input sequence & 1 state $+$ 72 TCP $+$ 32 action tokens \\
     & Image context tokens & $266+2\times266=798$ \\
     & Predicted TCP trajectory & $32\times20$ \\
     & Flow-matching inference & 4 Euler steps \\
    \bottomrule
  \end{tabularx}
\end{table}

\noindent\textbf{Model Size.}
We report total registered model parameters, including frozen modules and inherited heads retained in the implementation.
The \ours Stage~1 trackers contain approximately 1.522B parameters.
The full tracker--generator model contains approximately 2.272B parameters, comprising the Stage~1 tracker, 336.462M parameters in the added action and conditioning modules, a frozen DINOv2-Large wrist-image encoder with 304.369M parameters, and a frozen T5-base text encoder with 109.629M parameters.
The DiT accounts for approximately 294M of the added parameters.
The wrist-image encoder is shared across the left and right cameras and counted once.
Table~\ref{tab:model_parameters} provides the module-wise breakdown, including the inherited dense tracking head retained in the implementation.
Per-module counts are rounded independently, so their displayed sum may differ slightly from the reported total.

\begin{table}[!ht]
  \centering
  \caption{\textbf{Model parameter counts.} Counts include frozen modules and retained heads; the wrist encoder is shared across both cameras.}
  \label{tab:model_parameters}
  \small
  \setlength{\tabcolsep}{4pt}
  \renewcommand{\arraystretch}{1.08}
  \begin{tabularx}{\linewidth}{@{}>{\raggedright\arraybackslash}Xr@{}}
    \toprule
    \textbf{Module} & \textbf{Parameters (M)} \\
    \midrule
    Visual backbone & 1,136.491 \\
    Motion decoder & 283.340 \\
    TCP query encoder & 0.808 \\
    TCP trajectory prediction head & 0.800 \\
    Depth and ray geometry head & 50.115 \\
    Camera parameter decoder & 18.908 \\
    Dense tracking head (disabled) & 31.212 \\
    Shared wrist DINOv2-Large & 304.369 \\
    T5-base text encoder & 109.629 \\
    Added action and conditioning modules & 336.462 \\
    \midrule
    \textbf{Total} & \textbf{2,272.133} \\
    \bottomrule
  \end{tabularx}
\end{table}

\noindent\textbf{Hyperparameters.}
Table~\ref{tab:hyperparameters} summarizes the shared and stage-specific training hyperparameters.

\begin{table}[!ht]
  \centering
  \caption{\textbf{Training hyperparameters.} Common settings apply to both DROID Stage~1 and the RoboTwin Stage~2 experiments.}
  \label{tab:hyperparameters}
  \small
  \setlength{\tabcolsep}{6pt}
  \renewcommand{\arraystretch}{1.08}
  \begin{tabular*}{\textwidth}{@{\extracolsep{\fill}}p{0.17\textwidth}p{0.43\textwidth}p{0.32\textwidth}@{}}
    \toprule
    \textbf{Scope} & \textbf{Hyperparameter} & \textbf{Value} \\
    \midrule
    Common & Optimizer & AdamW \\
     & Adam coefficients $(\beta_1,\beta_2)$ & $(0.9,0.95)$ \\
     & Adam $\epsilon$ & $10^{-8}$ \\
     & Weight decay & $0.01$ \\
     & Mixed precision & bfloat16 \\
     & Gradient accumulation steps & 2 \\
     & Gradient clipping norm & $1.0$ \\
     & Learning-rate schedule & Cosine \\
     & Warm-up duration & 1,000 optimizer steps \\
     & Minimum / peak learning rate & $0.1$ \\
     & Training duration & 500,000 optimizer steps \\
    \midrule
    DROID Stage~1 & Peak LR: visual backbone & $10^{-5}$ \\
     & Peak LR: geometry head & $2\times10^{-5}$ \\
     & Peak LR: sparse TCP modules & $10^{-4}$ \\
    \midrule
    RoboTwin Stage~2 & Batch size per device & 8 samples \\
     & Observation history per sample & 4 frames \\
     & Peak LR: geometry head & $2\times10^{-5}$ \\
     & Peak LR: shared TCP decoder & $10^{-5}$ \\
     & Peak LR: backbone and other trainable modules & $10^{-4}$ \\
     & Frozen modules & Camera head; pretrained wrist-image and language encoders \\
    \bottomrule
  \end{tabular*}
\end{table}

\subsection{Frame Sampling Strategy}

\noindent\textbf{Stage~1 Sampling.}
Each minibatch contains a fixed total of 18 frames per device.
For each minibatch, we randomly choose the number of clips $S\in\{1,2,3,6,9\}$ and sample $18/S$ frames per clip, yielding the respective clip lengths $18$, $9$, $6$, $3$, and $2$.
Each clip comes from a single camera stream, with a temporal stride sampled uniformly from the feasible values in $\{1,2,3,4,5\}$ and a randomly chosen starting frame.
The sampled stride is constant within a clip and can differ across clips.
Each clip is reversed with probability $0.5$; images and all associated labels are reordered together.
The first frame in the resulting sequence serves as the TCP query frame, so queries can originate at different points within an episode and support tracking in either temporal direction.
Temporal supervision uses the signed time intervals between sampled frames, accounting for both subsampling and reversal.
Clips remain within valid trajectory segments and do not cross detected discontinuities.

\noindent\textbf{Stage~2 Sampling.}
For each sample, we randomly choose a current frame $t$ and use the four consecutive observations ending at $t$ as history ($L=4$), with the following 32 TCP states as targets ($h=32$).
Sampling is strictly forward with unit temporal stride and no reversal.
If fewer than four historical frames are available within the current valid segment, we repeat its earliest observed frame.
If fewer than 32 future states remain, we repeat the last valid state for storage and mask the padded positions in attention and loss computation.
The current robot state and wrist images are taken at $t$, and a language instruction is sampled from those associated with the episode.

\subsection{Sliding Window Inference}
\label{app:sliding_window}

\begin{figure}[!t]
\centering
\includegraphics[width=\linewidth]{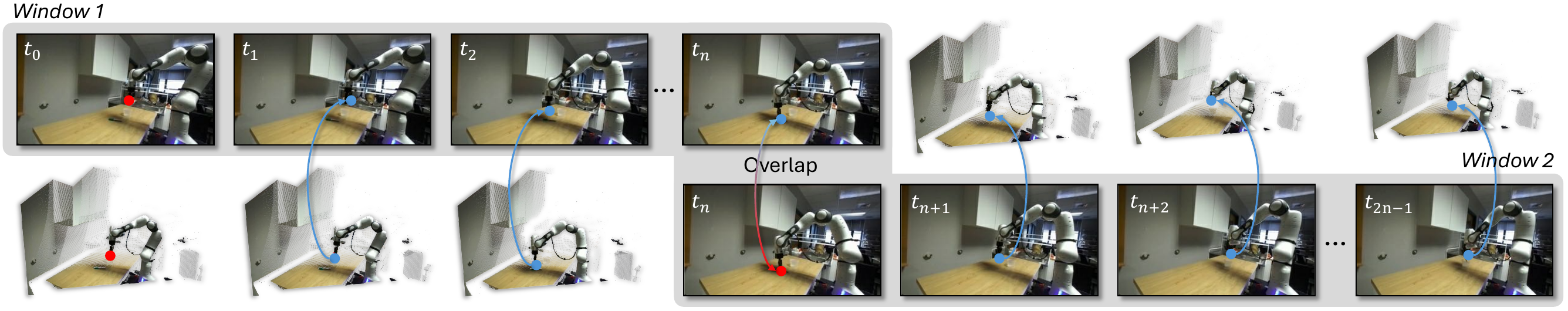}
\caption{\textbf{Sliding-window TCP tracking inference.} \ours supports inference on sequences of varying lengths.
To track a long sequence, only the TCP’s 2D location in the first frame needs to be specified.
The predicted 3D TCP position, \((x,y,z)\), at the final frame can then be projected onto the camera plane and used as the new query for the next window.}
\label{fig:sliding}
\end{figure}

As illustrated in Fig.~\ref{fig:sliding}, \ours extends TCP tracking to long videos by propagating visual queries across overlapping windows.
Only the initial frame requires 2D TCP queries for the available arms; subsequent windows use queries derived from the model's recovered TCP positions.

\noindent\textbf{TCP Tracking.}
For example, the tracker processes $n-1$ frame windows by default, with a configurable length of 2--18 frames.
Consecutive windows share exactly one boundary frame, so a full $n-1$ frame window advances the sequence by $n-2$ frames; the final window can be shorter.
After processing a window, we project its final-frame 3D TCP positions into the shared boundary image using the camera intrinsics decoded from the ray map.
These projected positions become the reference-frame queries for the next window, as shown in Fig.~\ref{fig:sliding}.

\noindent\textbf{Conditional TCP Generation.}
For each window, we project the predicted TCP positions in the $L$ history frames into their corresponding images and sample local feature tokens from the shared visual representations.
These tokens are fed into the DiT to condition future TCP generation.

\section{Tracking Evaluation}
\label{app:tracking_metrics}

This section describes the datasets, protocol, and metrics used for the tracking evaluation in Table~\ref{tab:main_table} and reports additional results.
All TCP poses are expressed in the camera coordinate system of the corresponding frame.

\subsection{Datasets}
\label{app:evaluation}

\noindent\textbf{DROID.}
The PointWorld-annotated DROID subset contains 33,966 episodes.  
We use a fixed episode-level split with seed 42, assigning approximately 90\% of the episodes to training. From the remaining 10\%, we randomly sample 100 episodes for evaluation.
Each DROID episode has two third-person view cameras and is sampled at 15\,Hz.

\noindent\textbf{RoboLab.}
The additional RoboLab set contains one episode for each of 100 distinct tasks and is used exclusively for in-the-wild evaluation.
We use $\pi_{0.5}$~\cite{intelligence2025pi05} as the policy to generate trajectories, including both successful and failed trajectories.
During data collection, we randomize the backgrounds and camera positions.
RoboLab is sampled at 15 Hz.

\subsection{Evaluation Protocol}
\label{app:protocol}

\begin{figure}[!t]
\centering
\includegraphics[width=\linewidth]{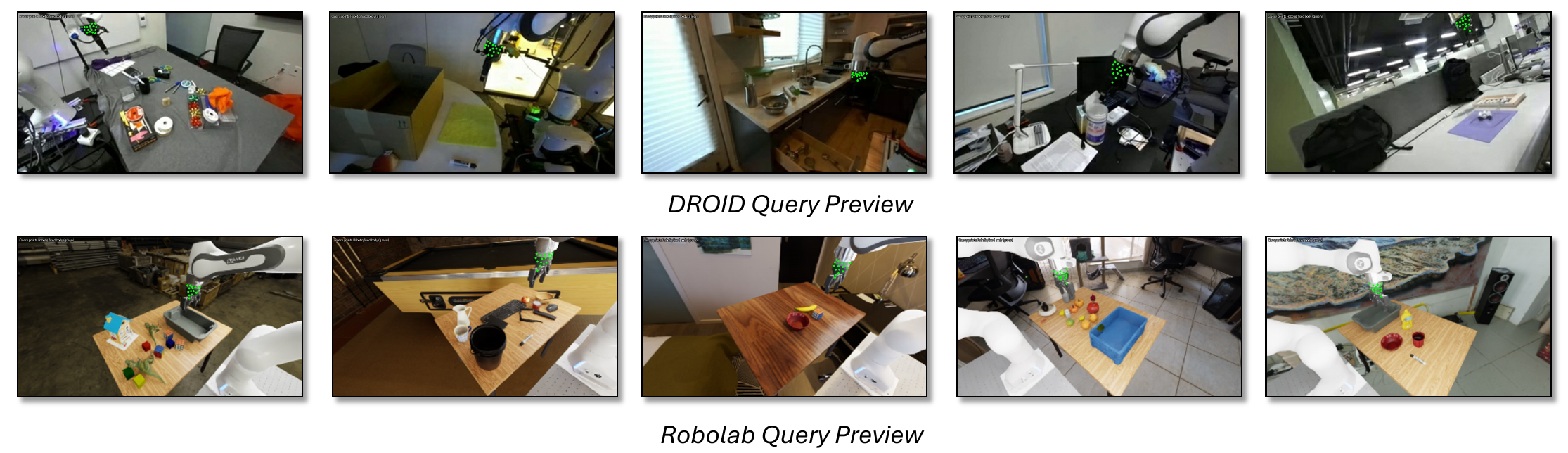}
\caption{\textbf{Query points on DROID and Robolab for baseline evaluation.} For baselines evaluated by rigid fitting of point tracks, we select 16 point groups on the gripper in the first frame of each window using ground-truth annotations, track them throughout the window, and fit the TCP pose from the tracked points.}
\label{fig:query}
\end{figure}

\noindent\textbf{Baseline Evaluation Windows.}
The 4RC~\cite{luo2026fourrc}, Any4D~\cite{karhade2026any4d}, Track4World~\cite{lu2026track4world}, VDPM~\cite{sucar2026vdpm4dvideoreconstruction}, MoSE3~\cite{cheng2026mose3}, and CoTracker3~\cite{cotracker} baselines use independent, non-overlapping nine-frame windows, with a potentially shorter final window. Each window's query frame is excluded from tracking metrics.

\noindent\textbf{TCP Pose Recovery from Point Tracks.}
For 4RC, Any4D, Track4World, VDPM, and CoTracker3, we recover TCP poses by rigid fitting of point tracks.
As shown in Fig.~\ref{fig:query}, each window is initialized with ground-truth-projected points on the fixed Robotiq gripper body.
Each query is associated with a known surface point on the gripper URDF model, and this query-to-model correspondence is retained throughout the window.
CoTracker3 uses VGGT-$\Omega$ depth predicted for every frame, together with the calibrated camera intrinsics, to lift its 2D tracks into 3D.
For all these baselines, including CoTracker3, we calibrate the predicted 3D point group's scale using the known metric dimensions of the corresponding URDF points at window initialization and keep this scale fixed throughout the window.

At each tracked frame, we robustly fit a rigid transformation from the corresponding gripper model points to their scale-calibrated 3D predictions.
The fitted rotation and translation describe the gripper body's pose in that frame's camera coordinates.
We apply the known fixed body-to-TCP offset to recover the TCP position and use the fitted body rotation as the TCP orientation, consistent with the dataset's TCP convention.
Repeating this fit across frames converts the point-group tracks into a TCP pose trajectory.

\noindent\textbf{MoSE3 Native \SE Motion Estimation.}
For MoSE3, we use its official rigidity-weighted per-point \SE estimation procedure.
At the start of each window, we perform the same URDF-based scale calibration and rigid fitting to initialize the TCP pose from the predicted 3D point group.
We then propagate this pose using the \SE motion associated with a fixed gripper-surface query, selected as the query nearest the image-space mean of the group, and transform it into each frame's camera coordinates using the predicted camera poses.
The initialization scale is applied to both motion and camera translations.
No additional CAD-to-track rigid fitting is performed at subsequent frames.
The same evaluation windows, query-frame exclusions, and metric definitions apply.

\noindent\textbf{Metrics and Alignment.}
We evaluate the recovered TCP positions with APD and MPE, and the recovered pose trajectory with ATE, RRE, and RTE in Table~\ref{tab:main_table}.
APD uses one GT-based global-median scale per episode--camera trajectory, computed by Eq.~\ref{eq:eval_gm_scale}.
MPE is computed without evaluation-time alignment; each method retains the scale of its recovered or directly predicted TCP trajectory.
ATE, RRE, and RTE share one position-based Sim(3) alignment per episode trajectory: its scale, rotation, and translation transform positions, and its rotation transforms orientations.
3D Tracking and \SE for those baselines are computed only over each method's successful windows.
SR reports the percentage of successfully processed windows.

\ours uses nine-frame windows with one shared boundary frame and propagates predicted queries after the initial ground-truth query, following the \textit{Sliding Window Inference} strategy in Appx.~\ref{app:sliding_window}.
Only its initial query frame is excluded, and overlapping frames are scored once using the preceding window's prediction.

\subsection{Tracking Success Rate (SR)}

For each camera sequence, we compute 
\begin{equation}
\mathrm{SR}=N_{\mathrm{succ}}/(N_{\mathrm{succ}}+N_{\mathrm{fail}}),
\end{equation}
and report the unweighted mean across episode.
Eligible targets have finite ground-truth TCP positions with positive depth and projections inside the image, with query-frame exclusions as described above; windows without eligible targets are excluded.
A window succeeds if initialization is valid and every eligible target has a finite TCP prediction with positive depth.
Failures include 1) invalid gripper-query initialization, 2) failed CAD-scale estimation, 3) failed initial TCP recovery or inference, and 4) missing, non-finite, or non-positive-depth target predictions.
Failed windows remain in the SR denominator but are excluded from the metrics in all the tables.

\subsection{Depth Metrics}

We reuse $\Omega_d$ from Appx.~\ref{app:geometry_loss}, restricted to each frame and ground-truth depths in $(0,3]$\,m.
Per-frame median alignment uses
\begin{equation}
    \tilde D_p=s_D\hat D_p,\qquad
    s_D=\frac{\operatorname{median}_{q\in\Omega_d}D_q}
                  {\operatorname{median}_{q\in\Omega_d}\hat D_q}.
    \label{eq:eval_depth_scale}
\end{equation}
Only a multiplicative scale is fitted from GT depth.
For metric-depth evaluation in Table~\ref{tab:additional_depth_metric}, we instead use the native prediction $\hat D_p$ directly, with no GT-based scale correction.
Define $r_p=\max(\tilde D_p/D_p,D_p/\tilde D_p)$ for $\tilde D_p>0$, and $r_p=\infty$ otherwise.
The three depth metrics are
\begin{equation}
    \begin{aligned}
        \mathrm{AbsRel}
        &=\frac{1}{|\Omega_d|}\sum_{p\in\Omega_d}
          \frac{|\tilde D_p-D_p|}{D_p},\\
        \mathrm{MAE}
        &=\frac{10^3}{|\Omega_d|}\sum_{p\in\Omega_d}
          |\tilde D_p-D_p|,\\
        \delta_1
        &=\frac{100}{|\Omega_d|}\sum_{p\in\Omega_d}
          \mathbf1[r_p<1.25].
    \end{aligned}
    \label{eq:eval_depth_metrics}
\end{equation}
MAE is reported in millimeters and $\delta_1$ in percent.

\subsection{3D TCP Tracking Metrics}

We retain the TCP notation from Appx.~\ref{app:losses}, omitting the arm index $a$ below; positions are in meters.
Let $\mathcal E$ contain the target frames from successful tracking windows, excluding query frames as described above.

\noindent\textbf{APD.}
We average distance-threshold accuracies at $\mathcal D=\{5,10,20,50\}$\,mm, using one global-median scale per episode--camera trajectory:
\begin{equation}
    s_{\mathrm{GM}}
    =\frac{\operatorname{median}_{i\in\mathcal E}\|\mathbf t_i\|_2}
           {\operatorname{median}_{i\in\mathcal E}\|\hat{\mathbf t}_i\|_2}.
    \label{eq:eval_gm_scale}
\end{equation}
The scale is fitted across retained successful windows, with no rotation or translation alignment.
Writing $e_i=10^3\|s_{\mathrm{GM}}\hat{\mathbf t}_i-\mathbf t_i\|_2$ in millimeters, we compute
\begin{equation}
    \mathrm{APD}
    =\frac{100}{4|\mathcal E|}
      \sum_{d\in\mathcal D}\sum_{i\in\mathcal E}\mathbf1[e_i\leq d].
    \label{eq:eval_apd}
\end{equation}

\noindent\textbf{Mean Position Error (MPE).}
MPE is the mean Euclidean TCP position error in metric coordinates, reported in millimeters:
\begin{equation}
    \mathrm{MPE}
    =\frac{10^3}{|\mathcal E|}\sum_{i\in\mathcal E}
      \|\hat{\mathbf t}_i-\mathbf t_i\|_2.
    \label{eq:eval_mpe}
\end{equation}
No evaluation-time alignment is applied; rigid-group baselines retain their initialization scale calibration.

For the GM-aligned comparison in Table~\ref{tab:additional_tracking}, we additionally report
\begin{equation}
    \mathrm{MPE}_{\mathrm{GM}}
    =\frac{10^3}{|\mathcal E|}\sum_{i\in\mathcal E}
      \|s_{\mathrm{GM}}\hat{\mathbf t}_i-\mathbf t_i\|_2.
    \label{eq:eval_mpe_gm}
\end{equation}
The scale $s_{\mathrm{GM}}$ is fitted once per episode trajectory using GT positions across all retained successful windows, as in Eq.~\ref{eq:eval_gm_scale}.
$\mathrm{PCK}_d$ thresholds the GM-aligned error, while $\mathrm{Acc}_d$ thresholds the metric error, both at $d$\,mm.

\subsection{\texorpdfstring{\SE}{SE(3)} Trajectory Metrics}

ATE, RRE, and RTE are reported as RMSE.
We fit one position-based Umeyama Sim(3) $(s_{\mathrm{aln}},\mathbf R_{\mathrm{aln}},\mathbf t_{\mathrm{aln}})$ per episode--camera sequence on $\mathcal E$, shared by all three metrics.
The aligned poses $\bar{\mathbf T}_i=(\bar{\mathbf R}_i,\bar{\mathbf t}_i)$ satisfy
\begin{equation}
    \begin{aligned}
        \bar{\mathbf t}_i
        &=s_{\mathrm{aln}}\mathbf R_{\mathrm{aln}}\hat{\mathbf t}_i
          +\mathbf t_{\mathrm{aln}},\\
        \bar{\mathbf R}_i
        &=\mathbf R_{\mathrm{aln}}\hat{\mathbf R}_i.
    \end{aligned}
    \label{eq:eval_aligned_pose}
\end{equation}
Relative errors use the full pose residual
\begin{equation}
    \mathbf E_{ij}
    =(\mathbf T_i^{-1}\mathbf T_j)^{-1}
      (\bar{\mathbf T}_i^{-1}\bar{\mathbf T}_j),
    \label{eq:eval_relative_pose}
\end{equation}
with $j=i+1$ in the original frame indices and valid poses at both retained endpoints.
RTE uses the translation norm of $\mathbf E_{ij}$, and RRE its rotation angle, computed using $d_R$ in Eq.~\ref{eq:app_rotation_distance}.

\subsection{Gripper-state Accuracy}
\label{app:gripper_metrics}

We threshold $\hat g_i$ and $g_i$ at $0.5$, treating values $\geq0.5$ as open and smaller values as closed.
Acc. is the average agreement between the resulting binary states, reported as a percentage.

\subsection{\ours Tracker Evaluation Results}
\label{app:tracker}

We extend Table~\ref{tab:main_table} with complete depth metrics, additional 3D TCP tracking and \SE statistics, and an evaluation of the camera head.

\noindent\textbf{Metric depth.}
Table~\ref{tab:additional_depth_metric} reports metric depth evaluation on DROID and RoboLab, using native depth predictions in meters without alignment to GT.
Results use the GT depth mask within 3\,m.
We compare DA3NESTED-GIANT-LARGE, Any4D, MapAnything, and \ours, for which unaligned metric-depth results are available.
\ours achieves the lowest AbsRel, MAE, and RMSE among these methods on both datasets, with MAE of 28.37\,mm on DROID and 113.92\,mm on RoboLab.
Its $\delta_1$ accuracies are 97.79\% and 90.53\%, respectively.
These results evaluate absolute scale accuracy together with depth reconstruction error.

\begin{table}[!ht]
\centering
\begin{minipage}{\textwidth}
\centering
\caption{\textbf{Metric depth evaluation on DROID and RoboLab.}
Native depth predictions are evaluated in meters without GT-based scale alignment, using the GT depth mask within 3\,m.
\protect\colorbox{successTableBest}{Best} and \protect\colorbox{successTableSecond}{second-best} values are highlighted independently for each dataset.}
\label{tab:additional_depth_metric}
\begingroup
\small
\renewcommand{\arraystretch}{1.1}
\begin{adjustbox}{max width=\linewidth}
\begin{tabular}{l*{10}{c}}
\toprule
Method & AbsRel $\downarrow$ & MAE (mm) $\downarrow$ & RMSE (mm) $\downarrow$ & SqRel (m)$\downarrow$
 & RMSE$_{\log}\downarrow$ & $\log_{10}\downarrow$ & SILog $\downarrow$
 & $\delta_1(\%)\uparrow$ & $\delta_2(\%)\uparrow$ & $\delta_3(\%)\uparrow$ \\
\midrule
\multicolumn{11}{l}{\textbf{DROID (Eval set)}} \\
\midrule
DA3-GIANT~\cite{lin2025depth} & \cellcolor{successTableSecond}0.2925 & \cellcolor{successTableSecond}280.36 & \cellcolor{successTableSecond}408.87 & \cellcolor{successTableSecond}0.1486 & \cellcolor{successTableSecond}0.3773 & \cellcolor{successTableSecond}0.1270 & 32.67 & \cellcolor{successTableSecond}49.98 & \cellcolor{successTableSecond}78.42 & \cellcolor{successTableSecond}91.50 \\
MapAnything~\cite{keetha2026mapanything} & 0.7756 & 616.20 & 732.90 & 0.7072 & 0.5838 & 0.2276 & \cellcolor{successTableSecond}27.03 & 13.73 & 43.24 & 74.69 \\
Any4D~\cite{karhade2026any4d} & 1.1698 & 906.84 & 1039.32 & 1.4912 & 0.7530 & 0.3029 & 28.42 & 8.18 & 27.84 & 51.74 \\
\midrule
\ours & \cellcolor{successTableBest}0.0303 & \cellcolor{successTableBest}28.37 & \cellcolor{successTableBest}110.75 & \cellcolor{successTableBest}0.0156 & \cellcolor{successTableBest}0.0995 & \cellcolor{successTableBest}0.0116 & \cellcolor{successTableBest}9.91 & \cellcolor{successTableBest}97.79 & \cellcolor{successTableBest}98.87 & \cellcolor{successTableBest}99.32 \\
\midrule
\multicolumn{11}{l}{\textbf{RoboLab (Unseen)}} \\
\midrule
DA3-GIANT~\cite{lin2025depth} & \cellcolor{successTableSecond}0.3066 & \cellcolor{successTableSecond}185.83 & \cellcolor{successTableSecond}245.87 & \cellcolor{successTableSecond}0.1868 & \cellcolor{successTableSecond}0.3237 & \cellcolor{successTableSecond}0.1134 & 24.26 & \cellcolor{successTableSecond}54.08 & \cellcolor{successTableSecond}83.15 & \cellcolor{successTableSecond}94.75 \\
MapAnything~\cite{keetha2026mapanything} & 1.1662 & 741.13 & 818.13 & 1.1475 & 0.7575 & 0.3172 & \cellcolor{successTableSecond}19.52 & 1.57 & 10.28 & 43.82 \\
Any4D~\cite{karhade2026any4d} & 1.5807 & 985.87 & 1058.42 & 1.9600 & 0.9258 & 0.3926 & \cellcolor{successTableBest}19.07 & 0.51 & 3.33 & 15.09 \\
\midrule
\ours & \cellcolor{successTableBest}0.2074 & \cellcolor{successTableBest}113.92 & \cellcolor{successTableBest}218.97 & \cellcolor{successTableBest}0.1752 & \cellcolor{successTableBest}0.2408 & \cellcolor{successTableBest}0.0659 & 21.44 & \cellcolor{successTableBest}90.53 & \cellcolor{successTableBest}95.24 & \cellcolor{successTableBest}96.16 \\
\bottomrule
\end{tabular}
\end{adjustbox}
\par\smallskip
\endgroup
\end{minipage}
\end{table}

\begin{table}[!ht]
\centering
\begin{minipage}{1\textwidth}
\centering
\caption{\textbf{Additional 3D TCP tracking metrics.}
Results are averages over successful windows.
$\mathrm{PCK}_d$ and $\mathrm{Acc}_d$ report percentages of position errors within $d$\,mm under GT scale alignment and metric scale, respectively.
\protect\colorbox{successTableBest}{Best} and \protect\colorbox{successTableSecond}{second-best} values are highlighted independently for each dataset.}
\label{tab:additional_tracking}
\begingroup
\small
\setlength{\tabcolsep}{4pt}
\renewcommand{\arraystretch}{1.1}
\begin{adjustbox}{max width=\linewidth}
\begin{tabular}{l*{10}{c}}
\toprule
 & \multicolumn{5}{c}{GT scale alignment (GM)} & \multicolumn{5}{c}{Metric scale (no GT alignment)} \\
\cmidrule(lr){2-6}\cmidrule(lr){7-11}
Method & MPE$_{\mathrm{GM}} (mm)$ $\downarrow$ & $\mathrm{PCK}_{5} (\%)\uparrow$ & $\mathrm{PCK}_{10}(\%)\uparrow$
 & $\mathrm{PCK}_{20}(\%)\uparrow$ & $\mathrm{PCK}_{50}(\%)\uparrow$
 & Median(mm) $\downarrow$ & P95(mm) $\downarrow$ & $\mathrm{Acc}_{10}(\%)\uparrow$
 & $\mathrm{Acc}_{20}(\%)\uparrow$ & $\mathrm{Acc}_{50}(\%)\uparrow$ \\
\midrule
\multicolumn{11}{l}{\textbf{DROID (Eval set)}} \\
\midrule
4RC~\cite{luo2026fourrc} & 103.14 & 0.15 & 0.76 & 4.34 & 29.73 & 84.70 & 276.93 & 0.96 & 6.19 & 32.33 \\
Track4World~\cite{lu2026track4world} & 141.96 & 0.00 & 0.08 & 1.43 & 13.20 & 174.06 & 338.74 & 0.03 & 0.54 & 5.01 \\
Any4D~\cite{karhade2026any4d} & 171.01 & 0.01 & 0.05 & 0.52 & 8.69 & 229.73 & 430.68 & 0.02 & 0.05 & 1.71 \\
VDPM~\cite{sucar2026vdpm4dvideoreconstruction} & 119.12 & 0.01 & 0.28 & 1.68 & 16.71 & 182.45 & 336.68 & 0.02 & 0.42 & 4.10 \\
CoTracker3~\cite{cotracker} & 87.26 & \cellcolor{successTableSecond}0.41 & \cellcolor{successTableSecond}2.34 & \cellcolor{successTableSecond}10.68 & \cellcolor{successTableSecond}42.73 & \cellcolor{successTableSecond}67.66 & \cellcolor{successTableSecond}241.52 & \cellcolor{successTableSecond}4.18 & \cellcolor{successTableSecond}14.81 & \cellcolor{successTableSecond}44.76 \\
MoSE3~\cite{cheng2026mose3} & \cellcolor{successTableSecond}86.58 & 0.17 & 1.18 & 7.19 & 40.98 & 83.69 & 245.35 & 1.12 & 5.95 & 32.29 \\
\midrule
\ours & \cellcolor{successTableBest}10.76 & \cellcolor{successTableBest}38.28 & \cellcolor{successTableBest}74.55 & \cellcolor{successTableBest}89.78 & \cellcolor{successTableBest}96.52 & \cellcolor{successTableBest}10.71 & \cellcolor{successTableBest}22.34 & \cellcolor{successTableBest}75.16 & \cellcolor{successTableBest}89.52 & \cellcolor{successTableBest}94.99 \\
\midrule
\multicolumn{11}{l}{\textbf{RoboLab (Unseen)}} \\
\midrule
4RC~\cite{luo2026fourrc} & 127.50 & 0.02 & 0.22 & 1.80 & 17.63 & 137.70 & 370.72 & 0.52 & 3.11 & 19.50 \\
Track4World~\cite{lu2026track4world} & 141.79 & 0.00 & 0.04 & 0.58 & 9.16 & 192.11 & 406.60 & 0.02 & 0.18 & 3.08 \\
Any4D~\cite{karhade2026any4d} & 179.75 & 0.00 & 0.02 & 0.16 & 4.42 & 215.42 & 469.83 & 0.01 & 0.15 & 1.50 \\
VDPM~\cite{sucar2026vdpm4dvideoreconstruction} & \cellcolor{successTableSecond}84.42 & 0.08 & 0.55 & 3.84 & \cellcolor{successTableSecond}35.15 & \cellcolor{successTableSecond}99.02 & \cellcolor{successTableSecond}235.26 & 0.32 & 2.46 & 22.80 \\
CoTracker3~\cite{cotracker} & 114.20 & \cellcolor{successTableSecond}0.15 & \cellcolor{successTableSecond}0.78 & 4.43 & 29.77 & 101.00 & 345.16 & \cellcolor{successTableSecond}1.72 & \cellcolor{successTableSecond}6.95 & \cellcolor{successTableSecond}30.20 \\
MoSE3~\cite{cheng2026mose3} & 110.98 & 0.10 & 0.64 & \cellcolor{successTableSecond}4.50 & 29.08 & 100.45 & 329.27 & 0.95 & 5.58 & 26.34 \\
\midrule
\ours & \cellcolor{successTableBest}21.55 & \cellcolor{successTableBest}2.83 & \cellcolor{successTableBest}16.08 & \cellcolor{successTableBest}60.86 & \cellcolor{successTableBest}95.07 & \cellcolor{successTableBest}26.32 & \cellcolor{successTableBest}51.54 & \cellcolor{successTableBest}10.04 & \cellcolor{successTableBest}48.42 & \cellcolor{successTableBest}87.21 \\
\bottomrule
\end{tabular}
\end{adjustbox}
\endgroup
\end{minipage}
\end{table}

\noindent\textbf{3D TCP tracking.}
Table~\ref{tab:additional_tracking} compares 3D TCP tracking in GT scale alignment and metric scale.
GT scale alignment rescales each episode--camera trajectory by $s_{\mathrm{GM}}=\operatorname{median}\|\mathbf t_i\|_2/\operatorname{median}\|\hat{\mathbf t}_i\|_2$.
Metric scale uses \ours predictions in meters and baseline predictions scaled to the known CAD dimensions at each window's initialization, following the protocol in Sec.~\ref{app:protocol}.
$\mathrm{MPE}_{\mathrm{GM}}$ and $\mathrm{PCK}_d$ use the rescaled positions; Median, P95, and $\mathrm{Acc}_d$ use metric positions.

\noindent\textbf{\SE trajectories.}
Table~\ref{tab:additional_se3} reports the mean, median, and P95 of RRE and RTE after per-camera Sim(3) alignment.
Relative-pose errors are computed between adjacent frames within each inference window.

\begin{table}[!ht]
\centering
\caption{\textbf{Additional \SE trajectory error statistics.}
Mean, median, and P95 are computed per camera over successful windows and then macro-averaged.}
\label{tab:additional_se3}
\begingroup
\small
\setlength{\tabcolsep}{4pt}
\renewcommand{\arraystretch}{1.1}
\begin{adjustbox}{max width=\linewidth}
\begin{tabular*}{\linewidth}{@{\extracolsep{\fill}}l*{6}{c}@{}}
\toprule
 & \multicolumn{3}{c}{RRE ($^\circ$) $\downarrow$} & \multicolumn{3}{c}{RTE (mm) $\downarrow$} \\
\cmidrule(lr){2-4}\cmidrule(lr){5-7}
Method & Mean & Median & P95 & Mean & Median & P95 \\
\midrule
\multicolumn{7}{l}{\textbf{DROID (Eval set)}} \\
\midrule
 4RC~\cite{luo2026fourrc} & 4.74 & 2.15 & 18.37 & 9.24 & 6.79 & 26.35 \\
 Track4World~\cite{lu2026track4world} & 6.40 & 3.50 & 21.93 & 8.65 & 5.98 & 25.04 \\
 Any4D~\cite{karhade2026any4d} & 6.13 & 2.66 & 23.22 & 9.26 & 6.60 & 26.41 \\
 VDPM~\cite{sucar2026vdpm4dvideoreconstruction} & 4.45 & 1.85 & 17.71 & 8.03 & 5.70 & 23.08 \\
 CoTracker3~\cite{cotracker} & 11.67 & 8.51 & 32.37 & 16.30 & 11.38 & 45.61 \\
 MoSE3~\cite{cheng2026mose3} & \cellcolor{successTableSecond}0.95 & \cellcolor{successTableSecond}0.73 & \cellcolor{successTableSecond}2.53 & \cellcolor{successTableSecond}4.50 & \cellcolor{successTableSecond}3.55 & \cellcolor{successTableSecond}11.60 \\
\midrule
\ours & \cellcolor{successTableBest}0.74 & \cellcolor{successTableBest}0.56 & \cellcolor{successTableBest}1.90 & \cellcolor{successTableBest}2.59 & \cellcolor{successTableBest}2.01 & \cellcolor{successTableBest}6.50 \\
\midrule
\multicolumn{7}{l}{\textbf{RoboLab (Unseen)}} \\
\midrule
 4RC~\cite{luo2026fourrc} & 3.76 & 1.58 & 15.28 & 5.79 & 3.98 & 17.06 \\
 Track4World~\cite{lu2026track4world} & 5.41 & 2.56 & 19.75 & 6.08 & 4.10 & 17.92 \\
 Any4D~\cite{karhade2026any4d} & 4.02 & 0.94 & 18.38 & 6.10 & 4.56 & 16.97 \\
 VDPM~\cite{sucar2026vdpm4dvideoreconstruction} & 2.72 & 0.83 & 12.61 & 5.56 & 3.58 & 17.17 \\
 CoTracker3~\cite{cotracker} & 11.39 & 8.38 & 31.78 & 12.93 & 9.34 & 36.86 \\
 MoSE3~\cite{cheng2026mose3} & \cellcolor{successTableBest}0.71 & \cellcolor{successTableBest}0.50 & \cellcolor{successTableBest}2.01 & \cellcolor{successTableSecond}3.58 & \cellcolor{successTableSecond}2.78 & \cellcolor{successTableBest}9.41 \\
\midrule
\ours & \cellcolor{successTableSecond}1.83 & \cellcolor{successTableSecond}0.72 & \cellcolor{successTableSecond}4.79 & \cellcolor{successTableBest}3.44 & \cellcolor{successTableBest}2.38 & \cellcolor{successTableSecond}10.21 \\
\bottomrule
\end{tabular*}
\end{adjustbox}
\endgroup
\end{table}

\noindent\textbf{Camera geometry.}
For this evaluation, we enable the retained camera head to predict camera extrinsics relative to the robot base and camera intrinsics.
Table~\ref{tab:additional_camera} reports mean errors on DROID and RoboLab using camera-wise macro averaging.

\begin{figure}[!t]
\centering
\includegraphics[width=0.95\linewidth]{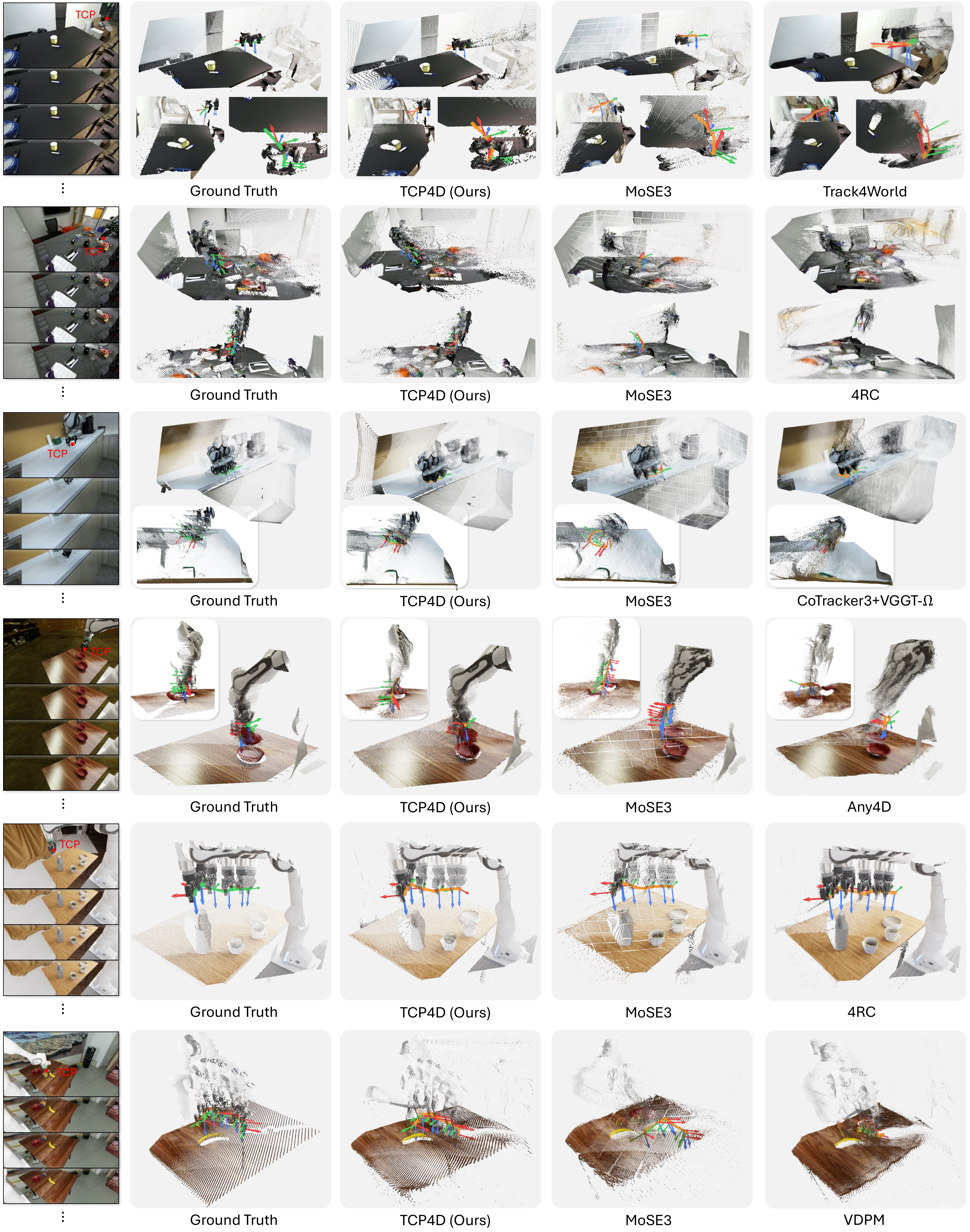}
\caption{\textbf{Qualitative comparison of TCP tracking.} We compare TCP tracking by \ours against dense tracking baselines followed by post-processing. 
The left side shows the input video. 
The red point in the first frame marks the query used by \ours, while the green points indicate the rigid point group used by the baselines.}
\label{fig:vis2}
\end{figure}

\noindent\textbf{More TCP tracking visualizations.} 
We show additional \ours tracking qualitative comparison in Fig.~\ref{fig:vis2}.

\begin{table}[!ht]
\centering
\footnotesize
\setlength{\tabcolsep}{2.5pt}
\caption{\textbf{Camera pose evaluation for \ours.}}
\label{tab:additional_camera}
\begingroup
\footnotesize
\setlength{\tabcolsep}{3pt}
\renewcommand{\arraystretch}{1.1}
\begin{tabular*}{\linewidth}{@{\extracolsep{\fill}}lcc@{}}
\toprule
Metric & DROID & RoboLab \\
\midrule
\multicolumn{3}{l}{\textit{Extrinsics}} \\
\midrule
Translation (mm) & 12.96 & 129.80 \\
Rotation ($^\circ$) & 1.26 & 2.42 \\
Camera center (mm) & 14.80 & 130.99 \\
\midrule
\multicolumn{3}{l}{\textit{Intrinsics}} \\
\midrule
$f_x$ rel. error (\%) & 0.248 & 0.477 \\
$f_y$ rel. error (\%) & 0.179 & 0.412 \\
Horizontal FoV ($^\circ$) & 0.139 & 0.268 \\
Vertical FoV ($^\circ$) & 0.096 & 0.221 \\
\bottomrule
\end{tabular*}
\endgroup
\end{table}

\begin{figure}[!t]
\centering
\includegraphics[width=1\linewidth]{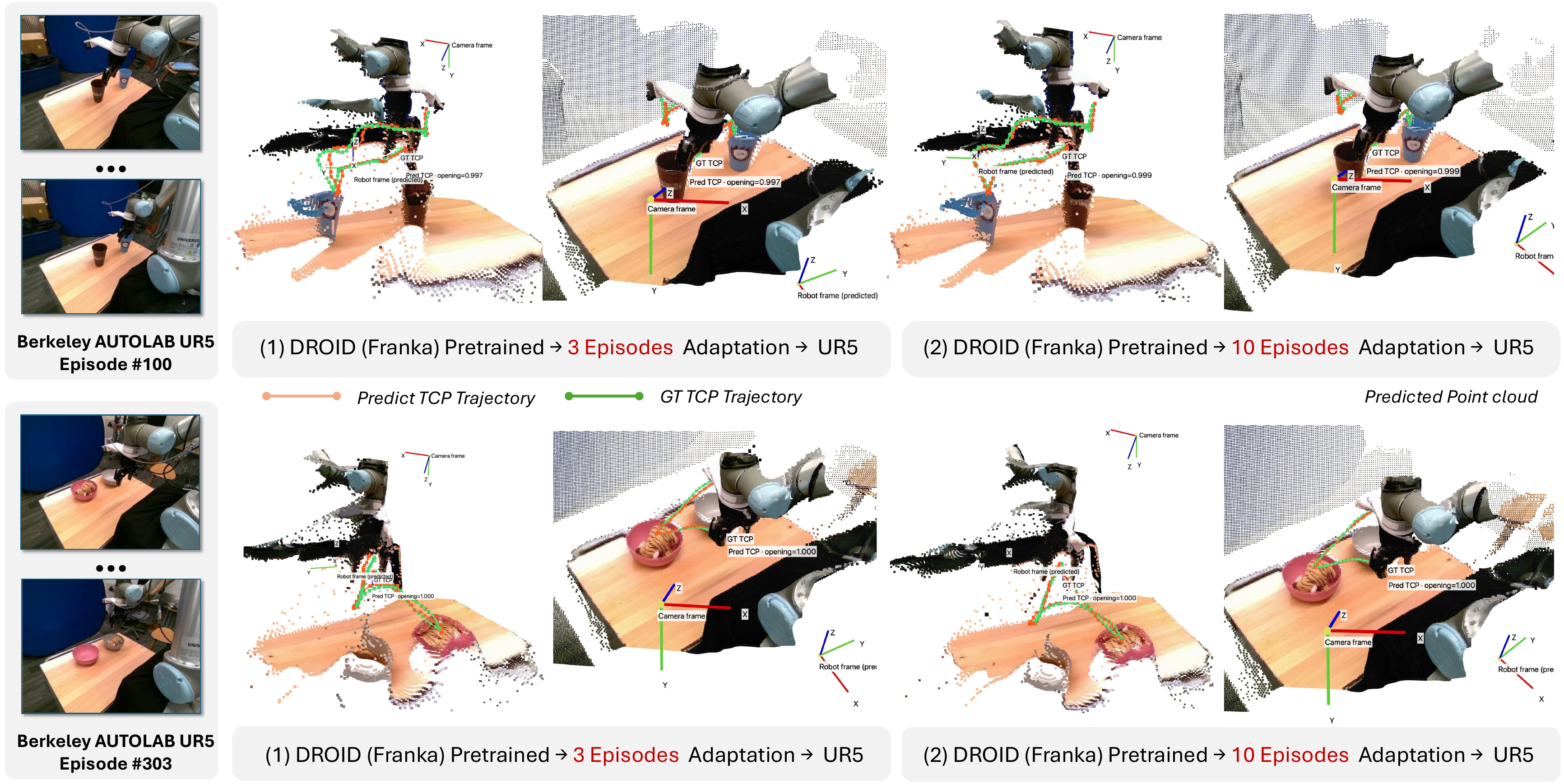}
\caption{\textbf{Visualization of \ours for few-shot cross-embodiment adaptation.} 
We fine-tune \ours using only three and ten randomly selected episodes for three-shot and ten-shot transfer, respectively. 
The orange and green paths represent the predicted and ground-truth TCP trajectories, respectively.
Our method can be readily transferred to new environments and robot embodiments.
}
\label{fig:transfer}
\end{figure}

\begin{figure}[!t]
\centering
\includegraphics[width=1\linewidth]{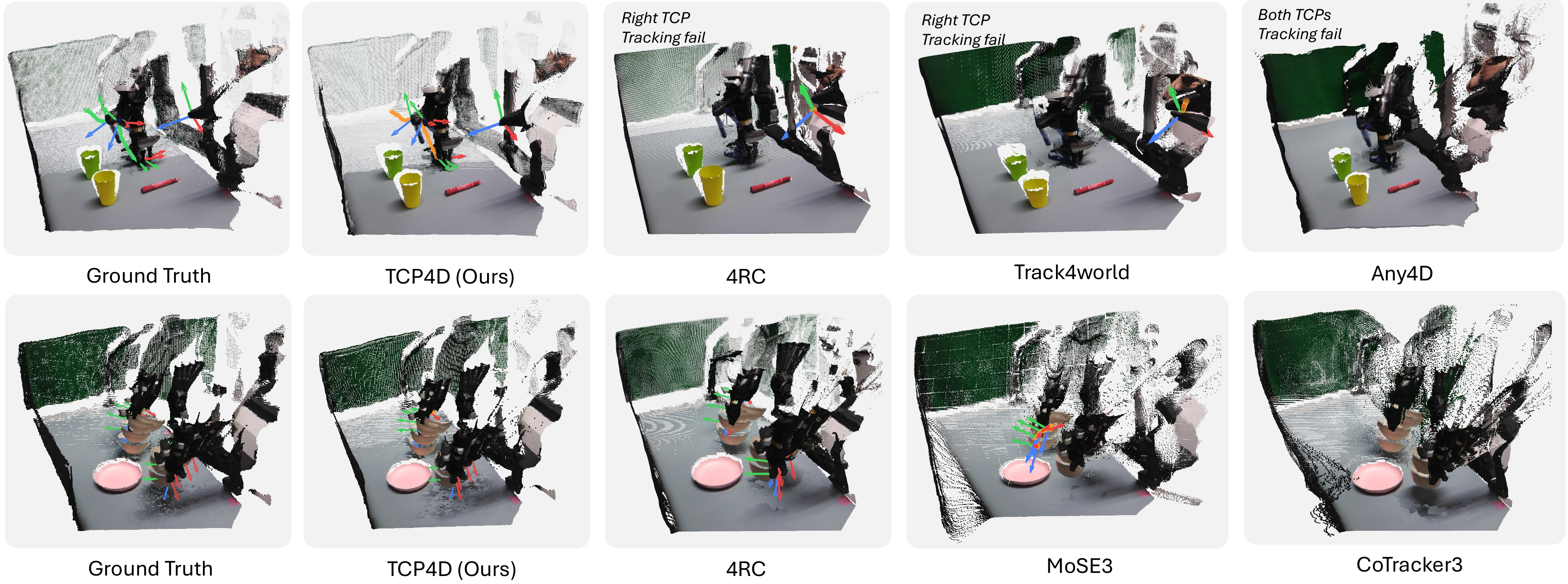}
\caption{\textbf{Qualitative results of \ours Tracker on real-world task.} 
Methods such as 4RC and MOSE3 can experience tracking failures.
}
\label{fig:real_tracker}
\end{figure}

\subsection{Few-Shot Cross-Embodiment Adaptation}
\label{app:adaptation}
After large-scale pretraining on the DROID dataset, \ours demonstrates strong few-shot cross-embodiment adaptation capabilities. 
We evaluate it on the Berkeley AUTOLAB UR5 dataset~\cite{BerkeleyUR5Website}, which contains 1,000 demonstration episodes across different tasks, comprising 97,939 video frames in total.
We fine-tune \ours under two settings: (1) using random 3 episodes for three-shot transfer and (2) using random 10 episodes for few-shot transfer. 
The raw depth data captured by the sensor are post-processed using LingBot-Depth~\cite{lingbot-depth2026}.
Fig.~\ref{fig:transfer} visualizes the three-shot and few-shot transfer results.

\FloatBarrier
\section{Cross-Embodiment Data Training Details}
\label{app:cross_exp}
\subsection{Augmented Dataset}
We augment 50 Aloha-AgileX demonstrations per task into ARX-X5 and UR5-WSG demonstrations across four RoboTwin tasks. For Place Burger Fries, Click Bell, and Handover Mic, we obtain 50 demonstrations for each target embodiment. For Shake Bottle, we obtain 49 ARX-X5 and 48 UR5-WSG demonstrations. This yields 397 augmented demonstrations from 400 source--target episode pairs, corresponding to a demonstration generation rate of 99.25\% (Table~\ref{tab:augmentation_counts}). This rate measures successful demonstration generation among attempted source--target episode pairs, rather than downstream task success. Combining the source and augmented datasets gives 150 demonstrations per task, except Shake Bottle, which contains 147.

\begin{table}[!ht]
\centering
\footnotesize
\setlength{\tabcolsep}{2.5pt}
\footnotesize
\setlength{\tabcolsep}{3pt}
\renewcommand{\arraystretch}{1.12}
\caption{\textbf{Augmented demonstration counts.} Generated demonstrations / attempted source--target episode pairs per target embodiment. Overall, 397/400 pairs yield augmented demonstrations (99.25\%).}
\label{tab:augmentation_counts}
\begin{tabular*}{\linewidth}{@{\extracolsep{\fill}}lrr@{}}
\toprule
\textbf{Task} & \textbf{ARX-X5} & \textbf{UR5-WSG} \\
\midrule
Click Bell & 50/50 & 50/50 \\
Shake Bottle & 49/50 & 48/50 \\
Handover Mic & 50/50 & 50/50 \\
Place Burger Fries & 50/50 & 50/50 \\
\midrule
\textbf{Total} & \textbf{199/200} & \textbf{198/200} \\
\bottomrule
\end{tabular*}
\end{table}

\subsection{Downstream Policy Training}
We use Diffusion Policy with a randomly initialized ResNet18 visual encoder and a conditional 1D U-Net action denoiser. The policy receives three consecutive observations, each comprising a third-person RGB image and a 14-dimensional robot state. The state contains six joint positions and one continuous gripper value for each arm. Recorded source states are unnecessary for the RGB-based augmentation pipeline; downstream policies use the evaluated robot's joint and gripper states as proprioceptive inputs. Architecture and training hyperparameters are summarized in Table~\ref{tab:augmentation_policy_config}.

\begin{table}[!ht]
\centering
\small
\setlength{\tabcolsep}{3pt}
\renewcommand{\arraystretch}{1.08}
\caption{\textbf{Downstream policy configuration.} Settings are shared across tasks and training-data regimes.}
\label{tab:augmentation_policy_config}
\begin{tabularx}{\linewidth}{@{}>{\raggedright\arraybackslash}X>{\raggedright\arraybackslash}X@{}}
\toprule
\textbf{Parameter} & \textbf{Value} \\
\midrule
Policy & Diffusion Policy \\
Visual encoder & ResNet18, from scratch \\
Action denoiser & Conditional 1D U-Net \\
U-Net channel widths & 256, 512, 1024 \\
RGB observation & One camera, $240\times320$ \\
State / action dimension & 14 / 14 \\
Observation / prediction horizon & 3 / 8 \\
Executed actions per prediction & 6 \\
Diffusion scheduler & DDPM (cosine noise schedule) \\
Diffusion / inference steps & 100 / 100 \\
State / action normalization & Min--max to $[-1,1]$ \\
\midrule
Training epochs & 600 \\
Batch size & 128 \\
Optimizer & AdamW \\
Learning rate & $10^{-4}$ \\
Weight decay & $10^{-6}$ \\
AdamW betas & $(0.95,0.999)$ \\
Learning-rate schedule & Cosine \\
Warmup optimizer steps & 500 \\
Maximum EMA decay & 0.9999 \\
Training seed & 0 (one run per setting) \\
\bottomrule
\end{tabularx}
\end{table}

All policies are trained from scratch, with separate models for different tasks. Within each task, the augmented-only setting uses independent ARX-X5 and UR5-WSG models, whereas the source-plus-augmented setting uses one shared model across Aloha-AgileX, ARX-X5, and UR5-WSG. For joint training, demonstrations from the three embodiments are concatenated, and valid observation--action sequences are sampled uniformly without explicit embodiment balancing. All training settings use the same number of epochs; consequently, the number of optimization updates varies with dataset size.

\Needspace{3\baselineskip}
For each task and evaluated embodiment, policy evaluation uses 50 episodes following a common environment-seed schedule across training settings. Success is determined by RoboTwin's task-specific criteria, and the success rate is the fraction of evaluation episodes completed successfully. Results are obtained from one training seed per setting. Downstream results are reported in Fig.~\ref{fig:augmentation_training} and discussed in Sec.~\ref{sec:augmentation_effectiveness}.

\begin{figure}[!t]
\centering
\includegraphics[width=\linewidth]{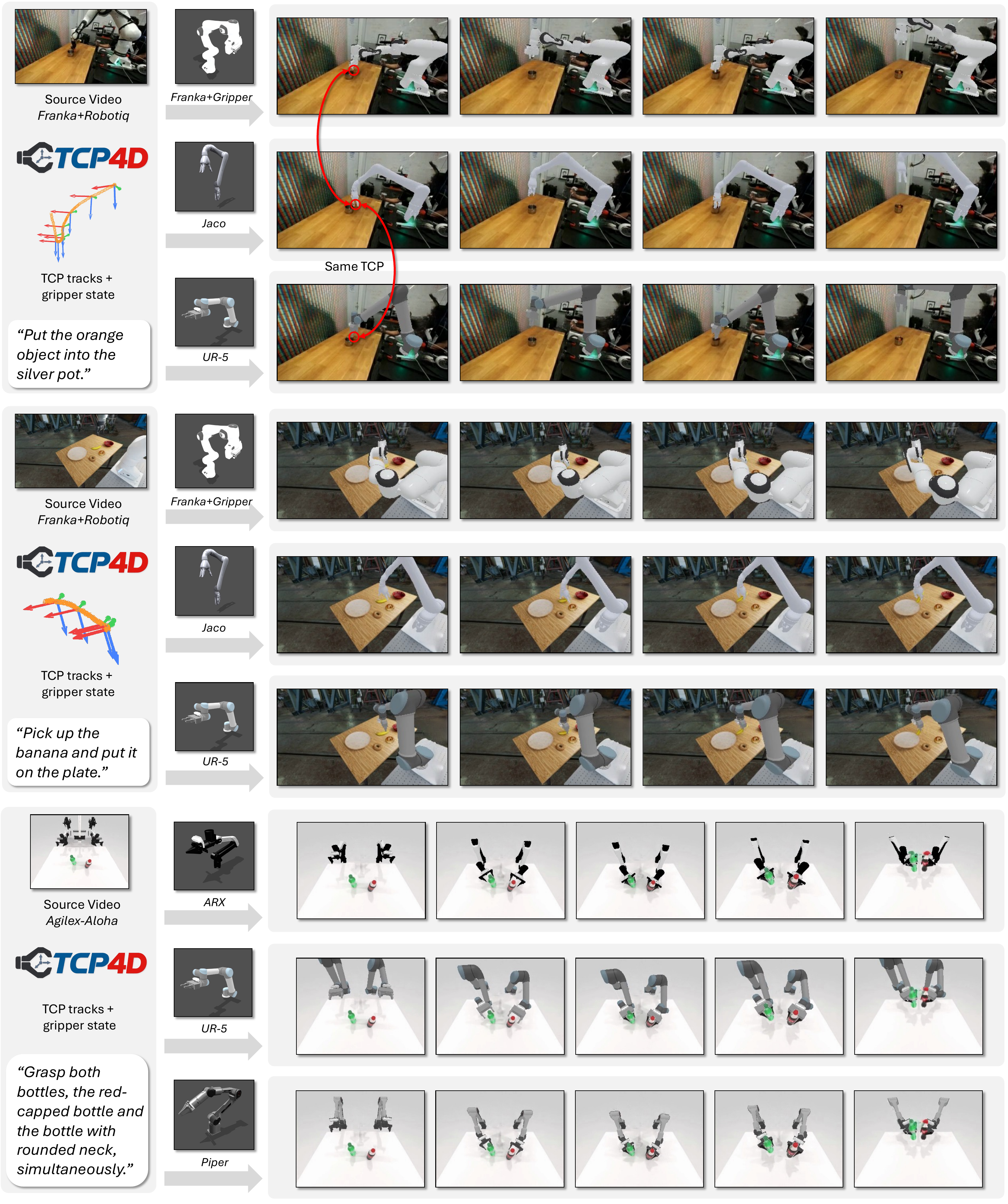}
\caption{\textbf{Qualitative results of our cross-embodiment augmented data.}  Given an RGB-only video, \ours reconstructs the TCP’s 3D trajectory and gripper state, enabling augmentation on any embodiment that satisfies \textit{reachability}. 
The right side shows the resulting augmented trajectories rendered on different target embodiments. }
\label{fig:vis4}
\end{figure}

\noindent\textbf{Qualitative Results.}
Fig.~\ref{fig:vis4} illustrates the rendered cross-embodiment demonstrations. These examples show the visual output of retargeting and composition; downstream policy performance is evaluated separately in Fig.~\ref{fig:augmentation_training}.

\section{Manipulation Evaluation}
\label{app:experiments}

\subsection{\ours Generator Result on RoboTwin}
\label{app:robotwin}

\noindent\textbf{Dataset.}
We collect 50 clean demonstrations for each of 48 tasks, totaling 2,400 training episodes, with third-person and left \& right wrist images as the scene observations.
We exclude the \textit{Dump\_Bin\_BigBin} and \textit{Put\_Object\_Cabinet} tasks because the TCP is totally occluded in the third-person camera view.
RoboTwin is sampled at 16.67\, Hz.

\noindent\textbf{Results.}
In \emph{clean2clean}, policies are evaluated in the clean environment; in \emph{clean2random}, they are evaluated in the randomized environment.
The latter setting tests generalization from clean demonstrations to environmental randomization.

Table~\ref{tab:robotwin2_full} reports per-task success rates of \ours for all 48 RoboTwin~2.0 tasks under \emph{clean2clean} and \emph{clean2random}.

\begin{table}[!ht]
\centering
\caption{\textbf{Per-task success rates (\%) of \ours on RoboTwin~2.0.}
The 48 tasks match Table~\ref{tab:robotwin2}.
E and H denote Easy (\emph{clean2clean}) and Hard (\emph{clean2random}), respectively.
Average reports the mean over all 48 tasks.}
\label{tab:robotwin2_full}
\label{tab:complete_result}
\begingroup
\small
\setlength{\tabcolsep}{3pt}
\renewcommand{\arraystretch}{1.10}
\begin{adjustbox}{max width=\linewidth}
\begin{tabular}{@{}lrr@{\hspace{12pt}}lrr@{\hspace{12pt}}lrr@{}}
\toprule
\textbf{Task} & \textbf{E} & \textbf{H} & \textbf{Task} & \textbf{E} & \textbf{H} & \textbf{Task} & \textbf{E} & \textbf{H} \\
\cmidrule(lr){1-3} \cmidrule(lr){4-6} \cmidrule(lr){7-9}
Adjust Bottle & 100 & 2 & Open Microwave & 77 & 3 & Place Object Scale & 33 & 1 \\
Beat Block Hammer & 82 & 1 & Pick Diverse Bottles & 86 & 8 & Place Object Stand & 79 & 12 \\
Blocks Ranking RGB & 63 & 1 & Pick Dual Bottles & 97 & 8 & Place Phone Stand & 88 & 2 \\
Blocks Ranking Size & 27 & 0 & Place A2B Left & 43 & 5 & Place Shoe & 56 & 4 \\
Click Alarmclock & 100 & 52 & Place A2B Right & 36 & 1 & Press Stapler & 89 & 41 \\
Click Bell & 97 & 41 & Place Bread Basket & 84 & 8 & Put Bottles Dustbin & 81 & 46 \\
Grab Roller & 100 & 22 & Place Bread Skillet & 75 & 3 & Rotate QRcode & 72 & 0 \\
Handover Block & 85 & 8 & Place Burger Fries & 93 & 15 & Scan Object & 67 & 2 \\
Handover Mic & 100 & 0 & Place Can Basket & 52 & 2 & Shake Bottle Horizontally & 99 & 81 \\
Hanging Mug & 33 & 0 & Place Cans Plasticbox & 89 & 3 & Shake Bottle & 97 & 72 \\
Lift Pot & 44 & 0 & Place Container Plate & 95 & 15 & Stack Blocks Three & 50 & 0 \\
Move Can Pot & 47 & 4 & Place Dual Shoes & 43 & 0 & Stack Blocks Two & 88 & 1 \\
Move Pillbottle Pad & 37 & 1 & Place Empty Cup & 100 & 0 & Stack Bowls Three & 54 & 3 \\
Move Playingcard Away & 85 & 6 & Place Fan & 24 & 1 & Stack Bowls Two & 86 & 9 \\
Move Stapler Pad & 31 & 0 & Place Mouse Pad & 26 & 0 & Stamp Seal & 12 & 0 \\
Open Laptop & 88 & 8 & Place Object Basket & 62 & 4 & Turn Switch & 69 & 29 \\
\midrule
\multicolumn{7}{l}{\textbf{Average over all 48 tasks}} & \textbf{69.19} & \textbf{10.94} \\
\bottomrule
\end{tabular}
\end{adjustbox}
\endgroup
\end{table}

\subsection{Real-world Task Performance}

We evaluate two tasks: \textit{1) placing pens} and \textit{2) stacking bowls} with the left and right arms.
We collect 60 demonstration trajectories per task for training.
We used LingBot-Depth~\cite{lingbot-depth2026} to process the raw depth data captured by the sensor.

\label{app:real_world_per}
\noindent\textbf{Qualitative results of \ours Tracker on real-world tasks.}
We provide a visualization of \ours Tracker on real-world tasks in Fig.~\ref{fig:real_tracker}.

\noindent\textbf{Task Rubric.}
The task-specific progress scoring rubric used in Fig.~\ref{fig:real_results} is summarized in Table~\ref{tab:our-real-world-progress-rubric}.

\begin{table}[!ht]
    \centering

    \caption{\textbf{Real-world task progress-score rubric.} Parentheses indicate the cumulative score after completing each ordered subtask.}
    \label{tab:our-real-world-progress-rubric}

    {
    \small
    \setlength{\tabcolsep}{3pt}

    \begin{tabularx}{\linewidth}{
        @{}
        >{\raggedright\arraybackslash}p{0.27\linewidth}
        >{\raggedright\arraybackslash}X
        @{}
    }
        \toprule
        \textbf{Task} & \textbf{Ordered subtasks} \\
        \midrule
        \shortstack[l]{Placing\_Pens\\(\(R=4\))}
        & \textbf{(1/4)} right arm picks up the pen on the right side of the table;
          \textbf{(2/4)} right arm places the pen in the right pen holder;
          \textbf{(3/4)} left arm picks up the pen on the left side of the table;
          \textbf{(4/4)} left arm places the pen in the left pen holder. \\
        \addlinespace
        \shortstack[l]{Stacking\_bowls\\(\(R=4\))}
        & \textbf{(2/4)} both arms simultaneously pick up two bowls;
          \textbf{(3/4)} right arm places its bowl on the plate;
          \textbf{(4/4)} left arm stacks its bowl on top of the first bowl. \\
        \bottomrule
    \end{tabularx}
    }
\end{table}

\begin{figure}[H]
\centering
\includegraphics[width=1\linewidth]{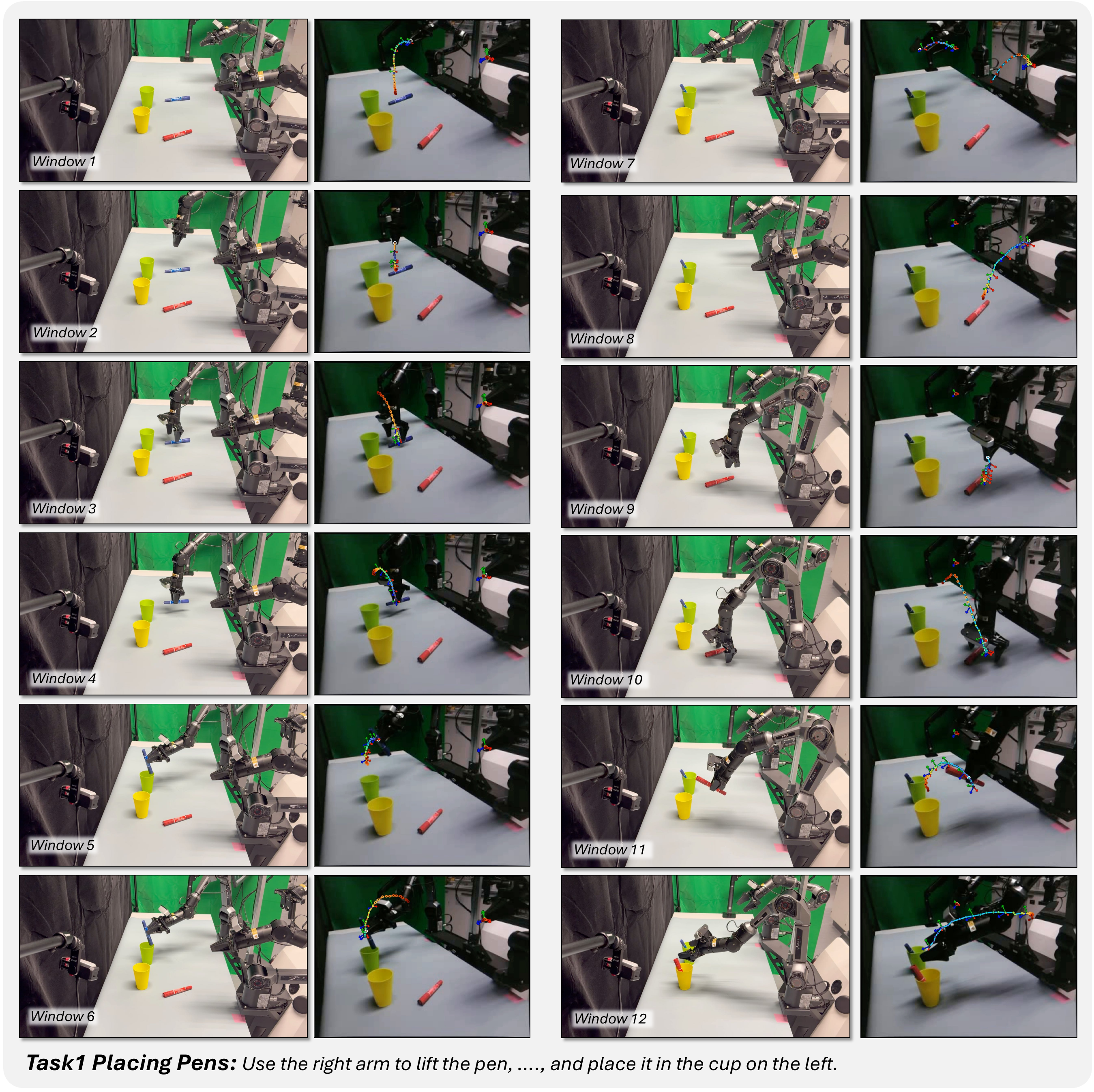}
\caption{\textbf{Visualizations of \ours on Task \textit{Placing Pens}.}  
Each window shows the future TCP trajectory predicted by \ours. The motion is then planned using the cuRobo~\cite{curobo_v2} library.}
\label{fig:task1_rollout}
\end{figure}

\begin{figure}[H]
\centering
\includegraphics[width=1\linewidth]{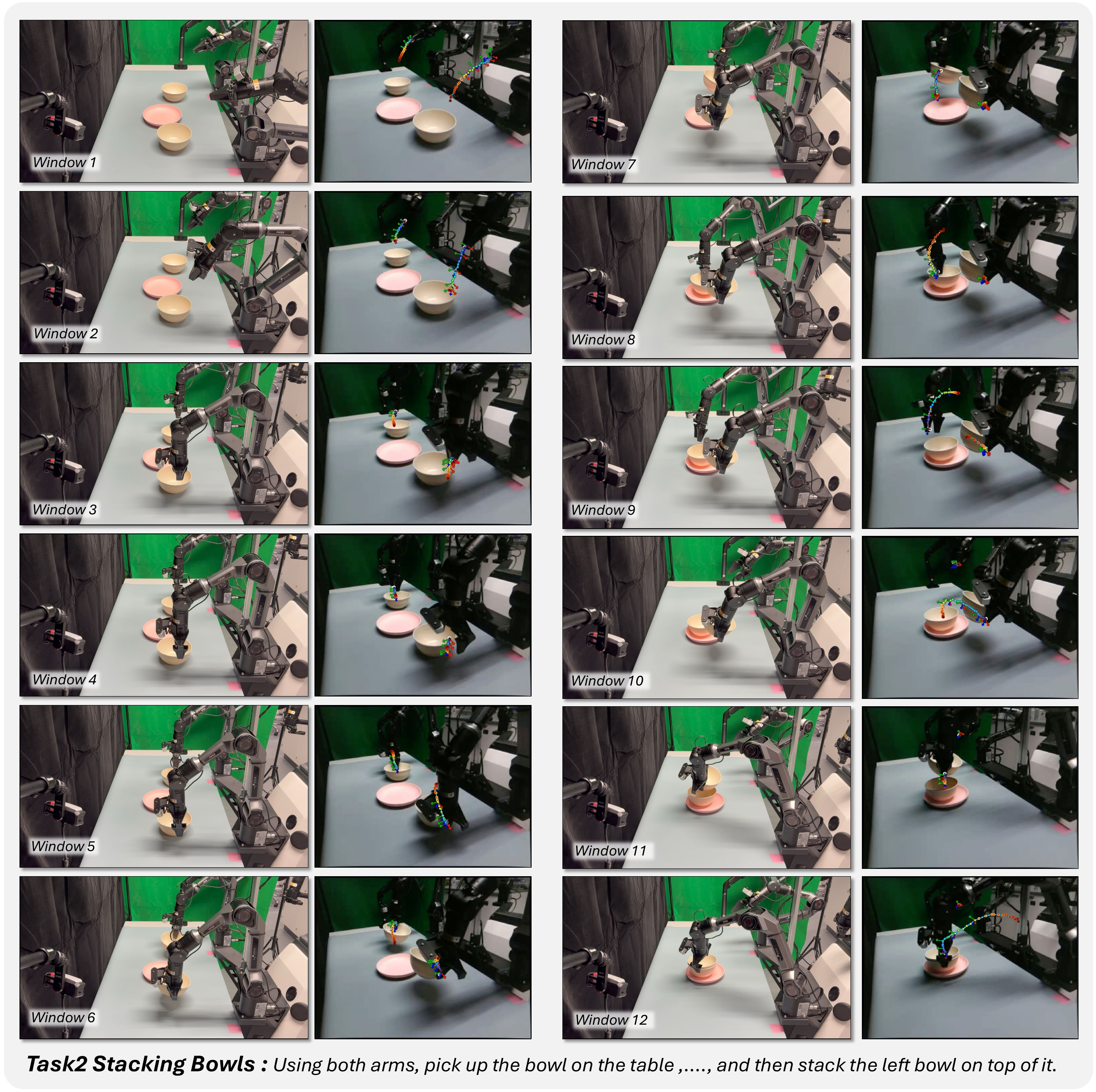}
\caption{\textbf{Visualizations of \ours on Task \textit{Stack Bowls}.}  
Each window shows the future TCP trajectory predicted by \ours. The motion is then planned using the cuRobo~\cite{curobo_v2} library.}
\label{fig:task2_rollout}
\end{figure}

\noindent\textbf{Real-world Task visualizations.}
We provide complete rollout visualizations for both tasks in Fig.~\ref{fig:task1_rollout} and Fig.~\ref{fig:task2_rollout}

\FloatBarrier

\par\medskip
\noindent\begin{minipage}{\linewidth}
\section{Limitations and Future Work}
\label{app:limitations}
\noindent\textbf{Embodiment Coverage.}
While \ours demonstrates strong performance in DROID setting, its training data currently covers only embodiments with Franka and Robotiq grippers, and other embodiment data remains limited.
After pretraining on DOIRD data, the \ours adapts effectively to UR5-WSG with only a few demonstrations (Table~\ref{tab:autolab_tracking}).
These results support the use of a shared \SE TCP representation for cross-embodiment transfer. With sufficient training data from a diverse range of embodiments, we expect \ours to generalize well to new robot platforms and gripper designs.

\noindent\textbf{Environmental Generalization.}
The \ours Generator shows strong performance under RoboTwin \emph{clean2clean}, but its performance under \emph{clean2random} remains limited.
The Generator is trained on a relatively small set of clean demonstrations without large-scale policy pretraining, providing little exposure to the environmental variations encountered during randomized evaluation.
This limited training diversity may partly explain the performance gap.
Large-scale pretraining on diverse manipulation data, together with domain randomization and targeted data augmentation, offers practical directions for addressing this gap and realizing the broader potential of 4D-conditioned TCP generation.

\end{minipage}

\end{document}